\documentclass[10pt]{article}
\usepackage[a4paper, top=2.5cm, bottom=2.5cm, left=2.5cm, right=2.5cm]{geometry}
\usepackage{amssymb, amsmath}
\usepackage{graphicx}
\usepackage{tabularx}
\usepackage{booktabs}
\usepackage{array}
\usepackage{float}
\usepackage{multirow}
\usepackage{url}
\usepackage{caption}
\usepackage{subcaption}
\usepackage{comment}
\usepackage{makecell}
\usepackage{mathrsfs}
\usepackage[numbers]{natbib}
\usepackage{soul}
\usepackage{xcolor}
\usepackage{hyperref}
\hypersetup{hidelinks}
\usepackage{amsthm}
\usepackage{fancyhdr}
\theoremstyle{definition}
\newtheorem{definition}{Definition}[section]

\def\argmin{\mathop{\rm argmin}\limits}
\def\argmax{\mathop{\rm argmax}\limits}

\newcommand{\tb}{\textbf}

\title{Bilevel Optimization for Neural Architecture Search}

\author{Abhishek Shukla$^{1}$ \and Ankur Sinha$^{2}$ \and Faiz Hamid$^{1}$}

\date{
$^{1}$Department of Management Sciences, IIT Kanpur, India\\
\texttt{abhiskl@iitk.ac.in, fhamid@iitk.ac.in}\\[6pt]
$^{2}$Krishnamurthy Tandon School of AI, IIM Ahmedabad, India\\
\texttt{asinha@iima.ac.in}
}

\begin{document}

\maketitle
\begin{abstract}
Bilevel optimization has become an influential and widely adopted framework for addressing hierarchical optimization problems in machine learning, providing an effective approach to modeling the interaction between two levels of optimization, with applications such as hyperparameter tuning, meta-learning, adversarial training, and data poisoning. Neural Architecture Search (NAS), a subfield of hyperparameter optimization, is a prime example of a bilevel optimization problem, with architecture parameters optimized at the outer-level and network weights optimized at the inner level. This paper presents a structured overview of NAS through the lens of bilevel optimization. We categorize existing NAS approaches into two main classes: sampling-based methods, which search optimal architectures using different architecture samplers, and bilevel theory-based methods, which solve the architecture search problem using bilevel optimization principles. We further highlight our current research direction, wherein the bilevel NAS formulation is addressed through an auxiliary mathematical programming framework. This framework enables the systematic integration of second-order information from the model's training loss function and ensures the optimality of the model parameters while modifying architecture parameters. By simultaneously updating the architecture and model parameters along their respective optimal descent directions derived from the auxiliary mathematical program, these methods achieve more principled and theoretically consistent results. The same auxiliary program can also be used for simultaneous hyperparameter and model fine-tuning. A comparative analysis shows that bilevel theory-based approaches generally outperform sampling-based methods, both in accuracy and efficiency. The paper concludes by outlining future research opportunities, emphasizing the expanding role of bilevel optimization in driving advances in NAS.
\end{abstract}


\section{Introduction}
Optimization holds a pivotal role in machine learning and mathematical modeling, encompassing a wide spectrum of problem formulations. Broadly, optimization tasks can be categorized according to the number of decision-making levels involved, leading to single-level and bilevel (two level) formulations, each characterized by distinct structural properties and application domains. The single-level formulation assumes that all decision variables are optimized jointly in a single layer of decision-making. Thus, it has a single-level of optimization task. On the contrary, in many real-world situations, decision-making involves two individuals: a leader and a follower. The leader decides first, and the follower responds by choosing the best possible action based on the leader’s choice. The interaction between the leader and follower forms a Bilevel Optimization Problem (BOP), where the leader’s decision-making task is called the upper-level problem, and the follower’s response is modeled as the lower-level problem. A real-life example of a BOP can be observed in pricing and production strategies within a supply chain. In this hierarchical structure, the manufacturer (leader) determines the wholesale price of a product, anticipating how the retailer (follower) will respond when deciding the order quantity. Given the manufacturer’s price, the retailer selects the quantity that maximizes its own profit. The manufacturer, in turn, chooses the price that maximizes its profit while accounting for the retailer’s optimal reaction. This setup captures the essence of bilevel decision-making, where the leader’s strategy depends on the follower’s best response. Therefore, unlike single-level optimization, where all variables are optimized at one-level, BOP explicitly consists of two hierarchical levels of optimization tasks. Owing to the hierarchical formulation, bilevel programming provides a unifying framework that allows the representation of many practical optimization problems that cannot be represented in single-level; in fact, every single-level optimization problem can also be viewed as a special case of bilevel programming~\citep{bracken1973mathematical}. It is noteworthy, however, that the follower’s optimal decision may not always align with the leader’s interests, as seen in problems such as toll pricing and adversarial learning. 
\begin{figure}
    \centering
    \includegraphics[width=0.8\linewidth]{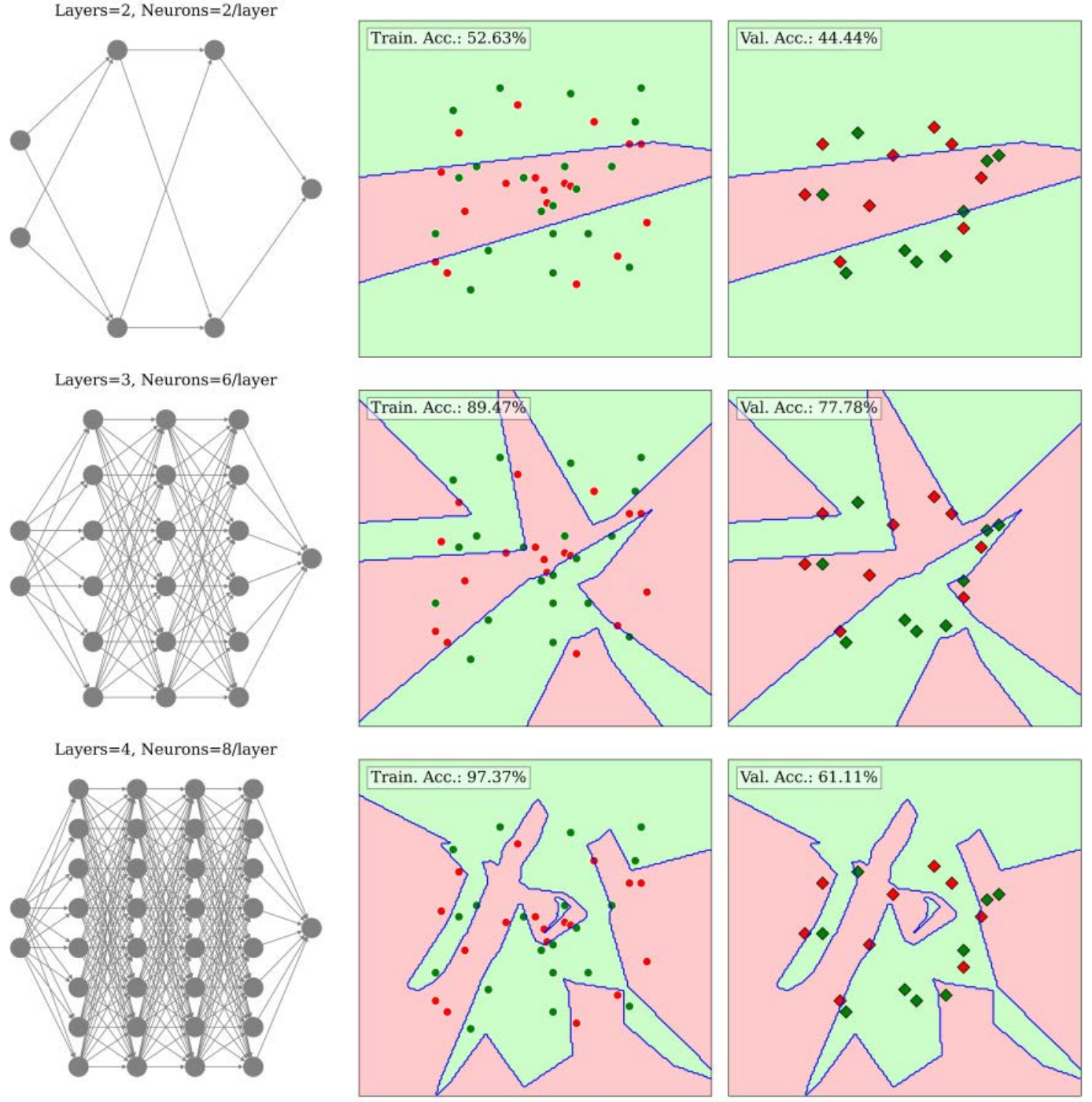}
    \caption{Representation power and overfitting in ANNs.}
    \label{fig:ann}
\end{figure}

Artificial Neural Networks (ANNs) constitute a subset of machine learning models that are conceptually influenced by the neural structure and functional characteristics of the human brain, i.e., they are computational models of biological neural networks. These networks process information by transmitting data through multiple layers of interconnected units, thereby emulating the signal propagation observed in biological neural systems. The trainable elements of an ANN, known as model parameters, primarily include the weights and biases associated with these connections. These parameters are iteratively adjusted during training to minimize a predefined loss function, thereby enabling the network to learn from data through error-driven updates. In contrast, hyperparameters are configuration settings that must be specified prior to training, as they are not directly learned from the data. They can be broadly categorized into three types: architectural parameters (e.g., the number of layers and neurons per layer), optimization hyperparameters (such as learning rate and momentum), and regularization hyperparameters (including weight decay and dropout)~\citep{sinha2024gradient}. The architecture of an ANN is primarily determined by factors such as the number of hidden layers, the number of neurons within each layer, the sequence of layers, and the choice of activation functions. These architectural design choices play a central role in shaping the network’s capacity to learn and represent complex, nonlinear relationships within the input data. Although increasing the network’s depth or width can enhance its expressive power and performance on training data, it also increases the risk of overfitting, a condition in which the model memorizes noise or spurious patterns in the training set, thereby impairing its ability to generalize to unseen data. This phenomenon is illustrated in Figure~\ref{fig:ann}, which is obtained by training and evaluating neural network models on the synthetic dataset described in Appendix~\ref{app_overfit_data}. The figure visualizes how the decision boundaries evolve as the number of layers and neurons increases, highlighting the growing representational capacity of the network. As the architecture becomes more complex, the decision boundaries adapt more finely to the training data, improving training accuracy but also increasing the risk of overfitting and degraded generalization to unseen data. In the third configuration, small isolated regions (islands) emerge in the decision boundary to correctly classify individual training points, clearly illustrating how excessive model flexibility can sacrifice generalization for perfect training performance.

\begin{figure}
    \centering
    \includegraphics[width=0.85\linewidth]{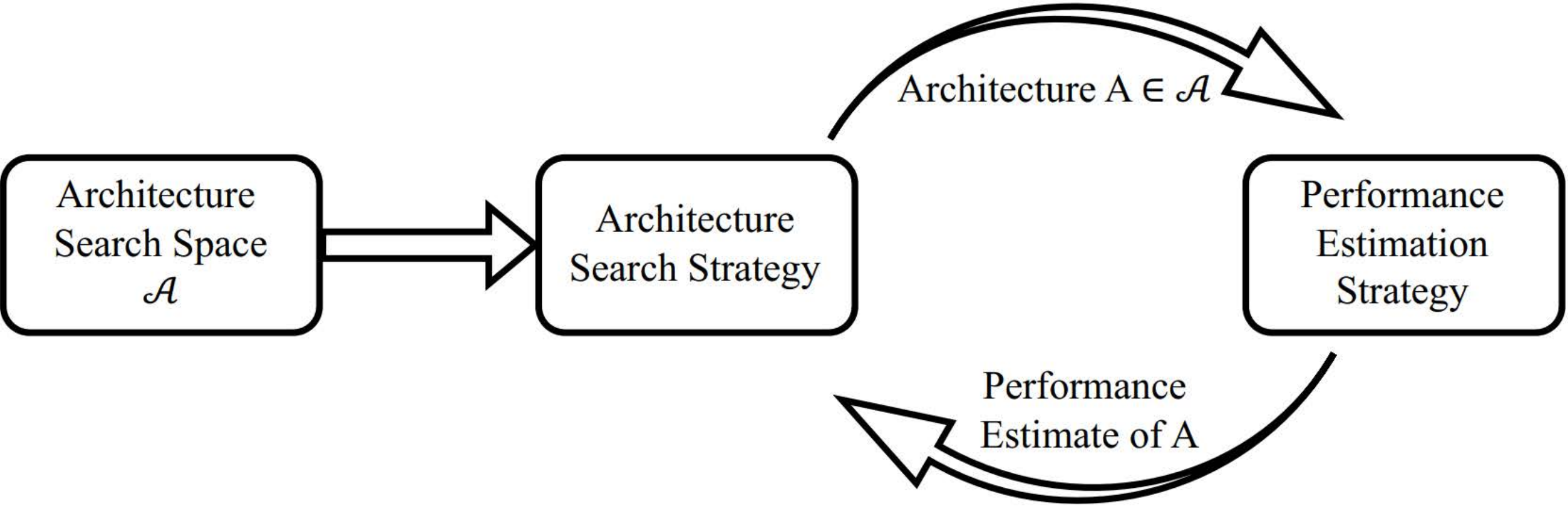}
    \caption{Conceptual illustration of NAS.}
    \label{fig:nas}
\end{figure}

Designing an effective neural network architecture is a critical step in developing high-performing deep learning models. Traditionally, this process has relied heavily on expert intuition, manual experimentation, and empirical trial-and-error. Neural Architecture Search (NAS), a sub-field of Hyperparameter Optimization (HPO) and Automated Machine Learning (AutoML), seeks to automate this design process. The central objective of NAS is to identify the optimal architecture from a vast and complex search space of possible configurations, tailored to a specific task and dataset. Figure~\ref{fig:nas} illustrates the conceptual framework of neural architecture search \citep{elsken2019neural}.
Formally, NAS is often formulated as a BOP, in which the upper-level seeks the optimal network architecture, while the lower-level focuses on training the corresponding model parameters to optimality. However, the NAS search space is typically high-dimensional and discrete, rendering the optimization process both computationally demanding and algorithmically challenging. Despite these complexities, NAS has demonstrated remarkable success in automating architectural design, substantially reducing human intervention, and producing models that match or even outperform manually crafted architectures. Building on these insights, this paper presents a unified view of NAS from the standpoint of bilevel optimization, with a particular focus on differentiable approaches that have a potential to significantly advance the field.

The remainder of this paper is organized as follows. Section~\ref{sec:bilevel_basics} introduces the fundamental concepts of bilevel optimization, establishing the theoretical foundation for subsequent discussions. Section~\ref{sec:neural_architecture_optimization} reviews both sampling-based and bilevel-theoretic approaches to NAS, highlighting their evolution from traditional hyperparameter optimization techniques to modern bilevel techniques. Section~\ref{sec:comparative} synthesizes findings from existing studies, providing a comparative overview of the reported performance of different NAS methodologies. Furthermore, Section~\ref{sec:emerging_trends} discusses the summary and emerging trends, highlighting the effectiveness of the auxiliary mathematical program-based approaches. Finally, Section~\ref{sec:conclusions} concludes the paper by summarizing key insights and outlining promising avenues for future research.

\section{Basics of Bilevel Optimization}\label{sec:bilevel_basics}
In this section, we formally discuss BOPs. For completeness, we first define a single-level optimization problem to distinguish it mathematically from a BOP.
\begin{definition}
A single-level optimization problem is defined as
\begin{equation}\label{single}
\min_{x, y} \; F(x, y) \quad \text{subject to} \quad G(x, y) \leq 0, \quad H(x, y) = 0,
\end{equation}
where $x = (x_1, \dots, x_n)^\top \in \mathbb{R}^n$ and 
$y = (y_1, \dots, y_m)^\top \in \mathbb{R}^m$ are the decision variables; 
$F : \mathbb{R}^n \times \mathbb{R}^m \to \mathbb{R}$ is the objective function; 
$G = (G_1, \dots, G_p)^\top : \mathbb{R}^n \times \mathbb{R}^m \to \mathbb{R}^p$ defines $p$ inequality constraints, i.e., $G_i(x,y) \leq 0$ for $i = 1, \dots, p$; 
and $H = (H_1, \dots, H_q)^\top : \mathbb{R}^n \times \mathbb{R}^m \to \mathbb{R}^q$ defines $q$ equality constraints, i.e., $H_j(x,y) = 0$ for $j = 1, \dots, q$.
\end{definition}

\begin{definition}
A bilevel optimization problem is defined as
\begin{equation}\label{bilevel}
\operatorname*{\text{``min''}}_{x} \; F(x, \hat{y}) \quad \text{subject to} \quad 
\begin{cases}
\hat{y} \in \argmin_{y} \left\{ f(x, y) \mid g(x, y) \leq 0,\; h(x, y) = 0 \right\}, \\
G(x, \hat{y}) \leq 0,\; H(x, \hat{y}) = 0,
\end{cases}
\end{equation}
where $x = (x_1, \dots, x_n)^\top \in \mathbb{R}^n$ and 
$y = (y_1, \dots, y_m)^\top \in \mathbb{R}^m$ are the upper-level and lower-level decision variables, respectively; 
$F : \mathbb{R}^n \times \mathbb{R}^m \to \mathbb{R}$ is the upper-level objective function; 
$f : \mathbb{R}^n \times \mathbb{R}^m \to \mathbb{R}$ is the lower-level objective function; 
$G = (G_1, \dots, G_p)^\top : \mathbb{R}^n \times \mathbb{R}^m \to \mathbb{R}^p$ and 
$H = (H_1, \dots, H_q)^\top : \mathbb{R}^n \times \mathbb{R}^m \to \mathbb{R}^q$ define the $p$ inequality and $q$ equality constraints of the upper-level problem, i.e., $G_i(x,\hat{y}) \leq 0$ for $i = 1, \dots, p$ and $H_j(x,\hat{y}) = 0$ for $j = 1, \dots, q$; 
$g = (g_1, \dots, g_r)^\top : \mathbb{R}^n \times \mathbb{R}^m \to \mathbb{R}^r$ and 
$h = (h_1, \dots, h_s)^\top : \mathbb{R}^n \times \mathbb{R}^m \to \mathbb{R}^s$ define the $r$ inequality and $s$ equality constraints of the lower-level problem, i.e., $g_k(x,y) \leq 0$ for $k = 1, \dots, r$ and $h_\ell(x,y) = 0$ for $\ell = 1, \dots, s$.
\end{definition}

\begin{table}[ht]
\centering
\caption{Key mathematical concepts and notations used in the formulation of BOPs.}
\label{tab:notations}
\renewcommand{\arraystretch}{1.6}
\begin{tabular}{@{}p{4.0cm} p{11.5cm}@{}}
\toprule
\textbf{Name and Notation} & \textbf{Description} \\
\midrule

Lower-level feasible region \newline
$\Omega : \mathbb{R}^n \rightrightarrows \mathbb{R}^m$ 
& $\Omega(x) = \{ y : g(x, y) \leq 0,\; h(x, y) = 0 \}$ \newline
Defines the set of feasible lower-level decisions for a given upper-level decision and is a set-valued (multi-valued) mapping from $\mathbb{R}^n$ into the power set of $\mathbb{R}^m$ \\

\addlinespace

Lower-level reaction set \newline
$\Psi : \mathbb{R}^n \rightrightarrows \mathbb{R}^m$ 
& $\Psi(x) = \left\{ \hat{y} : \hat{y} \in \argmin\limits_{y} \{ f(x, y) : y \in \Omega(x) \} \right\}$ \newline
Represents the set of optimal lower-level responses for a given upper-level decision and is a set-valued (multi-valued) mapping from $\mathbb{R}^n$ into the power set of $\mathbb{R}^m$ \\

\addlinespace

Relaxed feasible set \newline
$\Phi = \mathrm{gph}\, \Omega$ 
& $\Phi = \{ (x, y) : G(x, y) \leq 0,\; H(x, y) = 0,\; g(x, y) \leq 0,\; h(x, y) = 0 \}$ \newline
Describes the feasible region that jointly satisfies both upper- and lower-level constraints excluding the lower-level optimality constraint \\

\addlinespace

Inducible region \newline
$I = \mathrm{gph}\, \Psi$ 
& $I = \{ (x, \hat{y}) : (x, \hat{y}) \in \Phi,\; \hat{y} \in \Psi(x) \}$ \newline
Contains all pairs of upper-level decisions and associated lower-level optimal responses \\

\addlinespace

Choice function \newline
$\psi : \mathbb{R}^n \rightarrow \mathbb{R}^m$ 
& $\psi(x) \in \Psi(x)$ \newline
Defines a single-valued mapping that selects one optimal lower-level response, representing a deterministic follower behavior \\

\addlinespace

Optimal value function \newline
$\varphi : \mathbb{R}^n \rightarrow \mathbb{R}$ 
& $\varphi(x) = \min\limits_{y \in \Omega(x)} f(x, y)$ \newline
Outputs the minimum value of the lower-level objective function over the lower-level feasible set, given an upper-level decision \\

\addlinespace

Induced value function \newline
$\theta : \mathbb{R}^n \rightarrow \mathbb{R}$ 
& $\theta(x) = \min\limits_{\hat{y} \in \Psi(x)} F(x, \hat{y})$ \newline
Associates each upper-level decision with the minimum value of the upper-level objective function over the follower’s optimal response set \\

\bottomrule
\end{tabular}
\end{table}
In Definition 2, the quotes in the upper-level objective indicate ambiguity arising when multiple optimal solutions exist at the lower-level for a given upper-level decision. In such cases, the leader’s outcome depends on which solution is selected by the follower---cooperative or adversarial---thereby requiring an explicit selection rule to fully define the problem. At its core, a BOP involves upper-level and lower-level decision variables, where the follower solves a parametric optimization problem while treating the leader’s decision as a parameter. Crucially, a solution is feasible for the upper-level problem if and only if it corresponds to an optimal solution of the lower-level problem while satisfying the upper-level constraints. This nested structure distinguishes BOPs from standard single-level formulations.

\begin{figure}
    \centering
    \begin{subfigure}[c]{0.49\textwidth}
        \centering
        \includegraphics[width=\linewidth]{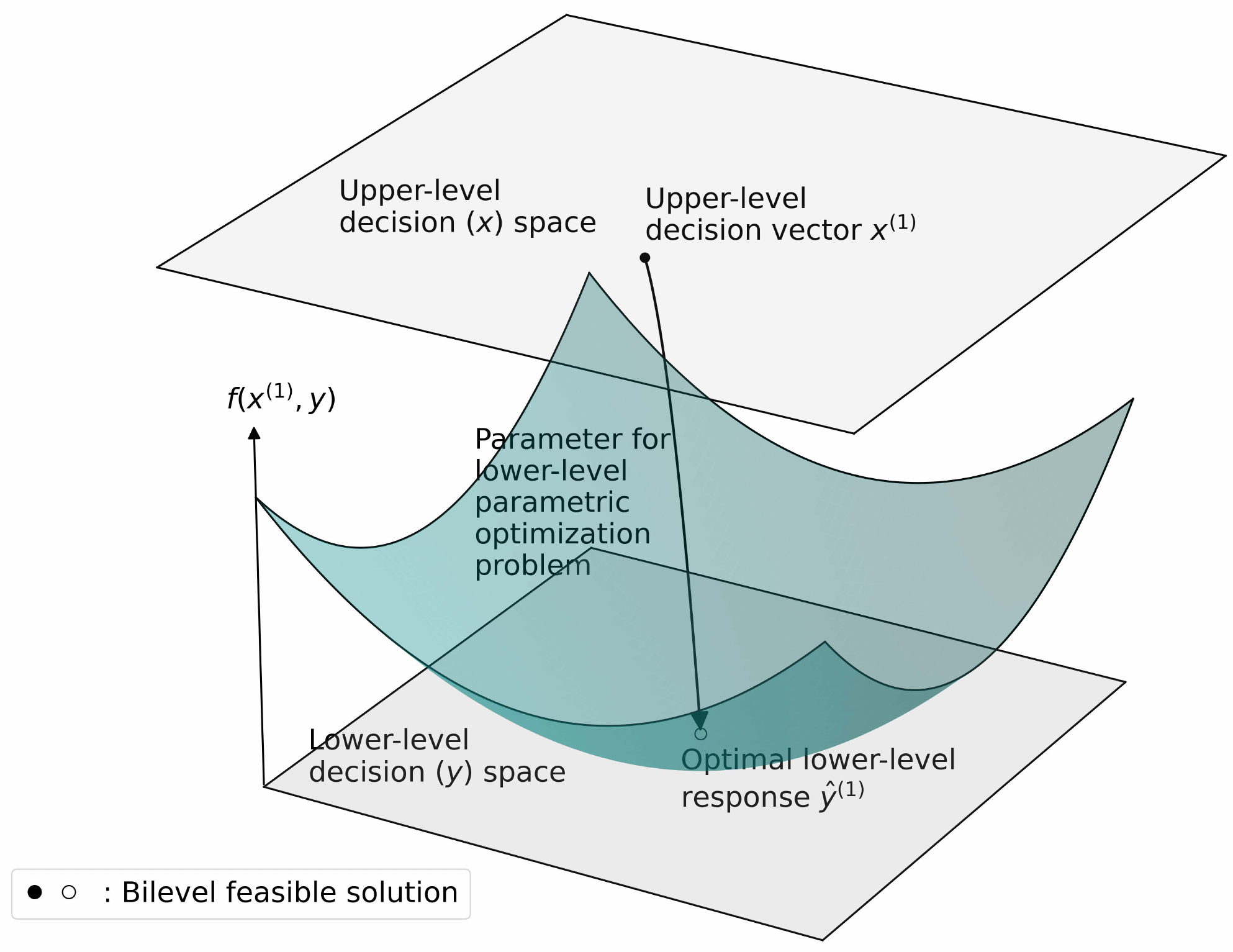}
        \caption{Single Lower-level Optimal Solution}
        \label{fig:single_llo}
    \end{subfigure}
    \hfill
    \begin{subfigure}[c]{0.49\textwidth}
        \centering
        \includegraphics[width=\linewidth]{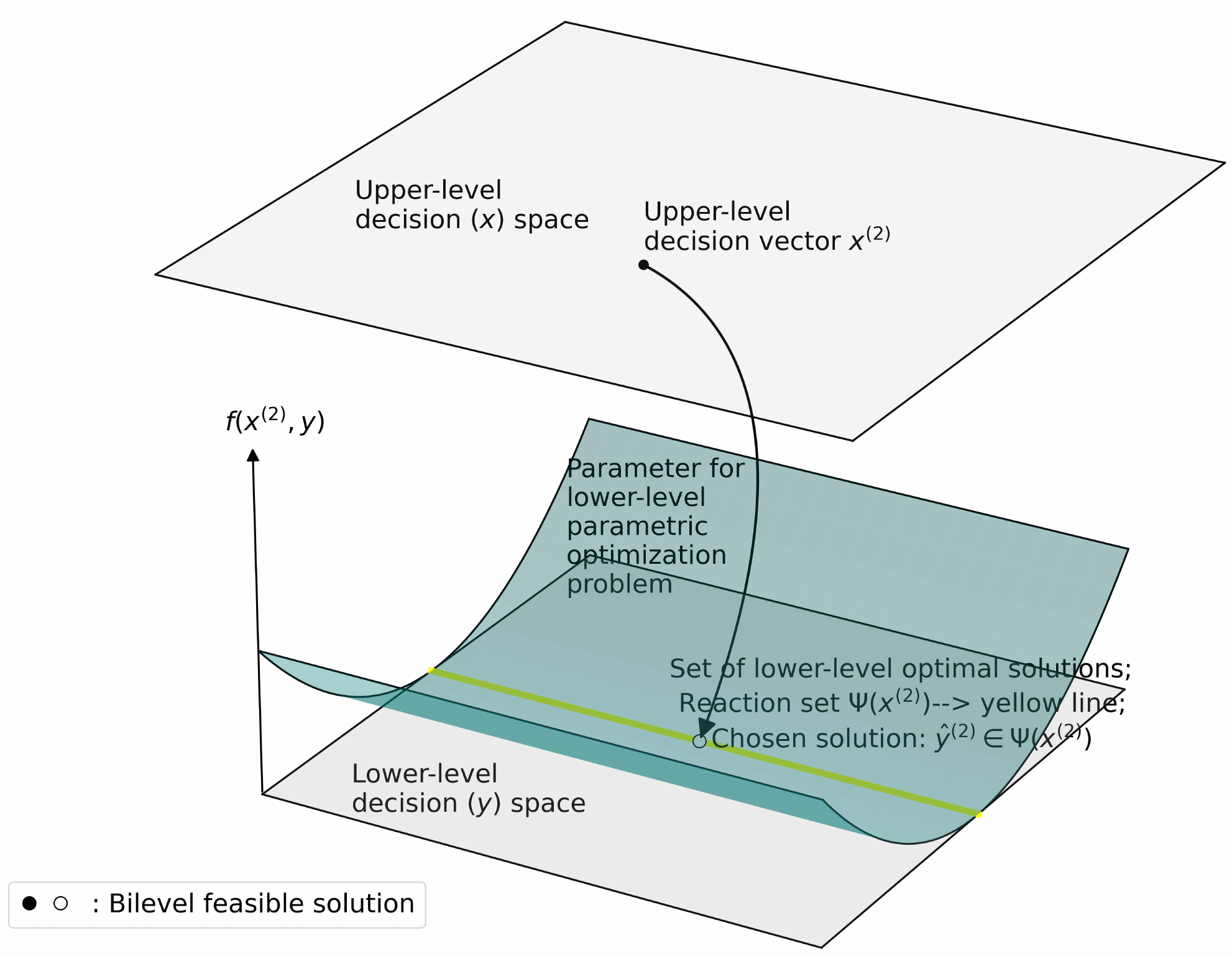}
        \caption{Multiple Lower-level Optimal Solutions}
        \label{fig:multiple_llo}
    \end{subfigure}
    \caption{Illustration of unique vs. multiple lower-level optimal solutions of a BOP.}
    \label{fig:llo_comparison}
\end{figure}

\begin{figure}
    \centering
    \includegraphics[width=0.8\linewidth]{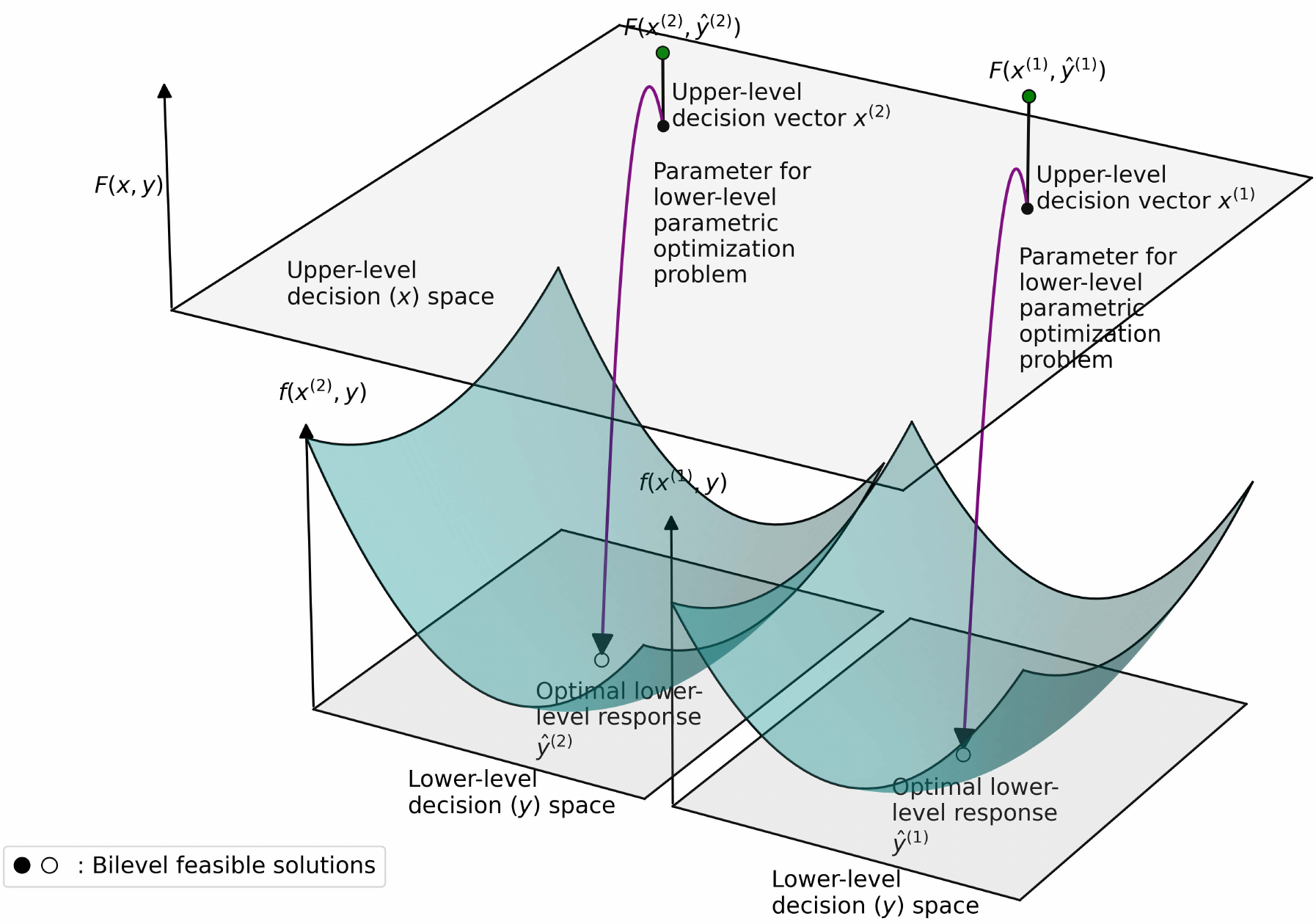}
    \caption{Interdependence between the upper- and lower-level problems.}
    \label{fig:blp}
\end{figure}

The BOP, as defined in~\eqref{bilevel}, establishes a hierarchical framework involving two interlinked levels of optimization. Table~\ref{tab:notations} summarizes the key symbols and terms commonly employed in the theory of bilevel optimization as described in \cite{sinha2017review}. The reaction set $\Psi(x)$, defined in Table~\ref{tab:notations}, is a set-valued mapping that depends explicitly on $x$ and is, in most practical cases, nonconvex and possibly disconnected. For the upper-level problem to remain feasible, the reaction set must be nonempty, i.e., \(\Psi(x) \neq \emptyset\). Moreover, \(\Psi(x)\) need not be unique. It may contain a single or multiple optimal responses for any given $x$, as illustrated in Figure~\ref{fig:llo_comparison}. A given upper-level and lower-level decision pair \((x,\hat{y})\) is therefore feasible to the leader only when $x$ is associated with \(\hat{y}\) that corresponds to an optimal solution of the lower-level problem, while satisfying the upper-level constraints too. A bilevel optimal solution thus represents a pair \((x^*, y^*)\) in which both the leader’s decision and the follower’s optimal response are jointly optimal, reflecting the intrinsic coupling between the two levels. This interdependence is depicted in Figure~\ref{fig:blp}, which highlights how variations in one level influence the feasibility and optimality of the other. Such hierarchical coupling substantially increases the analytical and computational complexity of bilevel optimization algorithms. The upper-level objective is no longer independent but is shaped by the outcome of an embedded lower-level optimization, which introduces several structural and algorithmic challenges, as summarized in Table~\ref{tab:blp_properties}.

\begin{table}[ht]
\centering
\caption{Key properties of BOPs.}
\label{tab:blp_properties}
\renewcommand{\arraystretch}{1.3}
\begin{tabular}{@{}p{4.0cm} p{11.0cm}@{}}
\toprule
\textbf{Property} & \textbf{Description} \\
\midrule

Hierarchical and nested & The upper-level and lower-level problems are interlinked, forming a nested decision structure \\

Non-convexity & The feasible region is typically non-convex and may even be disconnected, complicating solution methods \\

Computational complexity & BOPs are strongly NP-hard, making them computationally intractable in general cases \\

Feasibility & A solution is feasible only if it simultaneously satisfies the constraints of both levels, while satisfying lower-level optimality \\

Multiple lower-level optima & The existence of several follower optima adds ambiguity and complexity to the problem \\

Modes of interaction & Optimistic (Cooperative), where the follower’s choice is favorable to the leader; Pessimistic (Conflicting), where the follower’s choice is unfavorable to the leader \\

Non-differentiability & Arises from the implicit dependence of upper-level on the follower’s optimal response \\

Reformulation needs & Often necessitates reformulation using the Karush-Kuhn-Tucker (KKT) conditions, duality, or various mappings to improve tractability \\

Sensitivity & Apart from objectives and constraints, the stability and convergence of algorithms are affected by the sensitivity of the lower-level solution mapping \\

\bottomrule
\end{tabular}
\end{table}

This bilevel structure is the characteristic of many practical scenarios. Some of them are given in Table~\ref{tab:applications}. For instance, in HPO and NAS, model parameters are first optimized in response to a given hyperparameter configuration, and then the outer-level variables (hyperparameters or architecture parameters) are tuned to improve generalization in an iterative manner. In adversarial learning, robustness is achieved by proactively accounting for the worst-case perturbations that an adversary might introduce. In inverse optimization, the leader seeks to estimate model parameters such that the resulting optimal decisions replicate observed behavior as closely as possible, effectively learning from data to uncover underlying decision-making rules. In network design and resource allocation, the leader optimizes infrastructural variables while accounting for the system-optimal response of users, such as traffic routing. In electricity market bidding, a firm maximizes its revenue by anticipating how the market operator will allocate resources to minimize overall cost. In security resource allocation, the defender allocates available resources to minimize potential damage, considering the most damaging response of a rational attacker. To illustrate how bilevel problems are modeled and solved in practice, we present a representative case study in Appendix~\ref{app_toll_setting_problem}. This case study arises in the transportation domain and focuses on a toll-setting problem~\cite{sinha2015transportation,brotcorne2001bilevel,labbe1998bilevel}.
\begin{table}[ht]
\centering
\caption{Applications of bilevel optimization.}
\label{tab:applications}
\renewcommand{\arraystretch}{1.6}
\begin{tabular}{@{}>{\raggedright\arraybackslash}p{4.5cm} >{\raggedright\arraybackslash}p{11.0cm}@{}}
\toprule
\textbf{Application} & \textbf{Description} \\
\midrule

Hyperparameter optimization 
& $\displaystyle \min_{\lambda} \ \mathcal{L}_{v}(\lambda, \hat{W})$ \newline
$\text{s.t. } \hat{W} \in \argmin_{W} \ \mathcal{L}_{t}(\lambda, W)$ \newline
Tune hyperparameters $\lambda$ to minimize validation loss, assuming model weights $\hat{W}$ are trained optimally \citep{bennett2008bilevel,franceschi2018bilevel,okuno2021lp} \\

\addlinespace

Adversarial learning 
& $\displaystyle \min_{\theta} \ \mathbb{E}_{x}[\ell(\theta, x + \hat{\delta})]$ \newline
$\text{s.t. } \hat{\delta} \in \argmax_{\|\delta\| \leq \epsilon} \ \ell(\theta, x + \delta)$ \newline
Learn robust model parameters $\theta$ by accounting for worst-case adversarial perturbations $\hat{\delta}$ \citep{madry2017towards} \\

\addlinespace

Inverse optimization
& $\displaystyle \min_{\theta} \ \|\hat{y} - \tilde{y}\|^2$ \newline
$\text{s.t. } \hat{y} \in \argmin_{y} \ f(\theta, y)$ \newline
Optimize the model parameters $\theta$ such that the model-generated decision $\hat{y}$ closely matches the observed decision $\tilde{y}$ \citep{suryan2016handling,chan2025inverse} \\

\addlinespace

Network design \& resource allocation 
& $\displaystyle \min_{x} \ C(x, \hat{y})$ \newline
$\text{s.t. } \hat{y} \in \argmin_{y} \ T(x, y)$ \newline
Optimize resource allocations $x$ to minimize cost, considering optimal traffic routing $\hat{y}$ for a given $x$ \citep{marcotte1986network} \\

\addlinespace

Electricity market bidding 
& $\displaystyle \max_{b} \ R(b, \hat{y})$ \newline
$\text{s.t. } \hat{y} \in \argmin_{y} \left\{ \sum_{i} b_i y_i : \text{power balance and capacity constraints} \right\}$ \newline
Determine optimal bidding prices $b$ to maximize revenue, anticipating the Independent System Operator’s (ISO) cost-minimizing generation dispatch $\hat{y}$ \citep{fampa2008bilevel} \\

\addlinespace

Security resource allocation (defender–attacker)
& $\displaystyle \min_{x \in \mathcal{X}} \, R(x, \hat{y})$ \newline
$\text{s.t. } \hat{y} \in \argmax_{y \in \mathcal{Y}(x)} \, U(x, y)$ \newline
Defender allocates resources $x$ to minimize damage from an optimal attacker response $\hat{y}$ \citep{ramamoorthy2018multiple,dahan2022network,ramamoorthy2024exact} \\

\bottomrule
\end{tabular}
\end{table}

\subsection{Modeling Frameworks for Bilevel Optimization Problems}
Depending on how the leader anticipates the follower’s behavior and the inherent nature of their relationship in situations where there are multiple lower-level optimal solutions for a given upper-level decision, different modeling paradigms can be adopted. At a high level, the interaction between the leader and follower can be viewed through two fundamental lenses: cooperation or conflict. These perspectives then give rise to the well-known mathematical interpretations of the BOP, commonly categorized as the optimistic and pessimistic formulations. 

\subsubsection{Optimistic and Pessimistic Formulations}
The distinction between cooperation and conflict matters only when the follower has multiple optimal responses to the leader’s decision. If the follower’s best response is unique, their behavior is entirely predictable. Regardless of whether the leader and follower act cooperatively or competitively, the outcome remains the same. However, when the follower is indifferent among several equally good options, uncertainty arises about which one they will pick. The optimistic and pessimistic formulations resolve this ambiguity by defining how the follower’s choice should be interpreted.
\begin{description}
    \item[\tb{The Optimistic Formulation (Modeling Cooperation):}] The optimistic formulation is used when the leader assumes a cooperative relationship. It posits that if the follower has multiple optimal solutions, they will benevolently choose the one that is most favorable for the leader. This is often called the best-case scenario for the leader. The lower-level optimal response in the optimistic setting (choice function $\psi_o(x)$) is defined in Equation~\eqref{opt_response}, and the optimistic bilevel problem is modeled in Formulation~\eqref{bilevel_optimistic}.

    \begin{equation} \label{opt_response}
            \psi_o(x) = \argmin\limits_{\hat{y} \in \Psi(x)} F(x, \hat{y})
    \end{equation}
    \begin{equation}\label{bilevel_optimistic}
    \min_{x} \; F(x, \hat{y_o}) \quad \text{subject to} \quad 
    \begin{cases}
    \hat{y_o} = \psi_o(x), \\[6pt]
    G(x, \hat{y_o}) \leq 0,\; H(x, \hat{y_o}) = 0
    \end{cases}
    \end{equation}

    \begin{description}
    \item[\tb{Example:}] Consider a government (leader) that wants to subsidize green energy production to minimize carbon emissions. It offers a subsidy rate to a power company (follower), whose objective is to maximize its profit. Suppose for a given subsidy rate, the company can achieve the same maximum profit by either building a large solar farm or a slightly smaller solar farm and a small natural gas plant. The optimistic assumption is that the cooperative, environmentally-conscious company will choose to build the large solar farm, as this option best aligns with the government's goal of minimizing emissions.
    \end{description}

    \item [\tb{The Pessimistic Formulation (Modeling Conflict):}] The pessimistic formulation is applied when the leader assumes a conflicting relationship. It represents a more cautious, defensive stance, where the leader anticipates that the follower will choose the option that is least favorable for the leader from their set of optimal solutions. This is the worst-case scenario, and the leader seeks a decision that is robust against this adversarial response. The lower-level optimal response in the pessimistic setting (choice function $\psi_p(x)$) is defined in Equation~\eqref{pess_response}, and the pessimistic bilevel problem is modeled in Formulation~\eqref{bilevel_pessimistic}.

    \begin{equation} \label{pess_response}
             \psi_p(x) = \argmax\limits_{\hat{y} \in \Psi(x)} F(x, \hat{y})
    \end{equation}
    \begin{equation}\label{bilevel_pessimistic}
    \min_{x} \; F(x, \hat{y_p}) \quad \text{subject to} \quad 
    \begin{cases}
    \hat{y_p} = \psi_p(x), \\[6pt]
    G(x, \hat{y_p}) \leq 0,\; H(x, \hat{y_p}) = 0
    \end{cases}
    \end{equation}

    \begin{description}
        \item[\tb{Example:}] Reconsider the same government (leader) and power company (follower) setup. For a given subsidy rate, suppose the company again finds two profit-maximizing options: building a large solar farm or a smaller solar farm combined with a small natural gas plant. Under the pessimistic assumption, the government does not presume cooperation but instead prepares for the less desirable outcome, as the company may opt for a mixed solar-gas configuration, which yields higher emissions. Notably, the company’s choice is not necessarily adversarial; however, the pessimistic government designs its subsidy policy to remain effective even under such a cautious, worst-case interpretation of the company’s behavior.
    \end{description}
\end{description}
A nuanced discussion of bilevel problem formulations is presented in \citep{som2025bilevel}, where the authors distinguish between two modeling perspectives, namely, the real and standard bilevel formulations (discussed in Appendix~\ref{app_real_standard_formulations}). These formulations are applicable to both optimistic and pessimistic settings.

\subsubsection{High-point Formulation}
A high-point relaxation is a single-level relaxation of the BOP stated as
\begin{equation}\label{bilevel_relax}
\min_{x, y} \; F(x, y) \quad \text{subject to} \quad 
\begin{cases}
g(x, y) \leq 0,\; h(x, y) = 0, \\
G(x, y) \leq 0,\; H(x, y) = 0
\end{cases}
\end{equation}
The optimal value of the hight-point relaxation \eqref{bilevel_relax} provides a valid lower bound on the optimal value of the original BOP \eqref{bilevel} as we have removed the lower-level optimality requirement that was acting as a constraint. It is noteworthy that the lower-level constraints are still maintained.

\subsubsection{Illustrative Example}
Consider the BOP~\eqref{illustrative_example}, which serves to illustrate the distinctions among the optimistic, pessimistic, and high-point formulations.
\begin{equation}
\label{illustrative_example}
\operatorname*{\text{``min''}}_{x} \; F(x, \hat{y}) = x^2 + \hat{y}^2 + \hat{y} - x + \tfrac{1}{2}
\end{equation}
subject to
\begin{equation*}
\begin{cases}
-1 \leq x \leq 1, \\
-x + \hat{y} \leq 1, \\
x - \hat{y} \leq 1, \\
\hat{y} \in \underset{y}{\argmin} \left\{ f(x, y) = (y - x)^4 - (y - x)^2 
\;\middle|\; x^2 + y^2 \leq 1,\; x + y \leq 1 \right\}
\end{cases}
\end{equation*}
This problem exhibits a typical structural asymmetry of bilevel formulations. Although the upper- and lower-level objectives and constraints appear simple, their convexity properties differ. In particular,
\begin{description}
    \item[\tb{Upper-level Problem:}] The objective function is convex in $(x,\hat{y})$, but the (bilevel) feasible region is non-convex because of the embedded lower-level optimality condition
    \item [\tb{Lower-level Problem:}] The constraints are convex, yet the objective is non-convex, which leads to multiple local/global optima.
\end{description}
This mixture of convex and non-convex components illustrates why BOPs remain computationally challenging, even when each level alone may look relatively simple. The solutions of the optimistic, pessimistic and high-point formulation of the problem are given in Table~\ref{tab:formulation_solutions}. Further details are provided in the Appendix~\ref{app_solution_opt_pess_formulations}. Let us solve the high-point formulation of the considered BOP.

In the high-point relaxation, the lower-level variable $y$ is allowed to be any value satisfying the upper- and lower-level constraints, regardless of lower-level optimality. The resulting problem is
\[
\begin{aligned}
\min_{x, y} \quad & x^2 + y^2 + y - x + \frac{1}{2}, \\
\text{s.t.} \quad & -1 \leq x \leq 1, \\
& -x + y \leq 1, \\
& x - y \leq 1, \\
& x^2 + y^2 \leq 1, \\
& x + y \leq 1 \\
\end{aligned}
\]
Rewriting the objective
\[
F(x, y) = \left(x - \tfrac{1}{2}\right)^2 + \left(y + \tfrac{1}{2}\right)^2
\]
The above function is minimized at $(x^*, y^*) = (\tfrac{1}{2}, -\tfrac{1}{2})$, yielding an optimal value of
$F(x^*, y^*) = 0$, which is a lower bound to the bilevel optimal solution (optimistic/pessimistic).
\begin{figure}
    \centering
  \includegraphics[width=0.6\textwidth]{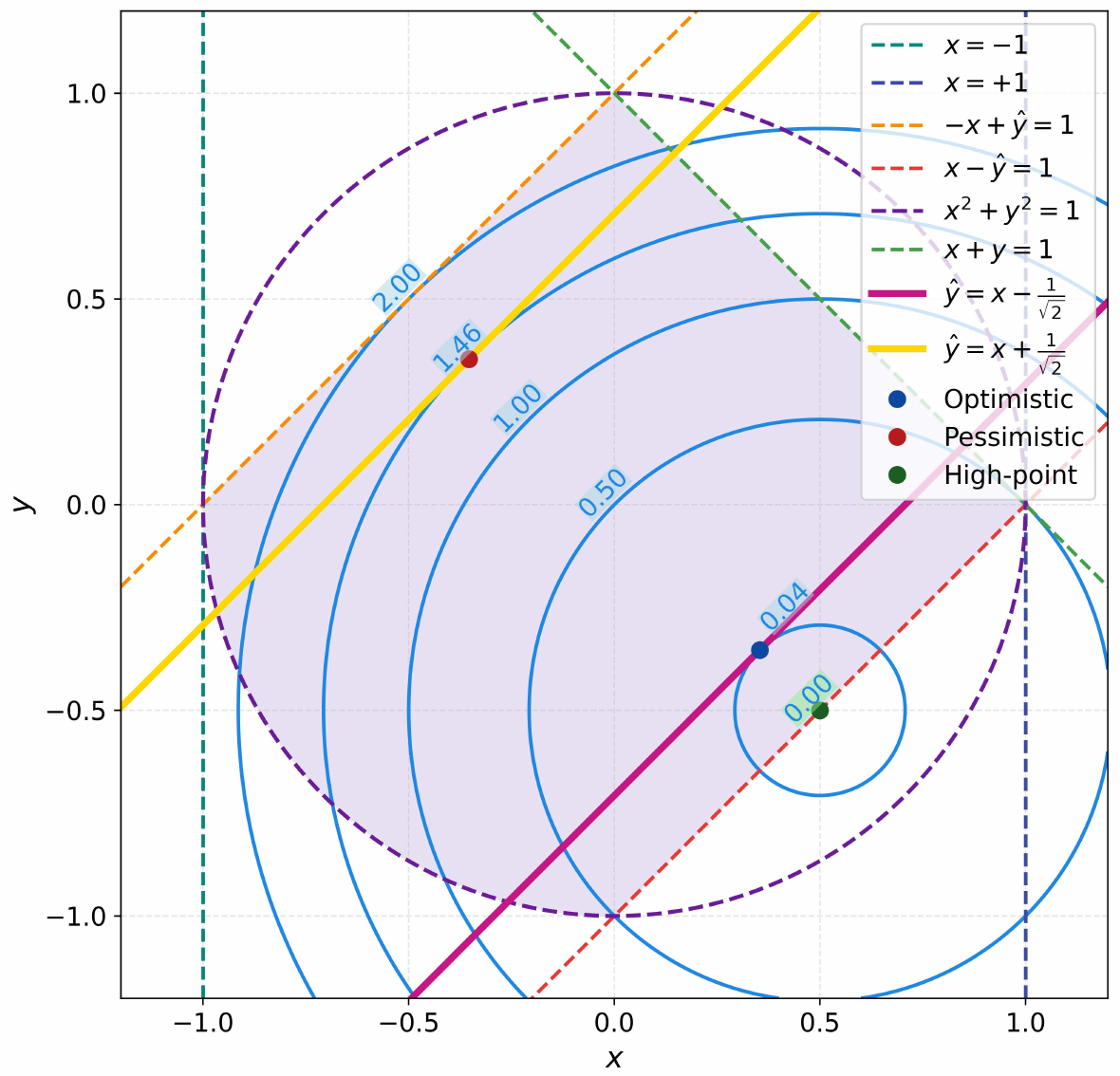}
    \caption{
    Geometric illustration of the considered BOP. }
    \label{fig:bilevel_plot}
\end{figure}
\begin{table}[ht]
\centering
\caption{Optimal solutions and objective values for different formulations.}
\label{tab:formulation_solutions}
\renewcommand{\arraystretch}{1.6}
\begin{tabular}{@{}p{4.5cm} p{11.0cm}@{}}
\toprule
\textbf{Formulation Type} & \textbf{Solution and Objective Values} \\
\midrule

Optimistic
& $(x^*, y^*) = \left(\tfrac{1}{2\sqrt{2}}, -\tfrac{1}{2\sqrt{2}}\right)$ \newline
$F^* = F(x^*, y^*) = \tfrac{3}{4} - \tfrac{1}{\sqrt{2}}$ \newline
$f^* = f(x^*, y^*) = \left(-\tfrac{1}{\sqrt{2}}\right)^4 - \left(-\tfrac{1}{\sqrt{2}}\right)^2 = \tfrac{1}{4} - \tfrac{1}{2} = -\tfrac{1}{4}$ \\

\addlinespace

Pessimistic 
& $(x^*, y^*) = \left(-\tfrac{1}{2\sqrt{2}}, \tfrac{1}{2\sqrt{2}}\right)$ \newline
$F^* = F(x^*, y^*) = \tfrac{3}{4} + \tfrac{1}{\sqrt{2}}$ \newline
$f^* = f(x^*, y^*) = \left(\tfrac{1}{\sqrt{2}}\right)^4 - \left(\tfrac{1}{\sqrt{2}}\right)^2 = \tfrac{1}{4} - \tfrac{1}{2} = -\tfrac{1}{4}$ \\

\addlinespace

High-Point
& $(x^*, y^*) = \left(\tfrac{1}{2}, -\tfrac{1}{2}\right)$ \newline
$F^* = F(x^*, y^*) = 0$ \newline
$f^* = f(x^*, y^*) = (-1)^4 - (-1)^2 = 1 - 1 = 0$ \\
\bottomrule
\end{tabular}
\end{table}
Figure \ref{fig:bilevel_plot} illustrates  the BOP geometrically. The contours correspond to the upper level objective and the shaded area represents the relaxed bilevel feasible region, where both upper- and lower-level constraints are satisfied, but the lower-level optimality condition is relaxed. The segments of the lines $\hat{y} = x + \frac{1}{\sqrt{2}}$ and $\hat{y} = x - \frac{1}{\sqrt{2}}$ within this region form the reaction set, indicating all possible lower-level optimal responses to a given upper-level decision. Medium blue circles depict contours of the upper-level objective. Three solutions are highlighted: the optimistic solution (blue), where the lower-level player acts in favor of the upper-level; the pessimistic solution (red), representing the worst-case lower-level response; and the high-point solution (green), corresponding to the single-level relaxation of the BOP.

This example vividly illustrates how differing assumptions regarding the follower’s behavior (optimistic, pessimistic, or based on a relaxed high-point formulation) can lead to substantially different optimal solutions. These distinctions are not merely theoretical but arise naturally across various application domains. Some of them have already been discussed earlier. 

\subsection{Solution Methods for Bilevel Optimization Problems}
\begin{table}[htb]
\centering
\caption{Solution methods for BOPs.}
\label{tab:bilevel_methods}
\renewcommand{\arraystretch}{1.6}
\begin{tabular}{@{}p{6.0cm} p{10.0cm}@{}}
\toprule
\textbf{Name and Category} & \textbf{Description and References} \\
\midrule
KKT-based Methods
& Utilize KKT conditions to reformulate and solve BOPs efficiently \citep{allende2013solving, dempe2014kkt, sinha2019using} \\
\addlinespace
Gradient Descent Methods
& Apply gradient-based optimization techniques for solving BOPs in differentiable settings \citep{SaGa94, ViSaJu94} \\
\addlinespace
Trust Region Methods
& Employ trust-region frameworks to enhance stability and convergence when solving BOPs \citep{liu1998trust, marcotte2001trust, colson2005trust} \\
\addlinespace
Penalty Methods
& Introduce penalty terms for constraint violations, enabling approximate solutions to BOPs \citep{IsAi92, WhAn93, sinha2024gradient, kleinert2021computing} \\
\addlinespace
Approximation of Mappings
& Employ various approaches to approximate reaction set mappings, or optimal value functions, thereby reducing the cost of repeated nested optimization \citep{sinha2024gradient, islam2017surrogate} \\
\addlinespace
Hypergradient-based Methods
& Leverage hypergradients via implicit or automatic differentiation to optimize upper-level variables by accounting for the sensitivity of lower-level optimal solutions \citep{franceschi2018bilevel, liu2019darts} \\
\addlinespace
Sampling-based Nested Methods
& Use sampling to approximate nested structures in BOPs, reducing direct computational complexity \citep{mathieu, zhu2006hybrid, angelo2015study, finding2014sinha} \\
\addlinespace
Sampling-based Hybrid Methods
& Hybridize sampling with other methods (e.g., surrogate models, Bayesian inference) to balance accuracy and computational cost \citep{sinha2017evolutionary, islam2017surrogate, kieffer2017bayesian, sinha2020bilevel,sinha2021solving} \\
\bottomrule
\end{tabular}
\end{table} 

Originating in the early 1970s, bilevel optimization has grown into a significant research domain with a robust theoretical foundation and diverse algorithmic advancements \cite{sinha2017review,bard2013practical, dempe2002foundations}. Various techniques have been developed to tackle the hierarchical nature of these problems. Among them, reformulations using the KKT conditions are prominent for transforming the lower-level problem into a set of constraints for the upper-level. Gradient-based strategies offer an alternative, leveraging sensitivity analysis to iteratively optimize the problem. Trust-region methods, known for their robustness, create a local approximation of the bilevel problem using an auxiliary problem. Penalty-based formulations introduce regularization terms that penalize constraint violations, simplifying problem structures. Approximation of reaction set mappings, or optimal value functions for bilevel problems has been a popular technique to solve these problems. Furthermore, the rise of sampling-based approaches has led to nested approximation techniques and hybrid methods that combine surrogate modeling or Bayesian optimization to mitigate computational complexity. A summary of bilevel optimization methods is given in the Table~\ref{tab:bilevel_methods} for reference.

The present study extends this line of research by employing a bilevel optimization perspective to analyze the structure underlying continuous hyperparameter tuning and architecture search tasks. The next section on NAS demonstrates how bilevel optimization principles are applied in practice. In particular, the subsection on bilevel theory-based NAS methods explores specific solution strategies designed to address the bilevel nature of the problem.

\section{Neural Architecture Search}\label{sec:neural_architecture_optimization}
To gain a clearer understanding of how neural networks are designed and optimized, it is important to distinguish between model parameters and hyperparameters, as both govern different aspects of the learning process. Table~\ref{tab:params} provides a concise comparison between these two categories, highlighting their respective roles and implications in network training and performance.
\begin{table}[ht]
\begingroup
\centering
\caption{Comparison between Model Parameters and Hyperparameters}
\label{tab:params}
\renewcommand{\arraystretch}{1.3}
\begin{tabularx}{\textwidth}{>{\raggedright\arraybackslash}X >{\raggedright\arraybackslash}X}
\toprule
\textbf{Model Parameters} & \textbf{Hyperparameters} \\
\midrule
Weights and biases in ANNs, filters in CNNs, attention and feedforward weights in Transformers, hidden state weights in RNNs/LSTMs/GRUs, etc. & Number of layers and neurons, learning rate, regularization parameters, kernel size, number of filters (CNNs), dropout rate, batch size, etc. \\
\addlinespace
Learned from training data via backpropagation and gradient descent & Set by experts or found via hyperparameter tuning techniques (e.g., grid search, random search, Bayesian optimization, Evolutionary search, etc.) \\
\addlinespace
Updated during training to minimize loss function & Remain fixed during training but significantly affect the training dynamics and final performance \\
\addlinespace
Determine how the model fits the data & Define the model’s structure and learning process \\
\bottomrule
\end{tabularx}
\endgroup
\end{table}

As explained earlier, NAS is an automated approach for discovering optimal neural network architectures tailored to specific tasks and datasets. It reduces the need for expert intervention by exploring a predefined search space using methods such as Reinforcement Learning (RL), evolutionary algorithms, gradient-based optimization, etc. The process involves defining the search space, selecting an NAS strategy, sampling and evaluating architectures, and iteratively refining selections based on performance metrics like validation accuracy and model complexity.

\subsection{Bilevel Structure in NAS} 
We consider the following bilevel NAS formulation (Formulation \eqref{formulation_darts}) from \cite{liu2019darts}, which is widely adopted in practice.
\begin{equation}\label{formulation_darts}
	\begin{aligned}
		\min_{A} \quad & \mathcal{L}_v(A, \hat{W}) \\
		\text{s.t.} \quad & \hat{W}= \argmin_{W \in \mathscr{W}} \mathcal{L}_t(A, W) \\
		& A \in \mathscr{A}
	\end{aligned}
\end{equation}
In practical machine learning settings, a single model is typically instantiated for a given architecture $A$ and trained to optimality, yielding an optimal model $\hat{W}(A)$. Consequently, for all practical purposes, a unique global lower-level optimal solution is often assumed. However, from a theoretical perspective, multiple lower-level optimal solutions may exist for the same architecture, i.e., multiple optimal models can correspond to a given upper-level value $A$. This may occur because different initializations or choices of other hyperparameters (e.g., random seed, learning rate, weight decay, and dropout) can lead to distinct converged solutions with identical training performance. When multiple lower-level optimal solutions exist, an optimistic stance is implicitly adopted. Specifically, if multiple weight configurations attain the same optimal training loss, the lower-level optimizer is assumed to select the model that minimizes the upper-level objective, i.e., the one with the smallest validation loss. Therefore, Formulation~\eqref{formulation_sinha_and_gunwal} from~\cite{sinha2025linear} provides a more general representation of the bilevel NAS problem, as it correctly reflects that $\hat{W}$ denotes an optimistic selection from the set of all optimal models corresponding to a given architecture or hyperparameter configuration.
\begin{equation}\label{formulation_sinha_and_gunwal}
	\begin{aligned}
		\min_{A} \quad & \mathcal{L}_v(A, \hat{W}) \\
		\text{s.t.} \quad & \hat{W} \in \argmin_{W \in \mathscr{W}} \mathcal{L}_t(A, W) \\
		& A \in \mathscr{A}
	\end{aligned}
\end{equation}
Referring to the two NAS formulations discussed above, in the NAS/HPO setting, both notations ($=$ and $\in$) are technically valid, as adopted in \cite{liu2019darts} and \cite{sinha2025linear}, respectively. In practice, since only one model is typically considered for a given architecture or hyperparameter configuration, the reaction set (i.e., the set of optimal models corresponding to that architecture or configuration) is effectively treated as a singleton, which justifies the use of the equality notation ($=$).

It is important to emphasize that this bilevel formulation is not merely a modeling convenience but is essential for the integrity and reliability of the NAS process. If architectures were evaluated using suboptimal or partially trained weights, the search could be significantly misled, favoring architectures that appear promising during training but fail to generalize. This highlights the deep interdependence between architectural design and weight optimization: an architecture's potential can only be meaningfully assessed when its weights have been sufficiently optimized. Unlike single-level approaches that collapse both training and validation into a unified optimization of training loss, bilevel methods explicitly account for generalization by optimizing the validation loss at the upper-level. Such single-level approaches are particularly prone to overfitting, often yielding architectures that perform well on the training distribution but poorly on unseen data. In contrast, bilevel optimization ensures that each candidate architecture is evaluated fairly and robustly, with its weights tuned specifically to its structure. Ultimately, the goal is not merely to minimize training error but to maximize the generalization capability of the discovered architectures.
\begin{figure*}
\centering
\begin{minipage}[t]{0.48\textwidth}
\centering
\includegraphics[width=\linewidth]{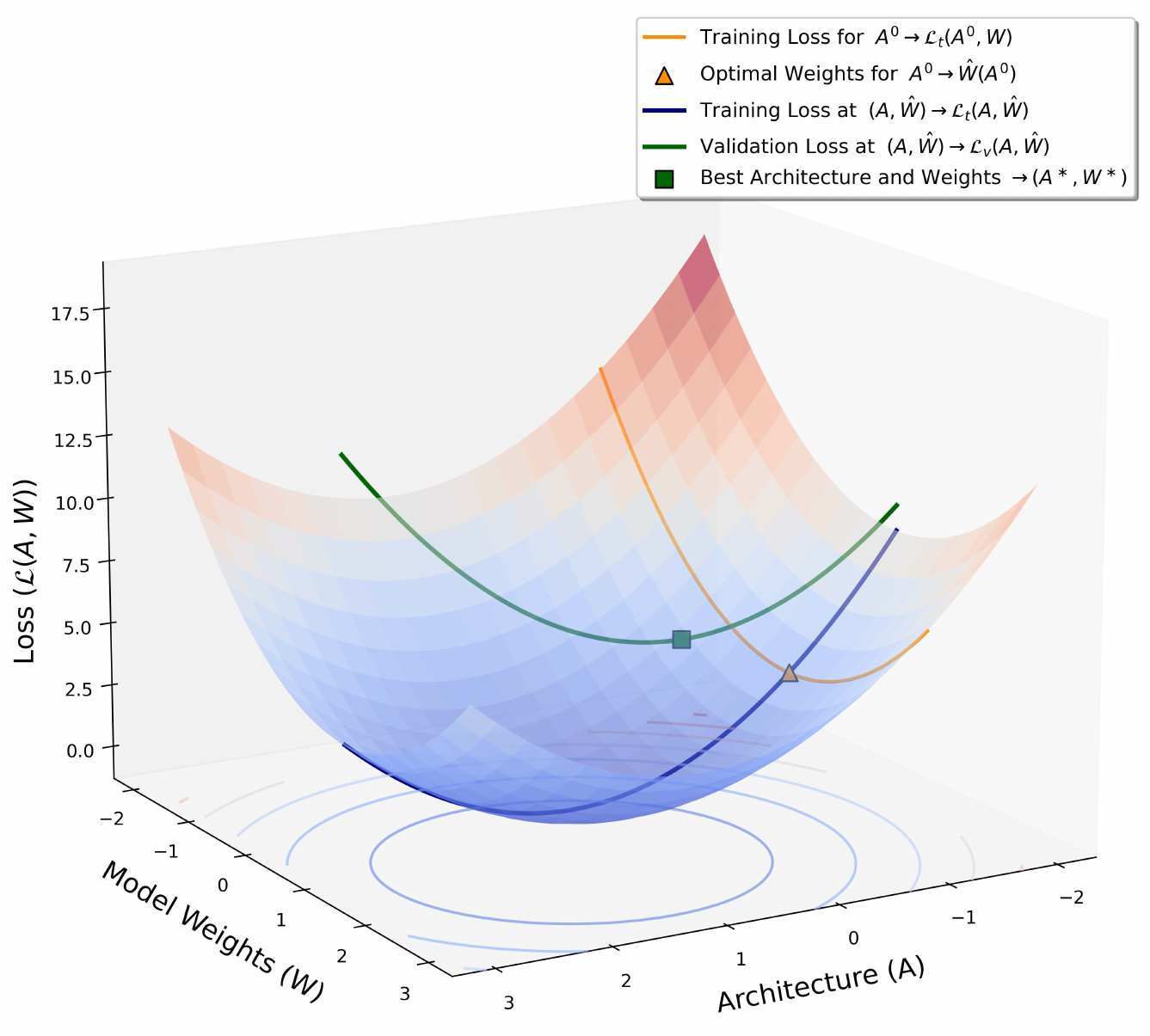}
\caption{Upper- and lower-level optimization in NAS.}
\label{loss_landscape}
\end{minipage}\hfill
\begin{minipage}[t]{0.48\textwidth}
\centering
\includegraphics[width=\linewidth]{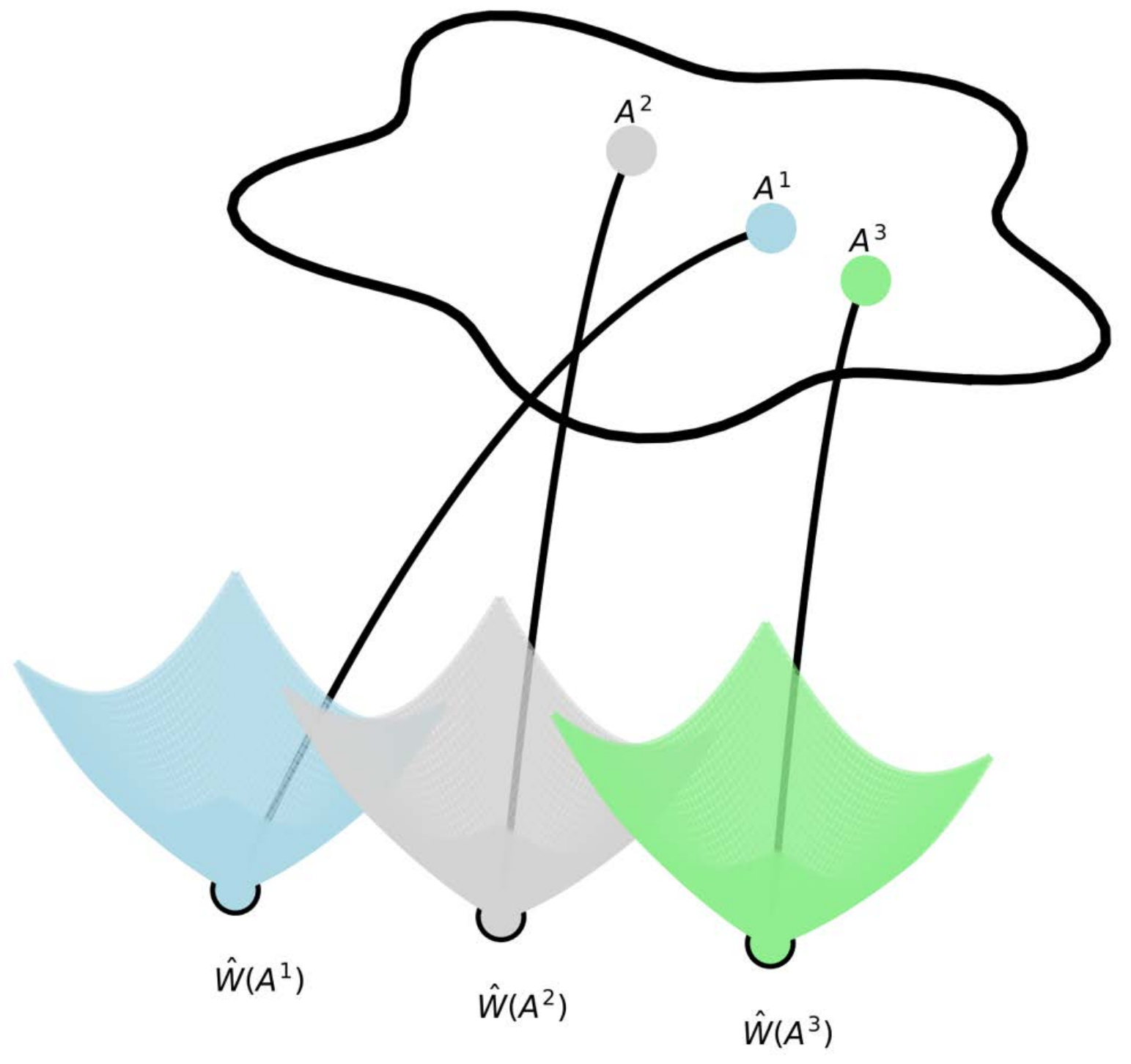}
\caption{Interplay between architecture selection and weight optimization.}
\label{fig:bilevel_nas}
\end{minipage}
\end{figure*}

A general assumption in NAS is that, for any given architecture parameter $A$, the lower-level problem admits at least one optimal solution $\hat{W}$. This assumption ensures that the bilevel problem has a well-defined, nonempty feasible region. The bilevel framework is particularly insightful in this regard, as it highlights two critical aspects: (1) the architecture space $\mathscr{A}$ directly shapes the lower-level optimization problem, and (2) only the optimal lower-level solutions, i.e., weight configurations that minimize the training loss, contribute to the upper-level objective defined by the validation loss. Consequently, the success of NAS fundamentally depends on the interaction between architecture selection and weight optimization, as illustrated in Figure~\ref{fig:bilevel_nas}. Notably, the lower-level problem is, in general, highly non-convex with respect to the lower-level variables and therefore constitutes a parametric non-convex optimization problem. However, for illustrative purposes, we assume the lower-level problem to be convex.

To build intuition for how bilevel optimization operates in NAS, we examine the loss landscape shown in Figure~\ref{loss_landscape}. The shaded surface depicts the overall loss function $\mathcal{L}(A, W)$, defined jointly over the architecture parameters $A$ and the model weights $W$. Each point on this surface corresponds to a particular pair $(A, W)$. If we fix an architecture $A^0$, the orange curve indicates how the training loss varies as the weights change. The orange triangle marks the optimal weights for this architecture, denoted by $\hat{W}(A^0) = \argmin_{W} \mathcal{L}_t(A^0, W)$. The corresponding training loss is then $\mathcal{L}_t(A^0,\hat{W}) = \mathcal{L}(A^0, \hat{W}; S^T)$, where $S^T$ denotes the training dataset. Repeating this process for all architectures yields the dark blue curve, which represents the optimized training loss $\mathcal{L}_t(A, \hat{W})$ after minimizing over the weights. The upper-level optimization is performed using the validation dataset $S^V$. The green curve in the figure represents the validation loss $\mathcal{L}_v(A, \hat{W}) = \mathcal{L}(A, \hat{W}; S^V)$, which guides the search toward the most promising architecture. The final solution is indicated by the green square, corresponding to the best architecture--weights pair $(A^*, W^*)$, where $A^* = \argmin_A \, \mathcal{L}_v(A,\hat{W})$. This marks the outcome of the bilevel optimization process.

\subsection{A Review of NAS}
NAS has rapidly evolved since the seminal work of Zoph and Le (2017) \citep{zoph2017neural}, which first demonstrated that automated search could match, and in some cases surpass, manually designed networks. Their formulation cast architecture search as an RL problem, where a controller network sampled candidate architectures and updated its policy using performance-driven feedback to enable more informed architecture sampling in subsequent iterations. This early paradigm, exemplified by NASNet \citep{zoph2018learning} and later ENAS \citep{pham2018efficient}, firmly established NAS as a promising alternative to manual architectural engineering. Although Zoph and Le ignited the field, their work built upon earlier efforts that explored architecture optimization in more restrictive forms. Approaches such as RL-based architecture search \citep{baker2016designing}, early evolutionary pipelines \citep{real2017large}, and Sequential Model-based Optimization (SMBO) driven modular frameworks like DeepArchitect \citep{negrinho2017deeparchitect} provided important precursors to modern NAS ideas. These foundations highlighted the need for automated design but lacked the scalability and generality that later methods achieved.

Shortly after the rise of RL-based NAS, Evolutionary Computation (EC) emerged as another major search strategy. Methods such as Regularized Evolution \citep{real2019regularized} explored populations of architectures through mutation and selection, offering a simple yet surprisingly effective alternative to RL-based controllers. These non-differentiable approaches, however, were computationally expensive, motivating the community to seek more efficient formulations. A major turning point arrived with the introduction of differentiable NAS. DARTS \citep{liu2019darts} proposed a continuous relaxation of the architecture space, enabling efficient gradient-based optimization of architectural parameters. Its influence was profound: follow-up methods such as PC-DARTS \citep{xu2019pcdarts}, ProxylessNAS \citep{cai2018proxylessnas}, and FBNet \citep{wu2019fbnet} improved scalability, addressed memory constraints, and introduced hardware-awareness for deployment on target mobile devices. The influence of DARTS has inspired a wide array of improved variants addressing stability, generalization, and performance issues, including \citep{chen2019progressive, zela2019understanding, fair-darts, liang2019darts+, qin2024darts, han2023ndarts, cai2023bhe, hasan2021darts, pang2021rl, cai2024sto, zhang2023differentiable, zhu2024relax, li2024lmd, de-darts, att-darts, yang2024ostr, guo2024semantic}. This differentiable paradigm subsequently expanded to transformers through AutoFormer \citep{chen2021autoformer}, and to platform-specific architectures such as MnasNet \citep{tan2019mnasnet}, which performs latency‑aware NAS on mobile devices, and EfficientNet \citep{tan2019efficientnet}, which uses multi-objective NAS to obtain a FLOPs-efficient baseline before compound scaling.

As NAS grew in popularity, the community recognized that reproducibility and fair comparison required standardized, well-defined search spaces. This led to the introduction of the NAS-Bench family, beginning with NAS-Bench-101 \citep{ying2019bench}, followed by NAS-Bench-201 \citep{dong2020bench}, NAS-Bench-301 \citep{siems2020bench}, NAS-Bench-NLP \citep{klyuchnikov2022bench}, and NAS-Bench-Graph \citep{qin2022bench}. These benchmarks provided fixed, exhaustively evaluated or surrogate-modeled search spaces, enabling rigorous evaluation without the confounding effects of noisy training pipelines. By clearly decoupling how architectures are searched (the NAS method) from where they are searched (the search space), these benchmarks played a pivotal role in stabilizing empirical results and making NAS research more accessible. Parallel to the emergence of benchmarks, flexible and task-adaptive search spaces grew in importance. Once-for-All networks \citep{cai2019once} introduced elastic supernets that encode a vast collection of subnetworks, enabling fast specialization through weight sharing. In the graph domain, GraphNAS \citep{gao2019graphnas} proposed a method for generating graph neural architectures, while NAS-Bench-Graph \citep{qin2022bench} provided a structured search space for evaluating them. Furthermore, recent efforts \citep{calhas2021automatic, perrin2024towards} have focused on automatically generating search spaces themselves, reducing reliance on hand-crafted motif design and expanding NAS to increasingly diverse domains.

A significant development in recent years has been the emergence of zero-cost proxies, such as SynFlow \cite{tanaka2020pruning}, TE-NAS \cite{chen2021neural}, and NASWOT \cite{mellor2021neural}, which estimate model quality using gradient-based or analytic signals computed at initialization. These methods dramatically reduce search costs and have redefined the efficiency frontier of NAS. In parallel, the field has moved beyond small convolutional cells to embrace large-scale language models \cite{lawton2023neural}, multimodal models \cite{sun2023rmnas}, and resource-efficient architectures \cite{xu2024survey}. This shift has motivated new paradigms such as LLM-guided NAS \cite{li2025collm, qin2024fl}, hybrid search strategies \cite{gao2025ranas, lee2024az}, and scalable supernet training \cite{yu2020train, hu2021improving} tailored for modern deep learning workloads.

Across this evolutionary trajectory, four broad families of NAS methods have crystallized: RL \citep{zoph2017neural}, EC \citep{liu2021survey}, Bayesian optimization \citep{gupta2024bayesian}, and gradient-based approaches \citep{ali2023gradient}. These are complemented by one-shot and supernet-based methods \citep{zhang2020one}, graph-based strategies \citep{oloulade2021graph}, hierarchical formulations \citep{liu2019auto}, and more recent zero-cost and hybrid approaches. In contrast, the search spaces explored by these methods have diversified from early cell-based designs \citep{dong2022cell}, such as the DARTS Search Space (DSS) \cite{liu2019darts}, to large-scale benchmarks and domain-specific collections in vision, language, and graph learning. Building on these richer search spaces and refined algorithms, NAS-designed architectures have consistently shown competitive or superior performance across diverse domains, including medical imaging \citep{xie2024lightweight, weng2019unet, wang2024mednas}, natural language processing \citep{sarah2024llama}, machine translation \citep{chitty2022neural}, smart city applications \citep{li2020autost}, and tasks like image classification, object detection \citep{zoph2017automl}, and semantic segmentation \citep{liu2019auto}. These advancements have democratized the network design process, reducing reliance on expert intervention and paving the way for broader accessibility in deep learning research \citep{meng2024evolution}.

Overall, the evolution of NAS reflects a shift from costly black-box exploration to principled optimization, standardized evaluation, and scalable, domain-aware search space design. With the foundational understanding established, we now turn our attention to two primary categories of neural architecture search methods: sampling-based approaches and bilevel theory-based approaches. This division reflects the conceptual framework adopted in this paper, wherein we organize the diverse NAS literature along these two axes. Each offers distinct strategies and theoretical underpinnings that have shaped the evolution of NAS.

\subsection{Sampling-based NAS Methods}
Sampling-based NAS methods explore the architecture search space by evaluating multiple candidate architectures through heuristic or probabilistic mechanisms. They are generally simpler to implement and often serve as baselines for more advanced techniques. By balancing exploration and exploitation, these methods can be effective, though they typically require a large number of architecture evaluations, making them computationally expensive. Importantly, several approaches in this category address the bilevel NAS problem by approximating the joint optimization of architecture parameters and model weights through carefully designed sampling procedures. The flexibility of the method allows to consider both discrete and continuous hyperparameters. Some of the prominent examples of this category are discussed next.

\subsubsection{Grid Search-based NAS}
\begin{figure*}
\centering
\begin{minipage}[t]{0.48\textwidth}
\centering
\includegraphics[width=\linewidth]{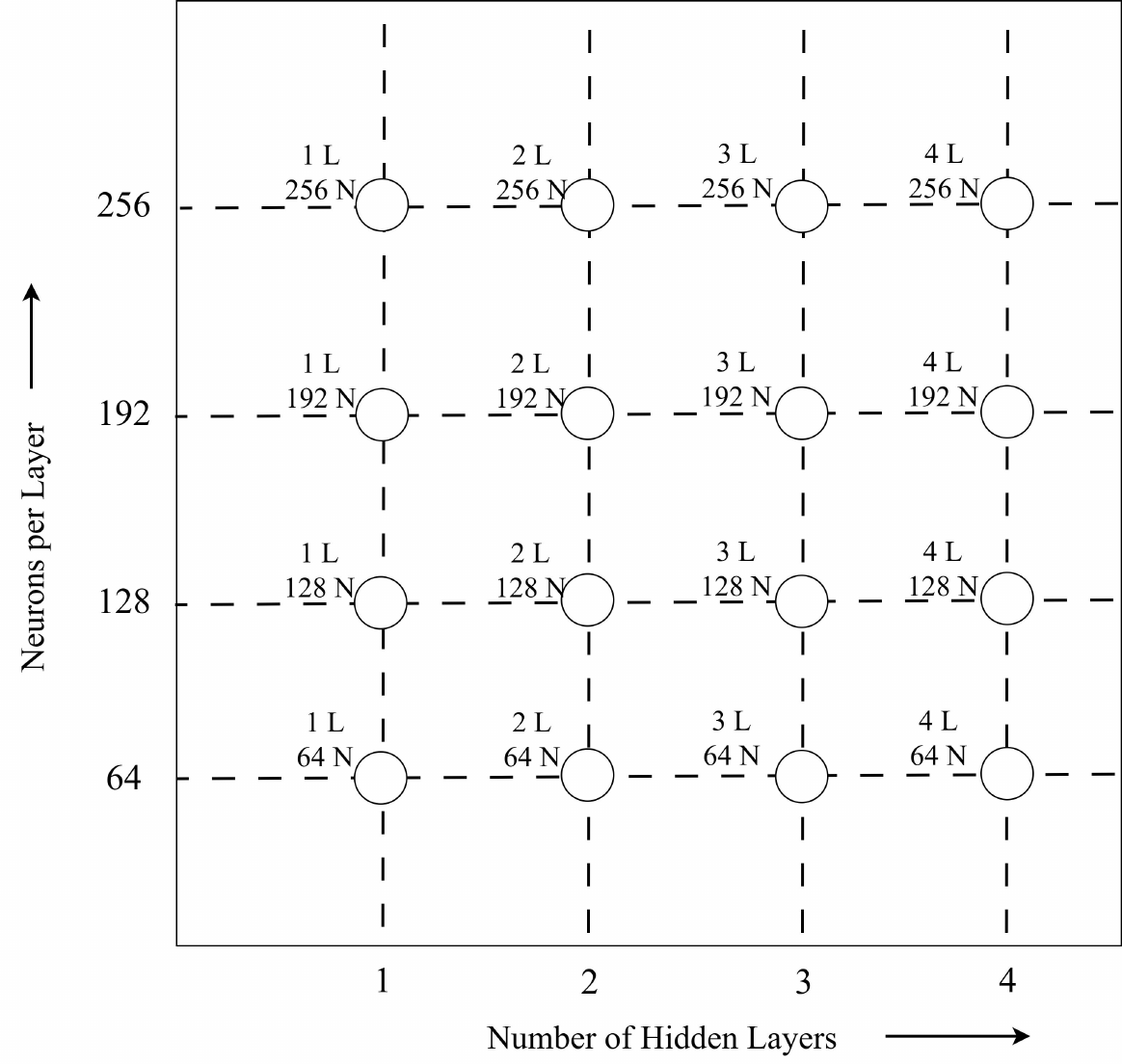}
\caption{Multilayer perceptron architecture search using grid search.}
\label{gs}
\end{minipage}\hfill
\begin{minipage}[t]{0.48\textwidth}
\centering
\includegraphics[width=\linewidth]{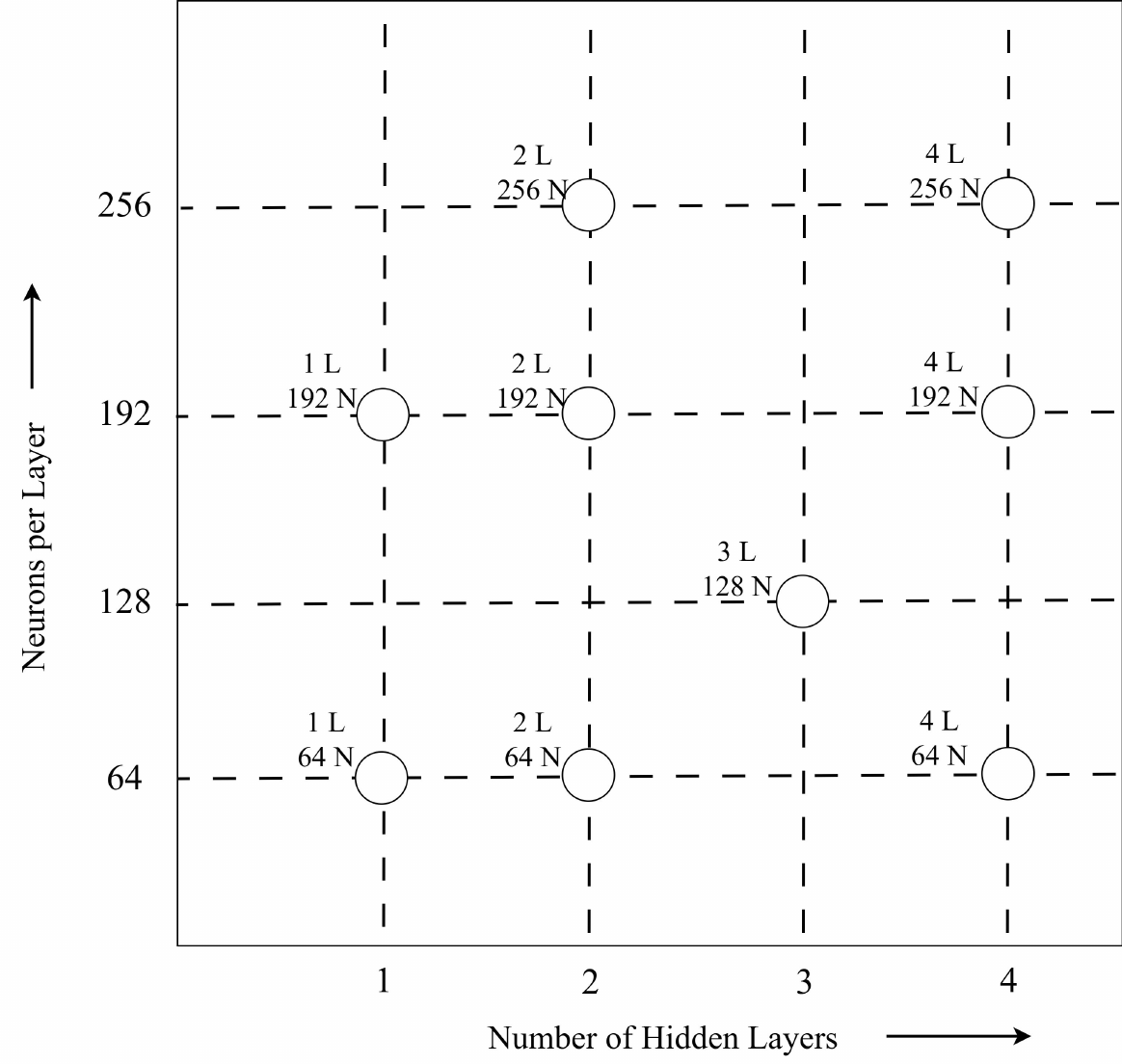}
\caption{Multilayer perceptron architecture search using random search.}
\label{rs}
\end{minipage}
\end{figure*}

The Grid Sampler, commonly referred to as grid search, represents one of the most fundamental and deterministic strategies in hyperparameter optimization as discussed in \citep{ogunsanya2023grid,dhilsath2021hyperparameter,ippolito2022hyperparameter,shekar2019grid}. Within the context of NAS, it functions by exhaustively evaluating all possible combinations of architectural and training-related hyperparameters, defined over a finite and discretized search space. This systematic exploration ensures that each configuration within the predefined grid is assessed, thereby offering a comprehensive approach to identifying potentially optimal neural architectures. Despite its conceptual simplicity and thoroughness, grid search becomes increasingly impractical as the dimensionality of the search space grows. NAS often involves high-dimensional and complex design spaces, incorporating choices such as layer types, connection schemes, kernel sizes, activation functions, and regularization strategies. In such cases, the exhaustive nature of grid search incurs a substantial computational burden, rendering it inefficient and often infeasible for large-scale or time-sensitive applications. Nevertheless, due to its deterministic behavior and ease of implementation, grid search remains a valuable baseline method, particularly in preliminary experiments or when exploring well-constrained architectural subsets. Figure~\ref{gs} provides an illustrative example of the grid search methodology as applied within the NAS framework.
\subsubsection{Random Search-based NAS} 
The Random Sampler, commonly referred to as random search, is one of the most fundamental and widely applicable strategies in NAS. It is particularly advantageous when the structure of the search space is unknown or difficult to model. In contrast to grid search, which systematically evaluates configurations drawn from a predefined grid, random search selects architecture configurations independently according to specified probability distributions, as discussed in \citep{bergstra2012random, bergstra2011algorithms}. This approach offers several practical advantages: it is simple to implement, inherently parallelizable, and robust to incomplete or redundant trials. Under a fixed computational budget, it often matches or even surpasses structured search methods, as model performance is typically sensitive to only a small subset of hyperparameters (low effective dimensionality). Consequently, grid search tends to waste evaluations on uninformative combinations, whereas random sampling distributes trials more effectively across influential regions of the hyperparameter space, rather than being constrained to a few discrete levels as in a grid, thereby achieving broader and more representative coverage~\cite{bergstra2012random}. Its ability to explore the architectural landscape in an unbiased and flexible manner makes it a strong baseline and a valuable tool for initializing more sophisticated search algorithms. Figure~\ref{rs} visually demonstrates the mechanism of random search within the NAS framework.

\subsubsection{Bayesian Optimization-based NAS}
\begin{figure}
	\centering
	\includegraphics[width=\linewidth]{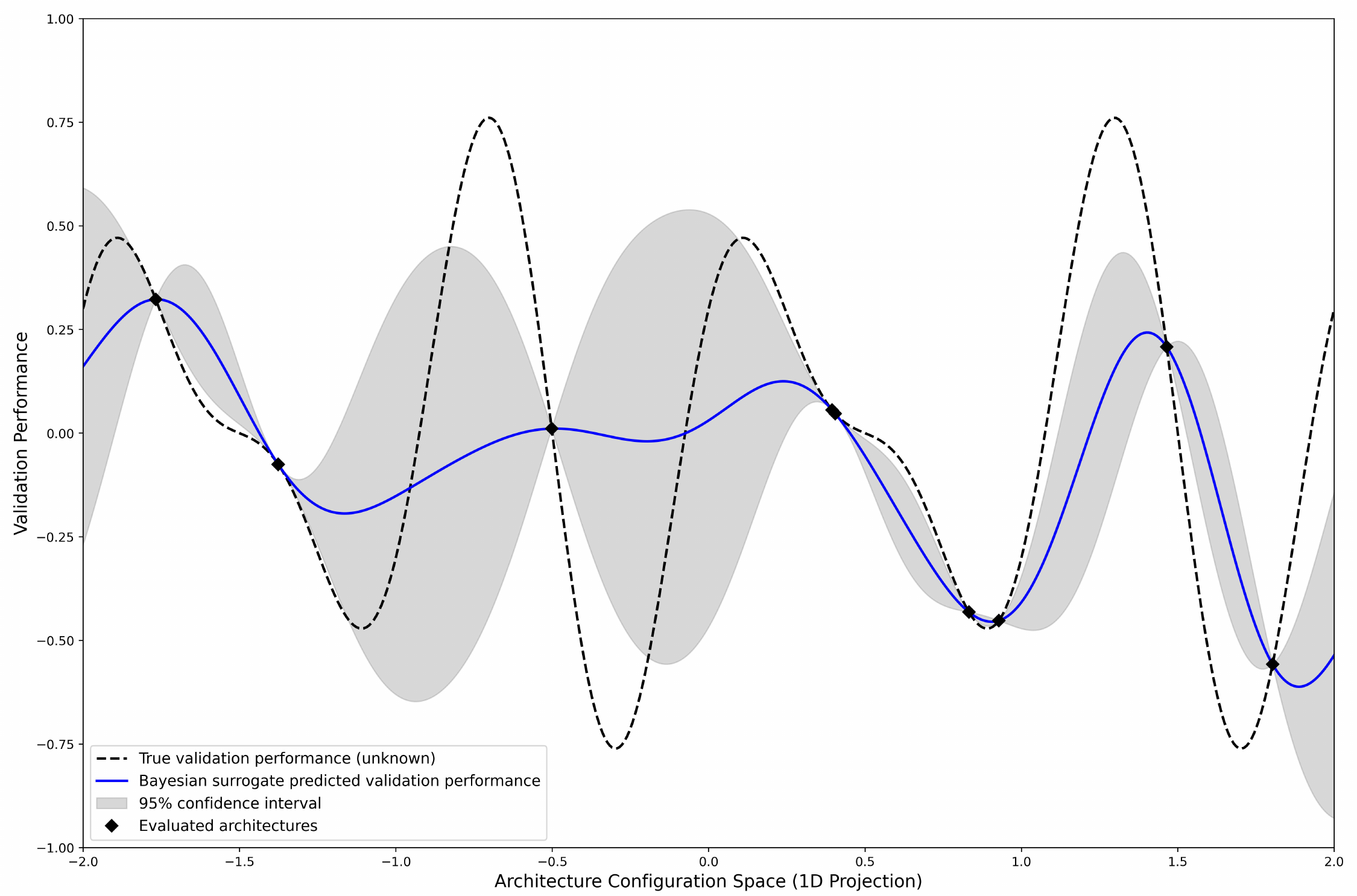}
	\caption{Bayesian optimization-based sampler in an NAS setting.}
	\label{bs}
    \end{figure}

Bayesian optimization introduces a more strategic and data-efficient alternative to conventional search methods such as grid and random search. This class of methods leverages probabilistic surrogate models to approximate the objective function, thereby guiding the search process towards promising regions of the architecture space as discussed in \citep{eggensperger2013towards, snoek2012practical, klein2017fast}. In the context of NAS, such surrogate models, commonly instantiated via Gaussian Processes or Tree-structured Parzen Estimators (TPE), enable the efficient exploration of complex, high-dimensional design spaces. The TPE, in particular, is a sophisticated variant of Bayesian optimization widely adopted for hyperparameter and architecture optimization tasks. Rather than evaluating configurations blindly or exhaustively, TPE constructs a probabilistic model that estimates the likelihood of achieving high performance for different architectural configurations. This model-driven approach allows TPE to balance the trade-off between exploration (searching new regions) and exploitation (refining known good regions) using an acquisition function that strategically selects the next candidates to evaluate. Compared to brute-force strategies like grid search and even the more flexible random search, TPE offers significantly improved efficiency by prioritizing evaluations that are statistically more likely to yield favorable outcomes. This targeted strategy makes Bayesian optimization, and TPE in particular, highly suitable for NAS problems, where evaluations are costly and the search space is vast and sparsely populated. Figure~\ref{bs} illustrates the functioning of a Bayesian optimization-based sampler within the NAS framework.

\subsubsection{Evolutionary Computation-based NAS}
\begin{figure}
	\centering
	\includegraphics[width=0.8\linewidth]{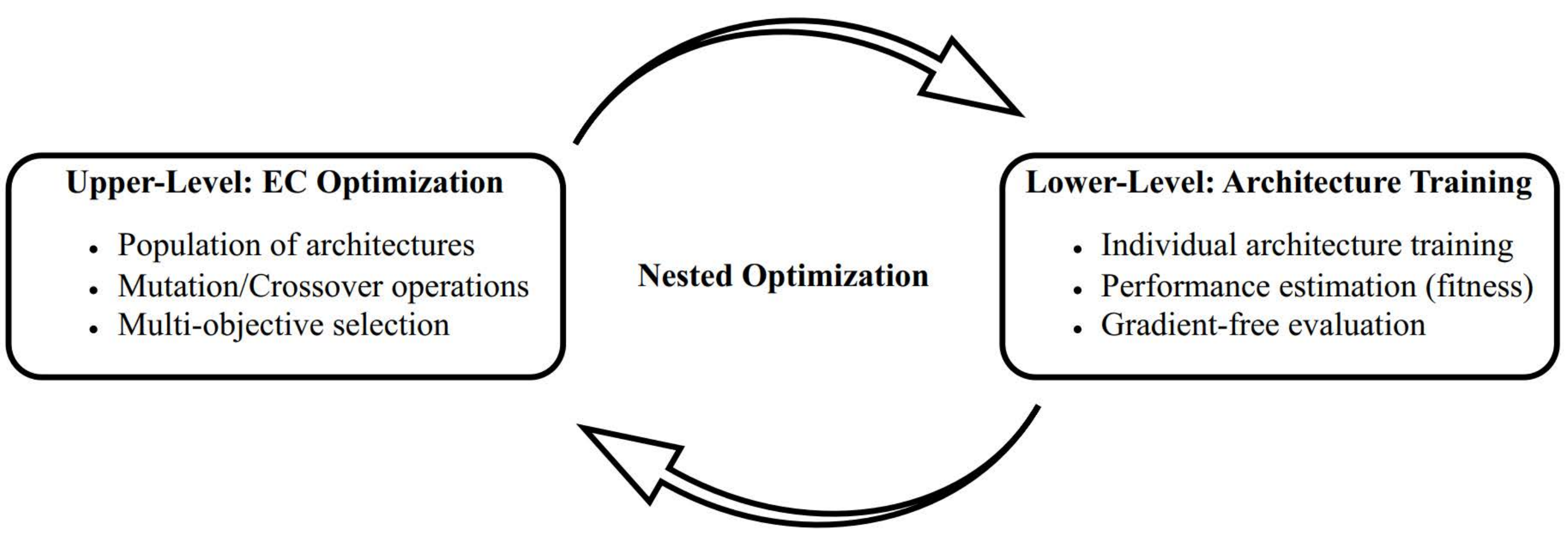}
	\caption{EC based samplers for NAS.}
	\label{ec}
\end{figure}

EC has emerged as a powerful approach to address the bilevel optimization challenges in NAS. At the upper-level, EC methods evolve a population of candidate architectures, while at the lower-level, each architecture is trained to obtain a performance estimate that serves as fitness. The iterative evolutionary cycle allows effective exploration of complex, non-convex, and discrete search spaces without relying on gradient information. Inspired by biological evolution, EC methods apply operators such as mutation and crossover to gradually refine architectures. For a high-level view of the evolutionary sampling process, please refer to Figure~\ref{ec}. Popular approaches include genetic algorithms and differential evolution, both of which have demonstrated success in discovering competitive architectures within large and irregular search spaces. A key advantage of EC lies in its flexibility to incorporate approximate evaluation strategies. Given the high computational cost of fully training each architecture, surrogate models are often employed to predict performance, or early generations may rely on partial training and low-fidelity evaluation. More accurate assessments are then reserved for later stages of evolution. Co-evolutionary strategies extend this paradigm by jointly evolving architectures and other hyperparameters, thereby capturing their mutual dependencies. Some of EC-based works are \citep{lorenzo2017particle, suganuma2017genetic, guo2020efficient, fan2022hybrid}. Furthermore, EC is naturally well-suited for multi-objective optimization, making it possible to balance competing criteria such as accuracy, latency, and model size. Several studies have explored multi-objective evolutionary approaches to neural architecture search (MOE-NAS), including \citep{lu2019nsga, elsken2018efficient, xue2023neural, lu2020nsganetv2, tong2022neural}. These methods extend standard evolutionary NAS frameworks by simultaneously optimizing multiple conflicting objectives, such as accuracy, computational cost, and model complexity, thereby producing a set of Pareto-optimal architectures.
\subsubsection{Reinforcement Learning-based NAS}
RL offers a powerful framework for neural architecture search by enabling an intelligent agent, typically implemented as a controller neural network, to learn a policy for sampling high-performing architectures through trial and feedback. In this setup, the controller (often an RNN or Transformer) generates candidate architectures sequentially, treating the process as a Markov Decision Process (MDP), where each architectural decision corresponds to an action within a discrete action space.
\begin{figure}[ht]
	\centering
	\includegraphics[width=0.78\linewidth]{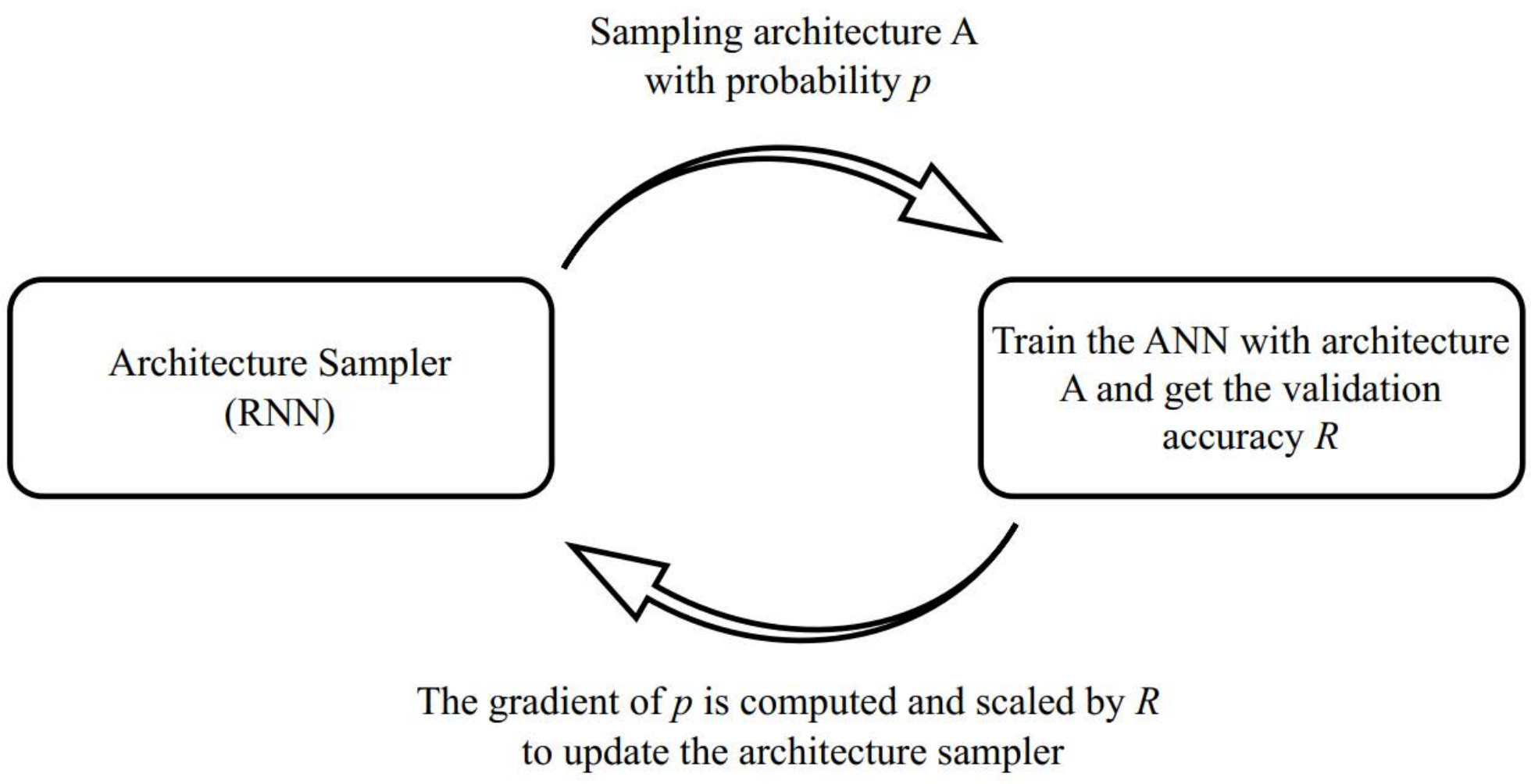}
	\caption{RL based sampler for NAS.}
	\label{rl}
\end{figure}

To address the NAS problem defined by Formulation~\eqref{formulation_darts}, an architecture sampler based on an RNN controller is employed to handle the upper-level task. At each iteration, the controller samples an architecture $A$ with probability $P_{\theta_c}(A)$, and this architecture is instantiated and trained on the training dataset to solve the lower-level problem using standard gradient-based methods. The performance of the resulting model, typically evaluated on a held-out validation set, serves as a reward signal $R(A)$ for the controller. The controller parameters $\theta_c$ are updated using policy gradient techniques to maximize the expected reward, thereby encouraging the generation of architectures that yield superior validation performance. Over successive iterations, the controller refines its sampling distribution, progressively favoring design choices that contribute to higher model accuracy. This adaptive exploration-exploitation mechanism enables RL-based NAS to effectively navigate vast architecture spaces and discover competitive network structures with minimal manual intervention. Figure~\ref{rl} provides a high-level overview of the RL process applied to NAS \citep{zoph2017neural}. Some other works in RL-based NAS are \citep{chu2020multi, mohan2023autorl, hsu2018monas, gao2019graphnas}.

\subsection{Bilevel Theory-based NAS Methods}
Bilevel theory-based NAS methods formulate NAS as a nested optimization problem, where architecture parameters serve as upper-level variables and model parameters as lower-level variables. The optimization of architecture parameters is guided by the performance of the trained model on a validation set, while the model parameters are optimized using the training data. These methods leverage bilevel optimization theory to provide a mathematical framework for solving the bilevel NAS problem, ensuring a theoretically grounded approach to jointly optimizing both sets of parameters. Bilevel NAS methods are particularly appealing for their theoretical rigor and enable efficient gradient-based optimization, though they often require careful handling of optimization dynamics and approximations. 
Most of these methods show benefit on continuous hyperparameters but can also be extended to discrete hyperparameters by doing a continuous relaxation. There is a tremendous potential of research in the context of discrete hyperparameters. Some of the influential methods in this category are discussed below.

\subsubsection{KKT Reformulation-based NAS}
These methods address the bilevel nature of NAS by reformulating it as a single-level optimization problem. Inspired by the approach in~\cite{mehra2021penalty}, this is achieved by replacing the lower-level problem with its corresponding KKT optimality conditions.\footnote{%
The KKT-based reformulation is presented only for illustrative purposes and is not intended to be an exact KKT reformulation of the bilevel NAS problem. Since only the first-order lower-level optimality condition is used, the resulting formulation should be interpreted as an approximation rather than a full KKT reformulation.} Consider the following tractable standard optimistic bilevel formulation of the NAS problem.
\begin{equation}\label{nas_2}
\begin{aligned}
\min_{A, \hat{W}} \quad & \mathcal{L}_v(A, \hat{W}) \\
\text{subject to:} \quad & \hat{W} \in \argmin_{W \in \mathscr{W}} \mathcal{L}_t(A, W), \\
& A \in \mathscr{A}
\end{aligned}
\end{equation}

This formulation is equivalent to the real optimistic formulation provided that the infimum of the validation loss over the reaction set of optimal models is attained (i.e., the minimum is achieved). Assuming the lower-level problem is unconstrained and sufficiently regular (e.g., \(\mathcal{L}_t\) is continuously differentiable), the first-order optimality condition is
\begin{equation*}
\nabla_W \mathcal{L}_t(A, \hat{W}) = 0,
\end{equation*}
which can be used to replace the lower-level problem, yielding the following single-level constrained optimization problem
\begin{equation*}
\min_{A, \hat{W}} \quad \mathcal{L}_v(A, \hat{W}) 
\quad \text{subject to} \quad 
\nabla_W \mathcal{L}_t(A, \hat{W}) = 0, \quad A \in \mathscr{A}
\end{equation*}

Solving this directly is difficult due to the nonconvexity introduced by the stationarity condition. To mitigate these issues, a common strategy is to relax the KKT condition using an \(\epsilon\)-approximation
\begin{equation*}
\|\nabla_W \mathcal{L}_t(A, \hat{W})\|^2 \leq \epsilon,
\end{equation*}
where \(\epsilon > 0\) controls the approximation accuracy. Alternatively, a penalty-based formulation is used
\begin{equation*}
\min_{A, \hat{W}} \quad \mathcal{L}_v(A, \hat{W}) + r \cdot \|\nabla_W \mathcal{L}_t(A, \hat{W})\|^2,
\quad A \in \mathscr{A},
\end{equation*}
where \(r > 0\) is a penalty parameter.

These relaxations convert the problem into an unconstrained and differentiable single-level formulation, enabling the use of gradient-based optimization methods. Gradients can be computed via implicit differentiation or unrolled optimization, and model and architecture parameters are updated either alternately or jointly using standard iterative methods such as stochastic gradient descent.

In summary, KKT-based reformulations provide a practical framework for approximating bilevel NAS problems as single-level problems. While effective and compatible with gradient-based solvers, they remain challenging due to nonconvexity, scalability, and approximation accuracy in high-dimensional search spaces.

\subsubsection{Surrogate Approximation-based NAS}
This class of methods utilizes surrogate models to alleviate the heavy computational demands of bilevel optimization in NAS. Instead of performing repeated and costly evaluations of candidate architectures or solving nested optimization problems from scratch, surrogate models are employed to approximate critical elements of the bilevel structure. These approximations serve to significantly speed up the search process while maintaining competitive solution quality. Surrogates can target different aspects of the optimization problem ranging from the upper-level objective and constraint sets to the lower-level solution mappings and optimal value functions \citep{sinha2024gradient}. By intelligently integrating these models, surrogate-based NAS approaches enable more efficient and scalable navigation of the architecture design space.

\begin{itemize}

\item \tb{Surrogate for the Upper-Level Problem}
\begin{equation*}
\min_{A, \hat{W}}~\tilde{\mathcal{L}}_v(A, \hat{W}) 
\quad \text{subject to} \quad
\begin{cases}
\hat{W} \in \argmin_{W \in \mathscr{W}} \mathcal{L}_t(A, W), \\
A \in {\mathscr{A}}
\end{cases}
\end{equation*}
Here, $\tilde{\mathcal{L}}_v$ is a surrogate model that approximates the upper-level validation loss. This formulation reduces evaluation overhead by directly learning a cheaper-to-evaluate proxy for the upper-level objective. Such an approach is typically employed when the upper-level function evaluation is computationally expensive, as may occur in engineering applications involving finite element analysis.\\

\item \tb{Surrogate for the Lower-Level Optimal Value Function ($\varphi$-mapping)}
\begin{equation*}
\min_{A, W}~\mathcal{L}_v(A, W) 
\quad \text{subject to} \quad
\begin{cases}
\mathcal{L}_t(A, W) \leq \tilde{\varphi}(A), \\
A \in \mathscr{A}
\end{cases}
\end{equation*}
In this formulation, $\tilde{\varphi}(A)$ approximates the optimal value function
\[
\varphi(A) := \inf_{W \in \mathscr{W}} \mathcal{L}_t(A, W)
\]
By replacing the inner optimization with a learned upper bound on the optimal value, this surrogate constrains feasible weights without explicitly solving the lower-level problem.\\

\item \tb{Surrogate for the Reaction Set Mapping ($\Psi$-mapping)}
\begin{equation*}
\min_{A, \hat{W}}~\mathcal{L}_v(A, \hat{W}) 
\quad \text{subject to} \quad
\begin{cases}
\hat{W} \in \tilde{\Psi}(A), \\
A \in \mathscr{A}
\end{cases}
\end{equation*}
Here, $\tilde{\Psi}(A)$ represents a surrogate set-valued mapping that approximates the reaction set
\[
\Psi(A) := \argmin_{W \in \mathscr{W}} \mathcal{L}_t(A, W)
\]
This relaxation accounts for uncertainty or variability in lower-level solutions and allows for more flexible modeling of solution behavior. The study in~\cite{lorraine2018stochastic} demonstrates this approach in the context of HPO tasks.

\item \tb{Surrogate for the Choice Function ($\psi$-mapping)}
\begin{equation*}
\min_{A}~\mathcal{L}_v(A, \tilde{\psi}(A)) 
\quad \text{subject to} \quad
A \in \mathscr{A}
\end{equation*}
In this case, $\tilde{\psi}(A)$ is a surrogate model that directly approximates the choice function
\[
\psi(A) \in \argmin_{W \in \mathscr{W}} \mathcal{L}_t(A, W)
\]
This surrogate choice function returns a singleton that is a functional approximation of a lower-level optimizer. The $\Psi$-mapping and $\psi$-mapping approximation approaches differ only in the presence of multiple lower-level optimal solutions.\\

\end{itemize}

Most of these approaches effectively transform the bilevel NAS problem into a single-level constrained optimization problem, wherein the solution to the inner problem is approximated through surrogate models or additional constraints. Subsequently, penalty or augmented Lagrangian techniques are employed to solve the resulting formulation efficiently. For instance, consider the following surrogate-based reformulation (surrogate $\varphi$-mapping), which represents NAS as a single-level nonlinear constrained optimization problem and thereafter a penalty-based solution approach is utilized:
\begin{equation}\label{nas_3}
\begin{aligned}
\min_{A, W} \quad & \mathcal{L}_v(A, W) \\
\text{subject to:} \quad & \mathcal{L}_t(A, W) - \tilde{\varphi}(A) \leq 0
\end{aligned}
\end{equation}

The constraint in Formulation~\eqref{nas_3} enforces that the current weights $W$ yield a training loss that is close to the optimal achievable value, thereby approximating the lower-level solution of the original bilevel formulation. To solve this constrained problem, we employ the augmented Lagrangian method. Let
\[
P(A, W) := \mathcal{L}_t(A, W) - \tilde{\varphi}(A)
\]
denote the constraint residual, $R > 0$ be a penalty parameter, and $\mu$ the Lagrange multiplier. The resulting optimization problem is
\begin{equation*}
\min_{A, W} \; Z(A, W) = \mathcal{L}_v(A, W) + \frac{R}{2} \, P(A, W)^2 + \mu \, P(A, W)
\end{equation*}

This augmented objective guides the optimization toward architectures that not only minimize validation loss but also produce training losses close to their optimal values. The penalty and multiplier terms progressively enforce the training loss constraint during optimization. This approach, referred to as Penalized Validation Method (PVM), is discussed in  \citep{sinha2024gradient}. By utilizing surrogate modeling in conjunction with augmented Lagrangian optimization, PVM eliminates the need for repeated inner-level training and enables tractable and scalable architecture search. Another work based on penalty methods for HPO is \cite{shi2021improved}.

As discussed earlier, trust-region-based approaches provide a well-founded methodology for solving bilevel problems. These methods address the problem iteratively by solving a sequence of locally approximated subproblems, thereby giving rise to another surrogate approximation-based strategy. At each iteration, a trust region-a localized neighborhood around the current solution-is established, within which the original bilevel objective is approximated. This approximation typically employs first-order Taylor expansions for the upper-level objective and constraints, and second-order expansions for the lower-level objective, resulting in a linear–quadratic BOP that is more tractable to solve. The solution to each subproblem then suggests a candidate update for the upper-level variables. Crucially, the size of the trust-region is adjusted dynamically: it is expanded when the predicted improvement aligns well with the actual performance gain, and contracted otherwise. This adaptive behavior promotes both stability and efficiency in the optimization process, particularly in settings characterized by high sensitivity to upper-level variations. By preserving the bilevel structure across iterations and employing localized approximations, trust-region-based methods achieve reliable convergence and exhibit strong generalization capabilities. In the unrestricted setting, the study in~\cite{mackay2019self} proposes adapting regularization hyperparameters in neural networks by constructing compact approximations of the best-response mapping that relate hyperparameters to their corresponding optimal weights and biases. This approach bears conceptual similarity to trust-region methods~\cite{colson2005trust}, which iteratively solve locally simplified bilevel subproblems to approximate the original formulation. This connection highlights the suitability of trust-region principles for NAS, where localized surrogate approximations can effectively balance approximation quality and computational efficiency.

\subsubsection{Hypergradient-based NAS}
To address the bilevel NAS, these methods compute the gradient of the upper-level objective (validation loss) with respect to the upper-level variables (architecture parameters) while explicitly incorporating the optimality conditions of the lower-level problem (model optimality). The resulting derivative is commonly referred to as the hypergradient. These approaches account for the dependence of the optimal weights on the architecture. 

Hypergradient-based NAS methods relax the discrete architecture search space into a continuous domain, enabling the use of gradient-based optimization. 

The indirect dependence is incorporated by differentiating through the lower-level optimization, yielding more informed updates at the cost of higher computation. Thus, it accounts for the indirect dependence of validation loss on architecture through the optimal weights. This is expressed by the total derivative 
\[
\frac{d \mathcal{L}_v(A, \hat{W})}{d A} = 
\frac{\partial \mathcal{L}_v}{\partial A} + 
\frac{\partial \mathcal{L}_v}{\partial\hat{W}} \cdot \frac{d \hat{W}}{d A}
\]
Here, the first term reflects the direct effect of \( A \), while the second term captures how changes in \( A \) influence the optimal weights and hence the validation loss. 

By approximating these hypergradients, typically via implicit differentiation or reverse-mode automatic differentiation one can carry out architecture optimization in a theoretically grounded and computationally efficient manner.

In practice, different variants of gradient descent can be employed for the architecture updates using the hypergradients. Batch Gradient Descent (BGD) computes exact gradients over the entire dataset, yielding stable but computationally expensive updates. Stochastic Gradient Descent (SGD) uses a single random sample per update, making the process faster but noisier. Mini Batch Gradient Descent (MBGD) strikes a balance by using small subsets of data, offering both efficiency and stability. Extensions such as momentum, RMSProp, and Adam further accelerate convergence and smooth the optimization trajectory by incorporating historical gradient information and adaptive learning rates. Overall, hypergradient-based NAS provides an efficient mechanism to optimize architectures by leveraging differentiability, achieving competitive performance while incurring only a fraction of the computational expense required by black-box NAS methods such as based on RL or EC. \cite{maclaurin2015gradient} demonstrates the hypergradient-based hyperparameter optimization.

A prominent example of hypergradient-based NAS is the Differentiable Architecture Search (DARTS) algorithm, which leverages this approach to iteratively refine architectures using gradient-based updates. Numerous variants evolved from the original DARTS framework, each addressing specific limitations or enhancing performance. The foundational method, DARTS \cite{liu2019darts}, introduced a differentiable neural architecture search framework operating in a continuous search space, which significantly improved efficiency and competitiveness over traditional RL and EC based approaches. To address computational and stability issues, P-DARTS \cite{chen2019progressive} was proposed as a progressive search space approximation with regularization, enhancing the robustness of the search process. Similarly, R-DARTS \cite{zela2019understanding} focused on robustness by analyzing the architectural space and integrating regularization to mitigate performance collapse. PC-DARTS \cite{xu2019pcdarts} improved efficiency and stability by employing a partially-connected search space, thereby reducing memory usage without compromising performance. Att-DARTS \cite{att-darts} incorporated attention mechanisms to enhance accuracy, while DARTS+ \cite{liang2019darts+} introduced an early stopping criterion to combat overfitting and excessive use of skip connections. Fair DARTS \cite{fair-darts} tackled the issue of unfair competition among candidate operations that led to suboptimal selections. Further extending the paradigm, DE-DARTS \cite{de-darts} employed dynamic attention networks to enhance architecture exploration, and EG-DARTS \cite{zhang2023enhanced} combined gradient-based search with evolutionary strategies to improve performance while reducing architectural complexity. More recent contributions such as Relax-DARTS \cite{zhu2024relax} aimed to alleviate overfitting and enhance generalization. STO-DARTS \cite{cai2024sto} introduced mechanisms to escape local minima, while OSTR-DARTS \cite{yang2024ostr} addressed search process degeneration using a magnitude-based operation selection scheme. To embed semantic understanding, Semantic DARTS \cite{guo2024semantic} utilized Masked Image Modeling (MIM) and classification to improve generalizability. Targeting search efficiency, LMD-DARTS \cite{li2024lmd} used dynamic sampling to prune weak operations and reallocate weights, accelerating the search and reducing memory demands. Finally, HN-DARTS \cite{li2024hn} redefined the traditional search space by introducing the \texttt{glore\_unit} in place of standard cells, achieving compact yet accurate architectures. Collectively, these contributions represented a rich and evolving landscape of hypergradient-based NAS methods, each advancing the scalability, reliability, and effectiveness of differentiable architecture search methods.

\begin{itemize}
\item \textbf{Differentiable Architecture Search.} 
NAS, as discussed earlier, is concerned with the automated discovery of network architectures that achieve superior task performance. In CNNs, this process operates over a predefined search space consisting of candidate operations such as convolutions, pooling layers, and skip connections. The goal is to identify an optimal composition of these operations that maximizes validation performance under the assumption that each candidate architecture is trained to convergence. This is depicted in Figure~\ref{fig:nas_1}. Traditional NAS methods formulate the problem as a discrete optimization task, where each edge in the network must select one operation from a finite set and each node must determine its incoming connections. This results in a combinatorial search space that grows exponentially with network depth and width. For instance, even a simple MLP with 10 layers and between 1 to 10 neurons per layer admits $10^{10}$ possible configurations. Similarly, a small NAS search space may contain an order of $10^{18}$ candidate architectures \cite{liu2019darts}. Exploring such spaces using RL or evolutionary algorithms often requires thousands of GPU-days, rendering them impractical. This computational bottleneck has motivated the development of DARTS, which addresses the challenge by relaxing the discrete optimization into a continuous formulation, thereby enabling efficient gradient-based search.
\begin{figure*}
\centering
\begin{minipage}[t]{0.55\textwidth}
\centering
\includegraphics[width=\linewidth]{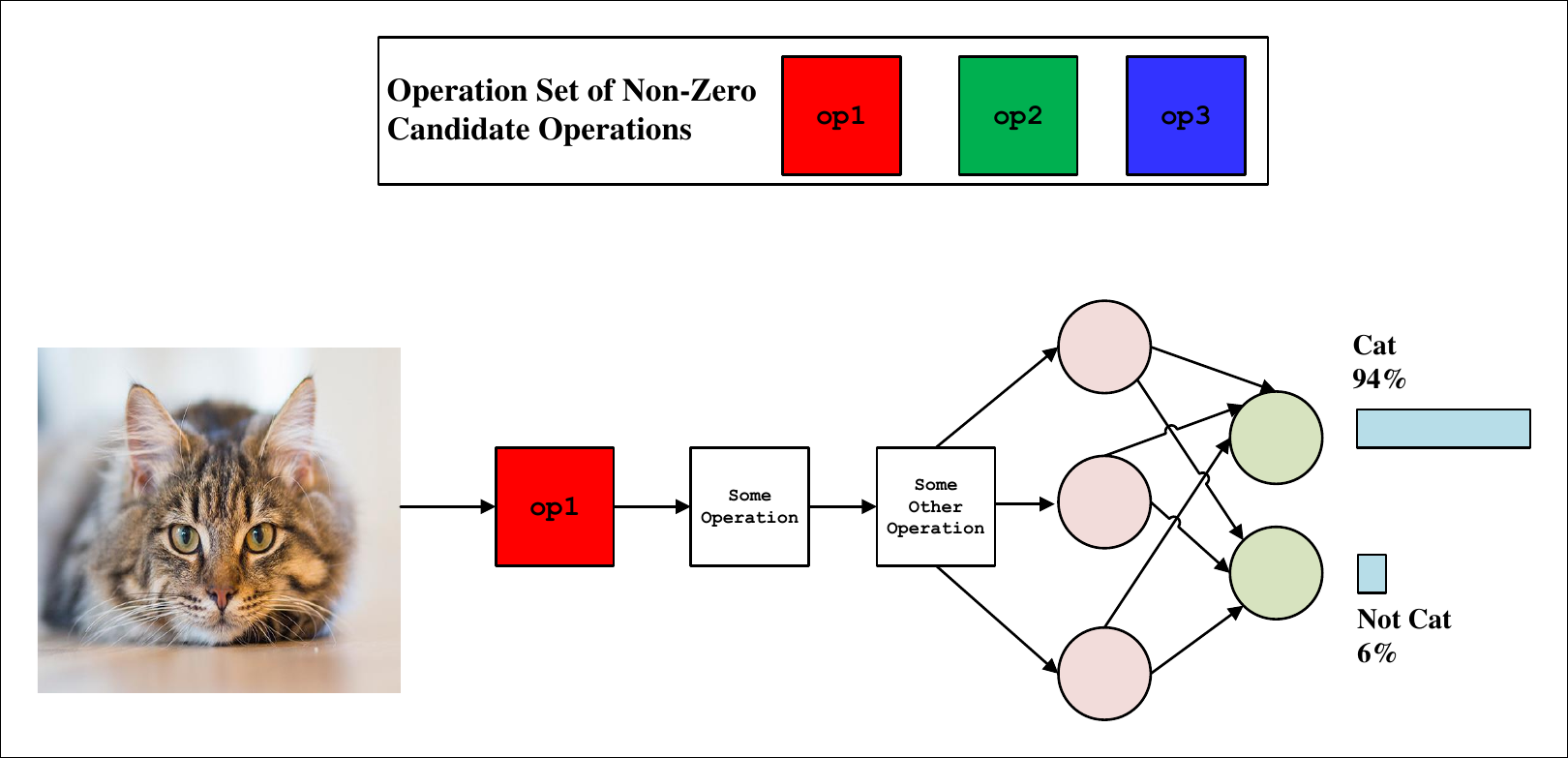}
\caption{Building task-specific networks by selecting operations from a search space.}
\label{fig:nas_1}
\end{minipage}\hfill
\begin{minipage}[t]{0.4125\textwidth}
\centering
\includegraphics[width=\linewidth]{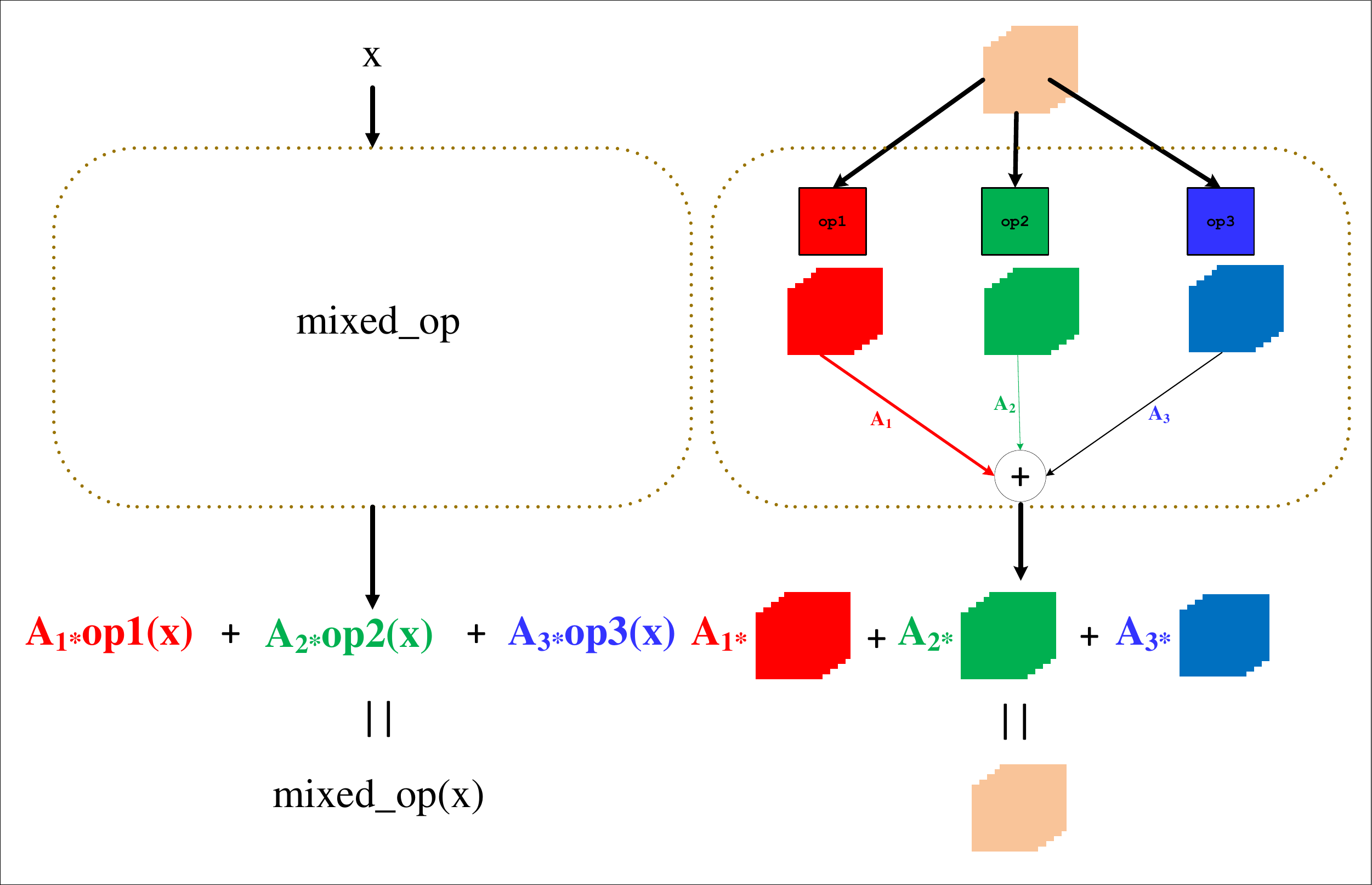}
\caption{Modeling edges as weighted sums of candidate operations.}
\label{fig:cont_relax}
\end{minipage}
\end{figure*}

\begin{description}
    \item[\tb{Continuous Relaxation:}] The core idea of differentiable NAS is to replace discrete operation with a continuous mixture of all candidate operations. Specifically, each edge $(i,j)$ in the Directed Acyclic Graph (DAG) is represented as a weighted sum of candidate operations, with weights determined by architecture parameters shared across the network. The relaxation is formalized as
\[
    \overline{o}^{(i, j)}(x) = \sum_{o \in \mathcal{O}} 
        \frac{\exp(A^{(i,j)}_o)}{\sum_{o' \in \mathcal{O}} \exp(A^{(i,j)}_{o'})}\; o(x),
\]
where $\mathcal{O}$ denotes the set of candidate operations and $A^{(i,j)}_o$ are learnable architecture parameters. The softmax ensures positive mixture coefficients summing to one. Figure~\ref{fig:cont_relax} illustrates this relaxation.\\

\item[\tb{Typical Cell Structure:}] To maintain tractability, most differentiable NAS methods restrict the search to a modular computational unit known as a cell, which is a DAG of nodes (feature maps) and edges (operations). Once an optimal cell is found, it is stacked repeatedly to construct the final architecture. A simplified cell structure is shown in Figure~\ref{fig:simple_cell}, where input feature maps are sequentially transformed through mixed operations to produce output feature maps. A typical cell, illustrated in Figure~\ref{fig:typical_cell}, is a DAG where nodes represent feature maps and edges denote candidate operations. Normal and reduction cells alternate to construct deeper architectures.

\begin{figure*}[htb]
\centering
\begin{minipage}[t]{0.47\textwidth}
\centering
\includegraphics[width=7cm, height=8.5cm]{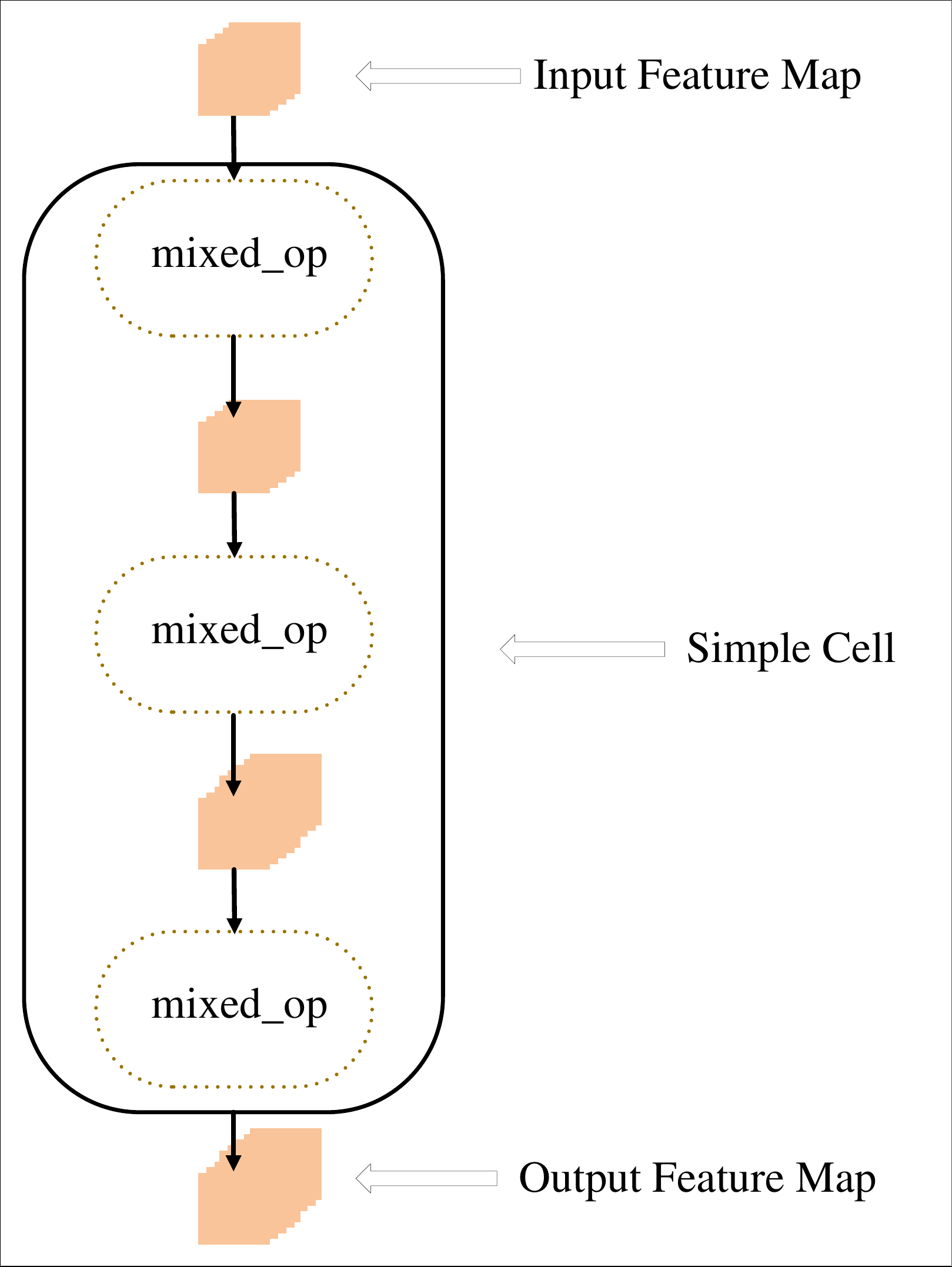}
\caption{A simple cell structure under continuous relaxation.}
\label{fig:simple_cell}
\end{minipage}\hfill
\begin{minipage}[t]{0.478\textwidth}
\centering
\includegraphics[width=7cm, height=8.5cm]{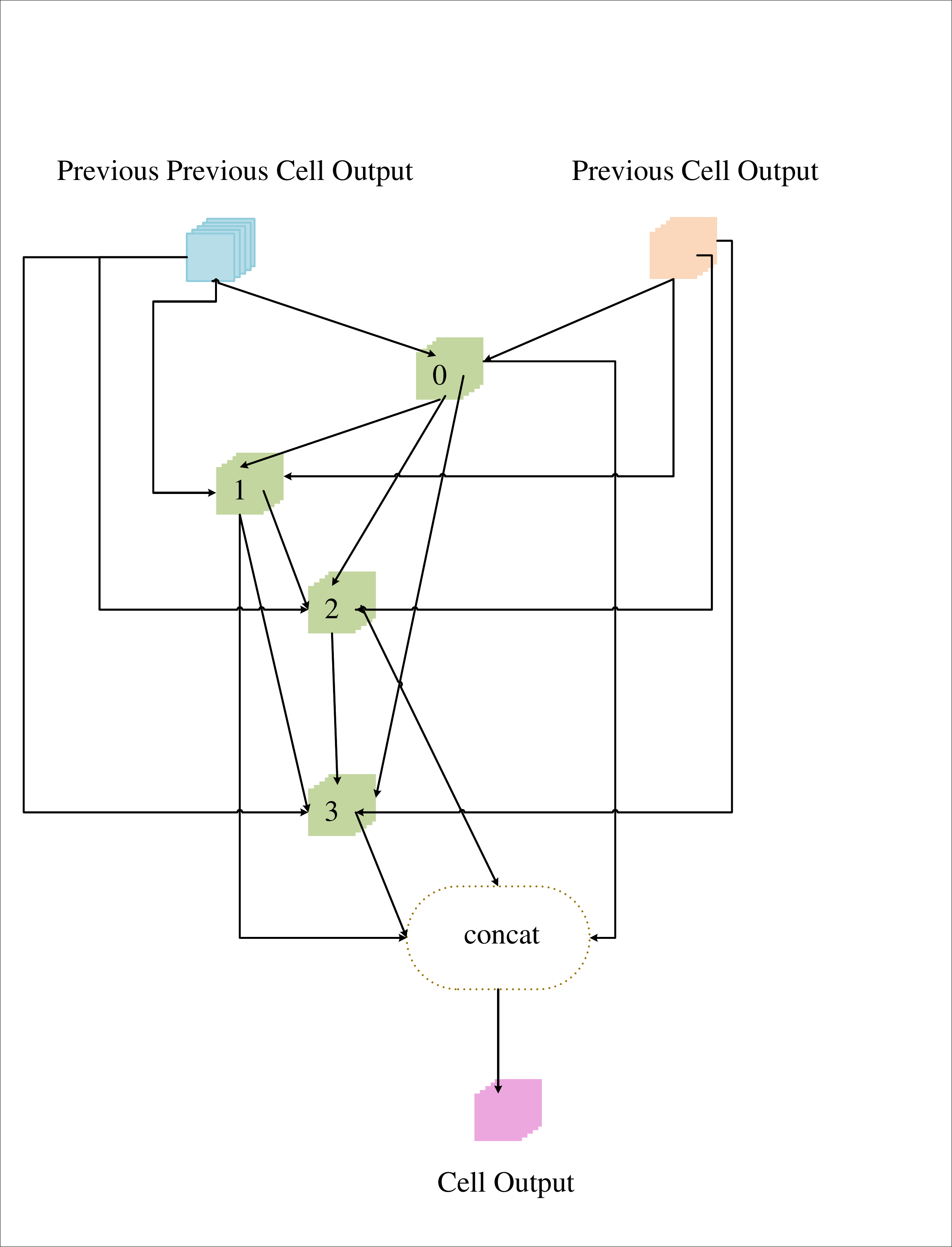}
\caption{A typical cell as a DAG.}
\label{fig:typical_cell}
\end{minipage}
\end{figure*}

\item[\tb{Optimization Procedure:}] DARTS \cite{liu2019darts} formulates NAS as a continuous BOP. Each cell is modeled as a DAG with two input nodes, four intermediate nodes, and one output node obtained by concatenating selected intermediates. The set of available operations, $\mathcal{O}$, comprises identity mappings (skip connections), separable convolutions with kernel sizes of $3 \times 3$ and $5 \times 5$, dilated separable convolutions with kernel sizes of $3 \times 3$ and $5 \times 5$, max pooling and average pooling with a $3 \times 3$ kernel, as well as a zero operation indicating no connection. Two distinct cell types are considered: normal cells, which preserve spatial resolution, and reduction cells, which downsample spatial dimensions. These cells are alternated in a stacking strategy to form the full architecture, as shown in Figure~\ref{nn_arch}, balancing feature extraction and hierarchical abstraction.
\begin{figure}[htb]
	\centering
	\includegraphics[width=\linewidth]{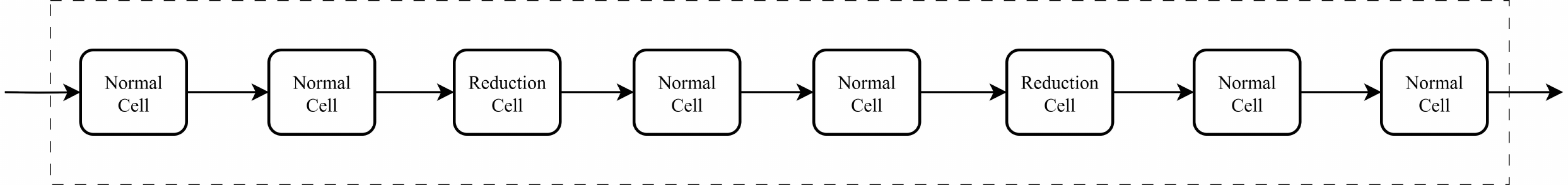}
	\caption{A cell-based neural network architecture.}
	\label{nn_arch}
\end{figure}
The feature representation at any node $j$, denoted as $x^{(j)}$, is obtained by aggregating the outputs from all its predecessor nodes. Formally, it is expressed as
\[
    x^{(j)} = \sum_{i < j} o^{(i,j)}\big(x^{(i)}\big),
\]
where $o^{(i,j)}$ represents the operation applied on the input $x^{(i)}$ coming from node $i$ to node $j$. The choice of which operation to use on each edge $(i,j)$ is determined through the DARTS optimization framework. In this framework, each edge is associated with a set of candidate operations $\mathcal{O}$, and a corresponding weight vector $A^{(i,j)} \in \mathbb{R}^{|\mathcal{O}|}$, where $|\mathcal{O}|$ is the number of candidate operations. The vector $A^{(i,j)}$ defines how much importance (or probability) is assigned to each candidate operation on edge $(i,j)$. After optimization, the final architecture is obtained by selecting, for each edge, the operation with the highest weight
\[
    o^{*(i,j)} = \argmax_{o \in \mathcal{O}} A_{o}^{*(i,j)},
\]
where $A^{*(i,j)}$ denotes the vector of optimized weights on edge $(i,j)$. The DARTS optimization procedure alternates between updating the architecture parameters \( A \), which minimize the validation loss, and the model weights \( W \), which minimize the training loss. Formally, to update the architecture, the gradient of the validation loss with respect to \( A \) is approximated as
\[
\nabla_A \mathcal{L}_v(A, \hat{W}) \approx \nabla_A \mathcal{L}_v\left(A, W - \xi \nabla_W \mathcal{L}_t(A, W)\right),
\]
where $\xi$ is a small learning rate for one-step unrolled optimization of the model training problem. Setting $\xi = 0$ yields a first-order approximation, while $\xi \neq 0$ introduces a second-order correction
\begin{equation}\label{approx_hypergrad}
\nabla_A \mathcal{L}_v(A,\hat{W}) \approx \nabla_A \mathcal{L}_v(A, W') - \xi \nabla^2_{A, W} \mathcal{L}_t(A, W) \nabla_{W'} \mathcal{L}_v(A, W'),
\end{equation}
with $W' = W - \xi \nabla_W \mathcal{L}_t(A, W)$. The second-order term captures how changes in the architecture affect the optimal weights and can be efficiently approximated via finite differences. For a detailed discussion of DARTS, readers may refer to \cite{liu2019darts}.
\end{description} 
We next discuss another bilevel theoretic NAS method based on auxiliary mathematical programming framework for differentiable neural architecture search that enables precise, localized updates by jointly optimizing over both the architecture and model parameter spaces. 
\end{itemize}

\subsubsection{Auxiliary Program-based NAS}\label{sec:proposed_approach}
\begin{figure}
    \centering
    \includegraphics[width=0.75\linewidth]{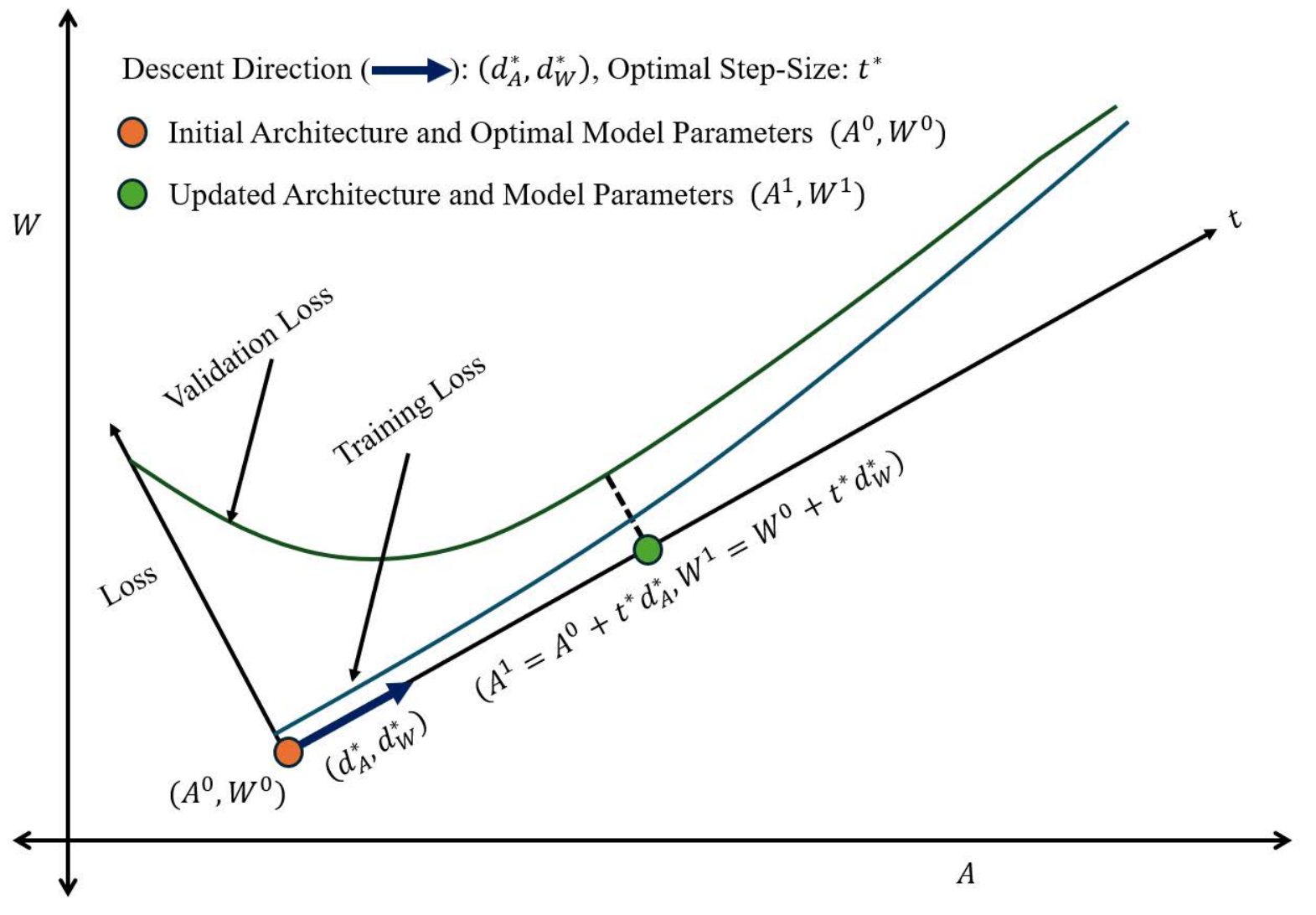}
    \caption{Determination of the optimal step-size \((t^*)\).}
    \label{fig:hls}
\end{figure}

This approach determines the steepest descent direction by solving a carefully formulated auxiliary mathematical problem that ensures updates to the architecture parameters consistently decrease the validation loss while simultaneously maintaining the optimality of the model parameters. By explicitly incorporating the optimality conditions of the lower-level (training) problem, the method guarantees feasibility within the bilevel optimization context and enforces a mathematically justified descent direction. Importantly, under certain simplifying assumptions, the resulting mathematical program may reduce to a simpler one, such as an LP, thereby offering a computationally efficient alternative, though the corresponding descent direction may not be the steepest. This mathematical optimization framework thus unifies rigorous optimization principles with practical considerations, providing an effective solution to the NAS problem. Let us formalize the approach by considering a neural network architecture parameterized by \( A \in \mathbb{R}^p \) (representing architecture parameters), and model weights denoted by \( W \in \mathbb{R}^q \). The objective is to address the following BOP
\[
\min_{A} \mathcal{L}_v(A, \hat{W}(A)) \quad \text{subject to} \quad \hat{W}(A) = \argmin_{W \in \mathscr{W}} \mathcal{L}_t(A, W),
\]
where \( \mathcal{L}_v \) is assumed differentiable and \( \mathcal{L}_t \) is twice differentiable. Considering the Hessian of \( \mathcal{L}_t \) with respect to \( (A, W) \) at the point \( (A^0, W^0) \), we write
\[
\nabla^2_{(A, W)} \mathcal{L}_t(A^0, W^0) =
\begin{bmatrix}
    H_{11} & H_{12} \\
    H_{21} & H_{22}
\end{bmatrix}
\]
By differentiating the first-order optimality condition of the lower-level problem, \( \nabla_W \mathcal{L}_t(A, W) = 0 \), with respect to \( A \), the following Jacobian equation is obtained
\[
H_{22} \frac{\partial W}{\partial A} + H_{21} = 0
\]
This relationship implies that for a small change \( d_A \) in the architecture parameters, the corresponding change \( d_W \) in the model parameters, required to preserve optimality of the lower-level problem, is given by
\[
d_W = \frac{\partial W}{\partial A} d_A = -H_{22}^{-1} H_{21} d_A
\]
To analyze the effect of these updates on the validation loss, we consider the first-order Taylor's expansion of \( \mathcal{L}_v \) around the current parameters \( (A^0, W^0) \)
\[
\mathcal{L}_v(A^0 + t d_A, W^0 + t d_W) \approx \mathcal{L}_v(A^0, W^0) + t \left( \langle \nabla_A \mathcal{L}_v, d_A \rangle + \langle \nabla_W \mathcal{L}_v, d_W \rangle \right)
\]
To ensure that the update leads to a decrease in the validation loss while maintaining lower-level optimality, we seek the steepest descent direction by solving the following optimization problem
\begin{equation}\label{math_program}
\begin{aligned}
\min_{d_A, d_W} \quad & \langle \nabla_A \mathcal{L}_v, d_A \rangle + \langle \nabla_W \mathcal{L}_v, d_W \rangle \\
\text{subject to:} \quad & H_{21} d_A + H_{22} d_W = 0 \\
& \|d_A\|_2 \leq 1
\end{aligned}
\end{equation}
This formulation is an SOCP, where the affine constraint enforces the preservation of lower-level optimality. For computational efficiency, the norm constraint can be relaxed to a box constraint, \( -1 \leq d_A \leq 1 \), resulting in an LP \eqref{linear_program}
\begin{equation}\label{linear_program}
\begin{aligned}
\min_{d_A, d_W} \quad & \langle \nabla_A \mathcal{L}_v, d_A \rangle + \langle \nabla_W \mathcal{L}_v, d_W \rangle \\
\text{subject to:} \quad & H_{21} d_A + H_{22} d_W = 0 \\
& -1 \leq d_A \leq 1
\end{aligned}
\end{equation}
After determining the optimal descent direction \( (d_A^*, d_W^*) \) by solving the above auxiliary mathematical problems, the parameters are updated as follows
\[
(A^{k+1}, W^{k+1}) \leftarrow (A^k, W^k) + t (d_A^*, d_W^*)
\]
Since \( W^{k+1} \) may not exactly satisfy the first-order optimality conditions due to approximation or inexact SOCP solutions, the model parameters are further refined by resolving the lower-level optimization with a warm start
\[
W^{k+1} = \argmin_{W \in \mathscr{W}} \mathcal{L}_t(A^{k+1}, W)
\]
This iterative process ensures that each update direction yields descent in the validation loss while maintaining, or restoring, the optimality of the model parameters with respect to the training loss. The optimal step-size \((t^*)\) used in the update of parameters can be obtained by a bit of additional experiment as illustrated in the Figure~\ref{fig:hls} \citep{sinha2025linear}. Starting from the current point \((A^0, W^0)\), the search direction \((d_A^*, d_W^*)\) is determined to achieve descent in the validation loss while maintaining lower-level optimality. The figure depicts the trajectories of the training and validation losses along this direction, with the optimal step-size \((t^*)\) identified by minimizing the validation loss along the update path \((A^0 + t d_A^*, W^0 + t d_W^*)\).

In Auxiliary Program-based Neural Architecture Search (AP-NAS), the computation of the Hessian matrix, which encapsulates second-order curvature information, poses significant computational and memory challenges due to the typically large number of model parameters involved. To make this process tractable, various Hessian approximation strategies have been introduced. One widely adopted technique is the L-BFGS method, which offers an efficient alternative by storing only a limited number of past updates to parameters and gradients. This allows for an implicit construction of the Hessian inverse, leading to a per-iteration complexity of \(O(mn)\), where \(m\) denotes the memory size (number of stored updates) and \(n\) represents the dimensionality of the parameter space. Another practical approximation involves representing the Hessian as a weighted sum of rank-one matrices, which are derived from the differences in parameter values and gradient evaluations between consecutive optimization steps \cite{shukla2026apnas}. These methods significantly reduce the computational burden while preserving enough curvature information for effective optimization. Given the impracticality of computing and storing the full Hessian in large-scale models, where it could demand memory in the order of tens of terabytes, reduction techniques are essential. One such strategy involves selecting only a small subset of model parameters for which second-order derivatives are evaluated \cite{shukla2026apnas}. This yields a reduced Hessian matrix that still retains critical information for guiding upper-level parameter updates, especially when paired with efficient approximations like L-BFGS.

This mathematical framework is broadly applicable for enhancing any differentiable neural architecture search algorithm. For example, the framework enhances DARTS, yielding two variants of DARTS, LP-DARTS and SOCP-DARTS, that reformulate the architecture update as linear and second-order cone programs, respectively. Beyond the problem class (LP vs. SOCP), the two variants also differ in how they select model parameters to construct the reduced Hessian matrix and the validation gradients used to solve the upper-level problem. LP-DARTS employs a deterministic criterion known as Gradient Norm Per Parameter (GNPP). This metric evaluates each tensor of model parameters based on the normalized magnitude of its training loss gradient, favoring selections with higher gradient norms relative to their size. Mathematically, GNPP is computed as the Euclidean norm of a parameter tensor’s gradient divided by the total number of its elements, thereby balancing the influence of parameter scale and sensitivity. The tensor with the maximum GNPP is selected for inclusion in the reduced Hessian computation, ensuring that the most impactful parameters in terms of gradient dynamics are prioritized. In contrast, SOCP-DARTS adopts a random selection strategy to choose parameter tensors for reduced Hessian and validation gradient evaluations. This approach mitigates bias in parameter updates by diversifying the subset of parameters considered across iterations. The strategy involves constructing a candidate set of parameter tensors whose size does not exceed a small fraction of the total parameter count. From this candidate set, a fixed number of tensors are randomly sampled, ensuring that the overall number of selected parameters remains small. This randomness allows for a broader coverage of the parameter space over multiple updates and avoids consistently favoring only the most active gradients. The random sampling mechanism emphasize diversity and fairness in parameter update opportunities. When integrated within a SOCP formulation, this sampling strategy ensures that updates to the architecture parameters maintain bilevel feasibility by enforcing the lower-level optimality condition, while achieving descent in validation loss. By combining systematic parameter selection methods with computationally efficient Hessian approximations, both LP-DARTS and SOCP-DARTS offer scalable and theoretically grounded frameworks for neural architecture search. The LP-DARTS method leverages gradient-based heuristics for informed parameter subset construction, while SOCP-DARTS emphasizes stochasticity and broad exploration. Interestingly, the architecture update in AP-NAS is similar to that in hypergradient-based methods. The key difference is that AP-NAS computes a descent direction that always preserves the optimality of the model parameters. In contrast, hypergradient-based methods update only the architecture parameters using the hypergradient direction, which generally disturbs the optimality of the model parameters.

Other than the NAS methods discussed above, there are several alternative perspectives for architecture search. For example, the authors in \cite{shukla2026ngma} considered Neuron Gating and Mixed Activation (NGMA) for architecture search. First, Neuron Gating (NG) can be modeled using continuous architecture parameters, where the parameters associated with hidden neurons control their effective participation in the network. Second, in Mixed Activation (MA)-based architecture search, each hidden neuron is associated with a mixed activation function defined as a convex combination of candidate activation functions drawn from a predefined set. By optimizing the corresponding mixing coefficients, the search process ultimately selects a single optimal activation function for each neuron. In the third setting, both perspectives are combined, leading to NGMA-based architecture search. These three approaches naturally give rise to bilevel optimization formulations, which can be solved using the bilevel theory-based NAS methods discussed above.

\section{Comparative Results of NAS Methods from Literature}\label{sec:comparative}
NAS has evolved into one of the most dynamic subfields in deep learning, with a diverse set of methodologies demonstrating competitive results across benchmark datasets such as CIFAR-10, ImageNet, and more recently, large-scale transformer and graph domains. To provide a unified perspective on performance trends, this section consolidates results from key NAS algorithms, highlighting accuracy, parameter efficiency, and computational requirements. The data presented are drawn from original studies and comparative analysis in the NAS literature \cite{zoph2018learning, pham2018efficient, real2019regularized, liu2019darts, xu2019pcdarts, cai2018proxylessnas, wu2019fbnet, tan2019mnasnet, tan2019efficientnet, cai2019once, chen2021autoformer, gao2019graphnas, chen2019progressive, cai2024sto}.

\subsection{NAS Methods on CIFAR-10 and ImageNet}
Table~\ref{tab:comparative_main} summarizes representative NAS methods spanning RL, EC, differentiable optimization, and meta-learning paradigms. Each approach is evaluated based on Top-1 accuracy, parameter count, and computational cost (architecture search time) measured in GPU days. The results clearly show the efficiency gains achieved by differentiable NAS and one-shot frameworks compared to earlier black-box methods.
\begin{table}[ht]
\centering
\caption{Comparative performance of representative NAS methods on CIFAR-10 and ImageNet benchmarks.}
\label{tab:comparative_main}
\resizebox{\textwidth}{!}{
\begin{tabular}{lcccccc}
\toprule
\textbf{Method} & \textbf{Search Type} & \textbf{Search Space} & \textbf{Dataset} & \textbf{Top-1 Acc. (\%)} & \textbf{Params (M)} & \textbf{GPU Days} \\
\midrule
NASNet-A \citep{zoph2018learning} & RL-based & Cell-based & CIFAR-10 (ImageNet) & 97.35 (74.0) & 3.3 (5.3) & 2000 \\
ENAS \citep{pham2018efficient} & RL (Weight Sharing) & Cell-based & CIFAR-10 & 97.11 & 4.6 & 0.5 \\
AmoebaNet-A \citep{real2019regularized} & Evolutionary & Cell-based & CIFAR-10 (ImageNet) & 96.66 (74.5) & 3.2 (5.1) & 3150 \\
DARTS \citep{liu2019darts} & Differentiable & Continuous & CIFAR-10 (ImageNet) & 97.24 (73.3) & 3.3 (4.7) & 4 \\
PC-DARTS \citep{xu2019pcdarts} & Differentiable & Continuous & CIFAR-10 (ImageNet) & 97.43 (74.9) & 3.6 (5.3) & 0.1 \\
ProxylessNAS \citep{cai2018proxylessnas} & Differentiable & Hardware-aware &  CIFAR-10 (ImageNet) & 97.92 (75.1) & 5.7 (7.1) & 4 \\
FBNet \citep{wu2019fbnet} & Differentiable & Hardware-aware & ImageNet & 74.9 & 5.5 & 9 \\
MnasNet \citep{tan2019mnasnet} & RL + Hardware & Mobile Search & ImageNet & 75.2 & 3.9 & 4.5 \\
EfficientNet-B0 \citep{tan2019efficientnet} & Compound Scaling & Platform-aware & ImageNet & 76.3 & 5.3 & - \\
Once-for-All \citep{cai2019once} & One-shot (Meta) & Elastic Supernet & ImageNet & 76.9 & 7.7 & 1.7 \\
Autoformer \citep{chen2021autoformer} & Differentiable & Transformer & ImageNet & 82.4 & 54.0 & - \\
GraphNAS \citep{gao2019graphnas} & RL-based & Graph Neural Nets & Cora (Citeseer) & 84.2 (73.1) & - & - \\
\bottomrule
\end{tabular}}
\end{table}

\subsection{DARTS Variants Under Computational Constraints}
Table~\ref{tab:combined_cifar10_imagenet} presents a consolidated summary of the time-constrained final evaluations on CIFAR-10 and the corresponding transferability results on ImageNet. The architectures for DARTS, LP-DARTS, and SOCP-DARTS were obtained through controlled architecture search and optimal architecture selection experiments conducted for approximately the same duration, ensuring a fair comparison under equal computational budgets \cite{shukla2026apnas}. For the remaining algorithms (P-DARTS, PC-DARTS, STO-DARTSv1, and STO-DARTSv2), the optimal architectures were adopted directly from their respective original publications \cite{chen2019progressive, xu2019pcdarts, cai2024sto}. All required experiments, including search, selection, and evaluation, were performed entirely on a CPU-based platform. Despite comparable search and evaluation times, both LP-DARTS and SOCP-DARTS achieve substantially higher accuracies and better parameter efficiency, demonstrating their effectiveness in discovering high-quality, competitive, and transferable architectures under limited computational resources.

\begin{table}[ht]
\centering
\caption{Evaluation metrics for the best architectures on CIFAR-10 and their transferability on ImageNet under time-constrained settings.}
\label{tab:combined_cifar10_imagenet}
\footnotesize
\resizebox{\textwidth}{!}{
\begin{tabular}{lc@{\hspace{0.25cm}}c@{\hspace{0.25cm}}c@{\hspace{0.5cm}}c@{\hspace{0.25cm}}c@{\hspace{0.25cm}}c}
\toprule
\multirow{2}{*}{\textbf{Method}} & \multicolumn{3}{c}{\textbf{CIFAR-10}} & \multicolumn{3}{c}{\textbf{ImageNet}} \\
\cmidrule(lr){2-4} \cmidrule(lr){5-7}
 & \textbf{Test Acc. (\%)}
 & \shortstack{\textbf{Eval. Cost}\\\textbf{(CPU hours)}} 
 & \textbf{Params (M)}
 & \textbf{Top-1 (5) Acc. (\%)} 
 & \shortstack{\textbf{Eval. Cost}\\\textbf{(CPU hours)}} 
 & \textbf{Params (M)} \\
\midrule
DARTS & 82.57 & 19.66 & 3.38 & 39.09 (64.10) & 99.36 & 5.81 \\
LP-DARTS & 92.56 & 19.91 & 1.66 & 42.50 (67.55) & 99.32 & 3.46 \\
SOCP-DARTS & 90.56 & 20.65 & 2.23 & 44.02 (68.76) & 101.32 & 4.11 \\
P-DARTS & 90.23 & 21.67 & 2.74 & 44.60 (69.38) & 98.96 & 4.94 \\
PC-DARTS & 89.10 & 20.84 & 2.90 & 42.40 (67.39) & 101.53 & 5.27 \\
STO-DARTSv1 & 89.04 & 19.79 & 2.20 & 42.72 (67.78) & 101.52 & 4.15 \\
STO-DARTSv2 & 87.89 & 20.85 & 3.06 & 42.89 (68.05) & 102.32 & 5.37 \\
\bottomrule
\end{tabular}}
\end{table}

\subsection{Analysis and Discussion}
The results from Table~\ref{tab:comparative_main} reveal several noteworthy trends. Early NAS methods such as NASNet and AmoebaNet, which rely on RL or evolutionary strategies, achieve strong accuracy but demand thousands of GPU-days due to discrete search and full training of candidate architectures. The introduction of differentiable NAS methods such as DARTS and PC-DARTS marked a breakthrough in efficiency, reducing computational cost by three orders of magnitude while maintaining competitive accuracy. Hardware-aware designs such as ProxylessNAS, FBNet, and MnasNet further refined this efficiency by explicitly incorporating latency constraints during search, making them viable for mobile and embedded systems. Meta-learning approaches, such as Once-for-All and Autoformer, illustrate a shift toward scalable and adaptive NAS frameworks that can generalize across architectures and modalities. Once-for-All introduces elastic supernetworks, allowing the reuse of pre-trained subnetworks across multiple deployment scenarios, while Autoformer demonstrates that NAS principles extend beyond CNNs to transformer-based architectures. Collectively, these methods highlight an ongoing trend toward generalizable and hardware-efficient NAS. Furthermore, auxiliary programming frameworks (AP-NAS), such as LP-DARTS and SOCP-DARTS, demonstrate superior performance compared to several DARTS variants under our time-constrained experimental settings. It is important to note that the architectures for P-DARTS, PC-DARTS, and STO-DARTS were adopted directly from their original studies, in which extensive experimental efforts were made to identify optimal experimental configurations and used better computational resources. Even under these conditions, the proposed AP-NAS frameworks exhibit competitive results, as summarized in Table~\ref{tab:combined_cifar10_imagenet}.

\section{Summary and Emerging Trends} \label{sec:emerging_trends}
The progression of NAS research reveals a clear trajectory toward greater automation, efficiency, and adaptability. Table~\ref{tab:comparative_summary} presents a concise comparison across NAS paradigms, emphasizing the trade-offs between computational cost, scalability, and generalization.

\begin{table}[ht]
\centering
\caption{Comparative characteristics of major NAS paradigms.}
\label{tab:comparative_summary}
\resizebox{\textwidth}{!}{
\begin{tabular}{lccccl}
\toprule
\textbf{Paradigm} & \textbf{Optimization Type} & \textbf{Scalability} & \textbf{Generalization} & \textbf{Cost (GPU Days)} & \textbf{Representative Works} \\
\midrule
Reinforcement Learning & Non-differentiable & Medium & Moderate & High (1000+) & NASNet, ENAS \\
Evolutionary Computation & Non-differentiable & High & Strong & High (1000+) & AmoebaNet, NSGA-NAS \\
Bayesian Optimization & Surrogate-based & Medium & Moderate & Medium (100–200) & BOHB, TPE-NAS \\
Hypergradient-based NAS & Differentiable & Very High & Strong & Low (1–10) & DARTS, PC-DARTS \\
Meta / One-shot NAS & Hybrid & Very High & Very Strong & Very Low (<2) & Once-for-All \\
Auxiliary Program-based NAS &  Differentiable & Very High & Strong & Low (1–10) & AP-NAS \\
\bottomrule
\end{tabular}}
\end{table}

Overall, differentiable and meta-learning approaches have clearly outperformed traditional RL and evolutionary methods in efficiency while achieving comparable levels of accuracy. The future of NAS is shifting toward multi-objective formulations, transferability across tasks, and integration with large-scale foundation models. These developments suggest a growing convergence between neural architecture search, model compression, and automated machine learning, marking a decisive step toward fully self-optimizing AI systems.
\begin{table}[ht]
\centering
\caption{Training, validation, and test metrics for CIFAR-10 considering 3 and 5 hyperparameters, with and without Hyperlocal Search.}
\label{tab:cifar10_hyperlocal}
\resizebox{\columnwidth}{!}{%
\renewcommand{\arraystretch}{1.2}
\setlength{\tabcolsep}{6pt} 
\begin{tabular}{
    >{\raggedright\arraybackslash}m{3.3cm}
    >{\centering\arraybackslash}m{1.3cm}  
    >{\centering\arraybackslash}m{1.3cm}
    >{\centering\arraybackslash}m{1.3cm}
    >{\centering\arraybackslash}m{1.6cm}
    >{\centering\arraybackslash}m{1.3cm} 
    >{\centering\arraybackslash}m{1.3cm}
    >{\centering\arraybackslash}m{1.3cm}
    >{\centering\arraybackslash}m{1.6cm}
}
\toprule
\multirow{2}{*}{\textbf{Metrics}} &
\multicolumn{4}{c}{\textbf{CIFAR-10 (3HP) Without Hyperlocal Search}} &
\multicolumn{4}{c}{\textbf{CIFAR-10 (3HP) With Hyperlocal Search}} \\
\cmidrule(lr){2-5} \cmidrule(lr){6-9}
& Grid & Random & TPE & QMC & Grid & Random & TPE & QMC \\
\midrule
Training Loss        & 1.95 & 1.91 & 2.04 & 1.89 & 1.60 & 1.64 & 1.63 & 1.56 \\
Validation Loss        & 2.95 & 2.92 & 2.96 & 2.92 & 2.59 & 2.61 & 2.56 & 2.56 \\
Test Loss          & 3.05 & 3.02 & 3.10 & 3.00 & 2.68 & 2.69 & 2.65 & 2.63 \\
Training Accuracy (\%)   & 75.84 & 75.99 & 74.77 & 76.09 & 81.12 & 80.32 & 80.79 & 80.78 \\
Validation Accuracy (\%)   & 71.14 & 71.18 & 70.55 & 71.24 & 75.64 & 74.99 & 75.65 & 75.47 \\
Test Accuracy (\%)     & 68.48 & 68.51 & 68.09 & 68.52 & 72.59 & 72.12 & 72.63 & 72.43 \\
Runtime (sec)      & 5348 & 5220 & 5316 & 5235 & 5597 & 5456 & 5552 & 5467 \\
\midrule
\multirow{2}{*}{\textbf{Metrics}} &
\multicolumn{4}{c}{\textbf{CIFAR-10 (5HP) Without Hyperlocal Search}} &
\multicolumn{4}{c}{\textbf{CIFAR-10 (5HP) With Hyperlocal Search}} \\
\cmidrule(lr){2-5} \cmidrule(lr){6-9}
& Grid & Random & TPE & QMC & Grid & Random & TPE & QMC \\
\midrule
Training Loss        & 2.20 & 2.92 & 2.52 & 2.36 & 1.94 & 2.02 & 2.07 & 2.01 \\
Validation Loss        & 3.25 & 3.32 & 3.48 & 3.42 & 3.02 & 3.05 & 3.06 & 3.07 \\
Test Loss          & 3.35 & 3.39 & 3.58 & 3.50 & 3.02 & 3.15 & 3.16 & 3.17 \\
Training Accuracy (\%)   & 74.97 & 73.62 & 70.90 & 73.37 & 78.83 & 77.60 & 77.10 & 78.31 \\
Validation Accuracy (\%)   & 70.60 & 69.40 & 67.19 & 69.09 & 74.80 & 72.97 & 72.56 & 73.43 \\
Test Accuracy (\%)     & 68.27 & 67.13 & 65.06 & 66.71 & 71.99 & 70.25 & 70.04 & 70.68 \\
Runtime (sec)      & 7455 & 7144 & 7367 & 7295 & 7789 & 7478 & 7712 & 7636 \\
\bottomrule
\end{tabular}%
}
\end{table}
The mathematical formulations presented by \eqref{math_program} and \eqref{linear_program} have also been effectively employed for tuning other hyperparameters, such as regularization strengths, to obtain models that generalize better. This approach allows hyperlocal search, i.e. simultaneous local search on certain hyperparameters and model parameters, and is discussed in detail in \cite{sinha2025linear}. To highlight its practical benefits, Table~\ref{tab:cifar10_hyperlocal} reports one of the key experimental results from that study. In this experiment, several sampling-based hyperparameter optimization methods, namely grid search, random search, TPE-based Bayesian optimization, and Quasi-Monte Carlo (QMC) sampling, were first used to obtain near-optimal hyperparameters and corresponding models. The auxiliary mathematical program was then employed to perform hyperlocal search, which locally refines both model parameters and hyperparameters around these optima. In essence, hyperlocal search enhances any given optimal model by locally optimizing its hyperparameters and model parameters. As shown in Table~\ref{tab:cifar10_hyperlocal}, this approach yields consistent performance improvements through refined regularization tuning. Building on the idea of hyperlocal search, the authors in \cite{shukla2026lift} extended the same principle to the fine-tuning of large transformer models that are prone to overfitting. A lightweight linear programming-based local search approach was employed to guide small, structured updates to the model parameters and hyperparameters during fine-tuning. When applied to GPT-2 Small on the WikiText-2 dataset, the proposed method effectively controlled overfitting and improved the model's generalization performance while enabling efficient fine-tuning across transformer layers, as summarized in Table~\ref{tab:lift}.

\begin{table}[ht]
\centering
\caption{Definitions of hyperparameters used in the experiments for fine-tuning GPT-2.}
\label{tab:hyperparams_gpt}
\renewcommand{\arraystretch}{1.3}
\begin{tabular}{@{}p{3.0cm} p{12.0cm}@{}}
\toprule
\textbf{Symbol} & \textbf{Description} \\
\midrule

$\mathbf{N}_{WE}$ & Number of warm-up epochs of training before initiating hyperlocal search \\

$\mathbf{N}_{TB}$ & Number of tunable transformer blocks considered during fine-tuning \\

$\mathbf{N}_{R}$ & Number of regularizers included in the bilevel regularization formulation \\

$\mathbf{N}_{\text{FULL/MLP}}$ & Structural scope of fine-tuning, indicating whether the entire transformer block (\texttt{FULL}) or only its MLP subcomponent (\texttt{MLP}) is tuned \\

\bottomrule
\end{tabular}
\end{table}

The various hyperparameter configurations used in the fine-tuning experiments are summarized in Table~\ref{tab:hyperparams_gpt} and the results in Table~\ref{tab:lift} report the training, validation, and test perplexity (and corresponding loss) with and without hyperlocal search.

\begin{table}[ht]
\centering
\scriptsize
\caption{Training, validation, and test perplexity (PPL)/loss for WikiText-2 dataset with and without hyperlocal search.\\}
\label{tab:lift}
\begin{tabular}{llllllccc}
\hline
\multicolumn{4}{c}{\textbf{HP}} & \textbf{\shortstack{Sample \\ Size}} & \textbf{\shortstack{PPL/ \\ Loss}} & \textbf{\shortstack{Without \\ HLS}} & \textbf{\shortstack{With \\ HLS}} & \textbf{\shortstack{Test PPL \\ $\downarrow$\;(\%)}} \\
\cline{1-4}
\(\mathbf{N}_{WE}\) & \(\mathbf{N}_{TB}\) & \(\mathbf{N}_{R}\) & \(\mathbf{N}_{\text{FULL/MLP}}\)
 & & & & \\
 \hline
  &  &  &  &  4718 & Training   & 35.84/3.58  & 35.84/3.58 \\
 0 & 3 & 2 & 1/2 &  487  & Validation & 35.60/3.57  & 35.60/3.57 \\
 & & & &  558  & Testing    & 34.51/3.54  &  34.51/3.54 & 0.00 \\
\hline
  &  &  &  &  4718 & Training   &  21.71/3.08 & 21.71/3.08 \\
 1 & 3 & 2 & 1/2 &  487  & Validation & 22.93/3.13  & 22.93/3.13 \\
 & & & &  558  & Testing    & 22.38/3.11  &  22.38/3.11 & 0.00 \\
\hline
 & & & &  4718 & Training   & 14.32/2.66  & 14.32/2.66  \\
 10 & 3 & 2 & 1/2 &  487  & Validation & 23.26/3.15  & 23.26/3.15 \\
 & & & &  558  & Testing    & 22.82/3.13  & 22.82/3.13 & 0.00 \\
\hline
 & & & &  4718 & Training   & 12.20/2.50  & 12.37/2.52 \\
 15 & 3 & 2 & 1/2 &  487  & Validation & 24.58/3.20  & 24.48/3.20 \\
 & & & &  558  & Testing    & 24.23/3.19  & 24.14/3.18 & 0.37 \\
\hline
 & & & &  4718 & Training   & 10.56/2.36  & 11.20/2.42 \\
 20 & 3 & 2 & 1/2 &  487  & Validation & 26.22/3.27  & 25.81/3.25  \\
 & & & &  558  & Testing    & 25.92/3.26   & 25.48/3.24 & 1.70 \\
\hline
 & & & &  4718 & Training   & 9.24/2.22  & 10.50/2.35 \\
 25 & 3 & 2 & 1/2 &  487  & Validation & 28.13/3.34  & 26.87/3.29 \\
 & & & &  558  & Testing    & 27.89/3.33  & 26.56/3.28 & 4.77 \\
\hline
 & & & &  4718 & Training   & 8.21/2.11  & 10.17/2.32 \\
 30 & 3 & 2 & 1/2 &  487  & Validation & 30.00/3.40  & 28.05/3.33 \\
 & & & &  558  & Testing    & 29.78/3.39  & 27.77/3.32 & 6.75 \\
\hline
 & & & &  4718 & Training   & 7.31/1.99   & 10.16/2.32 \\
 35 & 3 & 2 & 1/2 &  487  & Validation & 32.60/3.48  & 28.73/3.36  \\
 & & & &  558  & Testing    & 32.43/3.48   & 28.38/3.35 & 12.49\\
\hline
 & & & &  4718 & Training   & 1.30/0.26  & 1.62/0.48 \\
 25 & 4 & 4 & 2/2 &  487  & Validation & 205.62/5.33   & 152.73/5.03\\
 & & & &  558  & Testing    & 204.86/5.32   & 153.03/5.03 & 25.30 \\
\hline
 & & & &  4718 & Training   & 1.30/0.26  & 1.61/0.48 \\
 25 & 5 & 5 & 2/3 &  487  & Validation & 205.62/5.33 & 151.38/5.02 \\
 & & & &  558  & Testing    & 204.86/5.32 & 151.71/5.02 & 25.94 \\
\hline
\end{tabular}%
\end{table}

\section{Conclusions}\label{sec:conclusions}
This paper has reviewed the role of bilevel optimization in NAS, with an emphasis on how optimization theory can guide automated architecture design. We categorized existing NAS methods into two broad classes: sampling-based approaches, which rely on different sampling strategies for candidate architectures, and bilevel-theoretic approaches, which cast the search process in an optimization framework that explicitly accounts for the interaction between architecture parameters and network weights. Within the latter class, we placed particular focus on differentiable NAS methodologies, where a differentiable architecture is assumed and bilevel optimization approaches are used to search for the best architecture. A comparative analysis highlighted that bilevel-theoretic approaches generally achieve stronger performance than sampling-based methods, both in terms of accuracy and efficiency.

In summary, this paper highlights how bilevel optimization provides both theoretical rigor and empirical benefits for NAS. Results show that recent methods with differentiable architectures and second-order information hold significant promise, therefore, future research may explore scalable structured Hessian approximations (e.g., block-diagonal or staircase forms), develop hybrid optimization schemes that unify other differentiable NAS methods with the discussed auxiliary mathematical programming framework, and extend these ideas to broader search spaces and application domains. Collectively, these directions position bilevel optimization as a promising foundation for the next generation of efficient NAS frameworks.

\bibliography{references}

\appendix
\renewcommand{\thesubsection}{\Alph{subsection}}
\renewcommand{\thesubsubsection}{\Alph{subsection}.\arabic{subsubsection}}
\section*{\centering Appendix}

\subsection{Synthetic Dataset for Overfitting Illustration}\label{app_overfit_data}
This appendix provides the complete dataset used to create Figure~\ref{fig:ann} which illustrates the effect of model complexity on overfitting behavior in neural networks. The dataset is two-dimensional and synthetically constructed to enable clear visualization of classification boundaries. Each data point consists of an input vector $x = (x_1, x_2) \in \mathbb{R}^2$ and a corresponding binary label $y \in \{0,1\}$. We define the training and validation datasets respectively as,
\[
S^{T} = \{\, (x_i^{T}, y_i^{T}) \,\}_{i=1}^{38}, 
\qquad 
S^{V} = \{\, (x_i^{V}, y_i^{V}) \,\}_{i=1}^{18}
\]
where $S^{T}$ and $S^{V}$ are used for model training and validation respectively. The complete training dataset is given by
\[
\begin{aligned}
S^{T} = \{\, &
((-0.25,\,0.05),0),\,((-0.85,\,-0.65),0),\,((-0.80,\,0.15),0),\,((-0.75,\,-0.75),0),\,((-0.65,\,0.50),0),\\
&((-0.60,\,-0.20),0),\,((-0.15,\,0.75),0),\,((0.35,\,-0.85),0),\,((0.50,\,0.60),0),\,((0.80,\,0.40),0),\\
&((0.70,\,-0.60),0),\,((0.90,\,0.40),0),\,((0.95,\,-0.30),0),\,((0.075,\,-0.01),0),\,((0.25,\,0.05),0),\\
&((0.10,\,-0.125),0),\,((0.30,\,0.025),0),\,((0.00,\,0.10),0),\\
&((-0.85,\,-0.40),1),\,((-0.70,\,0.70),1),\,((-0.55,\,-0.90),1),\,((-0.50,\,0.20),1),\,((-0.40,\,-0.55),1),\\
&((-0.30,\,0.90),1),\,((0.15,\,-0.75),1),\,((0.30,\,0.85),1),\,((0.45,\,-0.40),1),\,((0.65,\,0.20),1),\\
&((0.80,\,-0.70),1),\,((0.85,\,0.60),1),\,((-0.65,\,0.10),1),\,((0.15,\,-0.40),1),\,((0.90,\,0.90),1),\\
&((0.15,\,-0.15),1),\,((0.35,\,0.00),1),\,((-0.10,\,0.10),1),\,((0.05,\,-0.20),1),\,((0.25,\,0.20),1)\,\}
\end{aligned}
\]

\noindent The validation dataset, used to monitor generalization performance, is given as
\[
\begin{aligned}
S^{V} = \{\, &
((-0.10,\,0.30),0),\,((-0.85,\,-0.05),0),\,((-0.70,\,0.40),0),\,((-0.55,\,-0.65),0),\,((0.25,\,0.55),0),\\
&((0.40,\,0.40),0),\,((0.50,\,-0.35),0),\,((0.65,\,0.10),0),\,((-0.30,\,-0.15),0),\\
&((-0.50,\,-0.75),1),\,((-0.40,\,0.50),1),\,((0.35,\,-0.60),1),\,((0.70,\,0.25),1),\,((0.80,\,0.30),1),\\
&((0.50,\,-0.20),1),\,((0.10,\,-0.65),1),\,((0.00,\,-0.55),1),\,((-0.70,\,-0.05),1)\,\}
\end{aligned}
\]

\subsection{Toll-Setting Problem} \label{app_toll_setting_problem}
Imagine a government or a private highway operator who wants to decide how much toll to charge for using a highway. The goal is to earn as much revenue as possible. However, if the toll is set too high, people may avoid using the tolled highway, leading to reduced revenue. On the other hand, if the toll is too low, many travelers may use it, but the revenue collected per traveler would be small. Hence, the toll must strike a balance between the toll rate and the number of users. Crucially, the government can not make this decision in isolation, it must account for how users will react to different toll rates. Users are interested in minimizing their own total cost, which depends on travel time, toll, and other monetary expenses. This naturally leads to a BOP, where the upper-level optimization represents the government’s task of setting the toll rate to maximize revenue, and the lower-level decision corresponds to users’ route choices for a given toll rate, i.e., selecting the fraction of distance on each route to minimize their travel cost.\\

\noindent \tb{Parameters and Decision Variables.} Consider a simplified transportation network between origin $A$ and destination $B$, connected by two parallel highways of normalized distance $1$. A traveler may choose to travel a fraction $y \in Y=[0,1]$ of the distance on the tolled highway and $(1-y)$ on the free highway. Switching between highways is usually possible at certain fixed locations along the route. If there are $n$ equally spaced switching points (including $A$ and $B$), then the set of feasible switching locations is $S = \left\{ 0, \tfrac{1}{n-1}, \tfrac{2}{n-1}, \ldots, 1 \right\}$. In the unrestricted case (switching possible everywhere), $S=[0,1]$. We assume that there is no cost involved in the switching process. The two participants in this decision process are:
    \begin{description}
    \item[Leader (Government/Operator):] Chooses the toll rate $\tau \geq 0$ to maximize revenue
    \item[Follower (Users):] Choose $y \in Y$ to minimize their own total cost.
    \end{description}
The parameters and decision variables for the problem are given in Table~\ref{tab:toll_params}.\\

\begin{table}[ht]
\centering
\caption{Parameters and decision variables in the toll-setting problem.}
\label{tab:toll_params}
\renewcommand{\arraystretch}{1.3}
\begin{tabular}{@{}p{3.0cm} p{12.0cm}@{}}
\toprule
\textbf{Symbol} & \textbf{Description} \\
\midrule

$\tau$ & Toll per unit distance (decision variable of the leader) \\

$y$ & Fraction of distance traveled on the tolled highway (decision variable of the follower) \\

$a$ & Travel time per unit distance on the tolled highway \\

$b$ & Travel time per unit distance on the free highway \\

$c$ & Intrinsic (non-toll) cost per unit distance on the tolled highway \\

$d$ & Cost per unit distance on the free highway \\

$w_1$ & Weight given to travel time in the user’s generalized cost \\

$w_2$ & Weight given to monetary cost in the user’s generalized cost \\

\bottomrule
\end{tabular}
\end{table}

\noindent \tb{Lower-Level Problem.} The total travel time and monetary cost are linear functions of $y$, given as follows
\begin{equation}
    T(y) = ay + b(1-y) = (a-b)y + b, 
    \qquad 
    C(y,\tau) = (c+\tau)y + d(1-y) = (c+\tau-d)y + d
\end{equation}
The user evaluates a generalized cost that combines time and money
\begin{equation}
    G(y,\tau) = w_1\big[(a-b)y+b\big]^2 + w_2\big[(c+\tau-d)y+d\big]^2
    \label{eq:gen_cost}
\end{equation}
Given a toll $\tau$, the user solves
\begin{equation}
    \min_{y \in Y} \; G(y,\tau)
    \label{eq:lower_problem}
\end{equation}
This is a convex optimization problem in $y$.\\

\noindent \tb{Upper-Level Problem.} Anticipating the user’s optimal reaction $\hat{y}(\tau)$, the operator seeks to maximize toll revenue as follows
\begin{equation}
    \max_{\tau \ge 0} \; R(\tau) = \tau \cdot \hat{y}(\tau)
    \label{eq:upper_problem}
\end{equation}

\noindent \tb{Bilevel Formulation.} The entire problem can be written compactly as
\begin{equation}\label{bilevel_toll}
\max_{\tau \geq 0} \; \tau \cdot \hat{y} 
\quad \text{subject to} \quad 
\begin{cases}
\hat{y} \in \argmin_{y \in Y} \; G(y, \tau)
\end{cases}
\end{equation}

\noindent \tb{Interpretation.} Formulation~\eqref{bilevel_toll} clearly demonstrates the hierarchical dependency characteristic of bilevel optimization. In this framework, the lower-level problem models the users’ equilibrium behavior, where each user selects the route that minimizes their perceived travel cost. The upper-level problem, on the other hand, represents the leader’s objective of maximizing total revenue while anticipating the users’ optimal reactions to the toll rate~$\tau$. This interdependence between decision layers is what imparts the hierarchical nature to the problem. Although the lower-level problem is convex, ensuring that users’ route choices are uniquely determined, the resulting upper-level objective, $R(\tau) = \tau\, \hat{y}(\tau)$, is typically non-convex. The nonlinearity and piecewise smoothness of the follower’s best-response mapping~$\hat{y}(\tau)$, arising from the argmin operator, make the leader’s optimization landscape inherently challenging. Hence, convexity at the lower-level does not guarantee overall tractability at the upper-level. To explore the interplay between the leader and follower decisions, we first analyze the follower’s optimization problem, where the goal is to determine the optimal traffic split~$y$ minimizing the generalized travel cost for a fixed toll~$\tau$. The leader then seeks the toll rate that maximizes total revenue by anticipating this equilibrium response. To demonstrate the system’s behavior under varying conditions, three representative parameter settings, each reflecting a distinct operational regime, are analyzed. The corresponding optimal solutions are reported in Table~\ref{tab:toll_optima}, with detailed analytical derivations presented as follows.
\subsubsection{The Follower's Response: A Convex Quadratic Problem} \label{app_11}
The follower's objective is to solve
$$
\hat{y}(\tau) \in \underset{y}{\argmin} \Big\{ G(y, \tau) \;\Big|\; 0 \le y \le 1 \Big\},
$$
where the generalized cost $G(y,\tau)$ from \eqref{eq:gen_cost} is a quadratic function of the decision variable $y$. Let's expand this function to make its structure clear. Assuming weights $w_1=w_2=1$
\begin{align*}
    G(y, \tau) &= \big(ay + b(1-y)\big)^2 + \big((c+\tau)y + d(1-y)\big)^2 \\
    &= \big((a-b)y + b\big)^2 + \big((c+\tau-d)y + d\big)^2 \\
    &= \underbrace{\left[ (a-b)^2 + (c+\tau-d)^2 \right]}_{A(\tau)}\,y^2 + \underbrace{\left[ 2b(a-b) + 2d(c+\tau-d) \right]}_{B(\tau)}\,y + \underbrace{\left[ b^2 + d^2 \right]}_{\text{const}}
\end{align*}
The problem is thus a convex quadratic program, $\min_{y \in [0,1]} A(\tau)y^2 + B(\tau)y$, whose solution is unique. The unconstrained minimizer of this quadratic can be found by setting the derivative with respect to $y$ to zero: $2A(\tau)y + B(\tau) = 0$. This gives
\begin{equation}
    y^{\mathrm{unc}}(\tau) = -\frac{B(\tau)}{2A(\tau)}
    \label{eq:yunc_def}
\end{equation}
However, the follower's choice $y$ is constrained to the interval $[0,1]$. The optimal solution $\hat{y}(\tau)$ is therefore the projection of the unconstrained minimizer onto this interval. This can be derived from the KKT conditions discussed next.

\subsubsection{Derivation of the Follower's Response via KKT Conditions}\label{app_12}
To apply the KKT conditions, we first state the problem in standard form. Let $f(y) = A(\tau)y^2 + B(\tau)y$. The constraints are $g_1(y)=-y \le 0$ and $g_2(y)=y-1 \le 0$. The Lagrangian is
$$ \mathcal{L}(y, \mu_1, \mu_2) = f(y) + \mu_1 g_1(y) + \mu_2 g_2(y) = A(\tau)y^2 + B(\tau)y - \mu_1 y + \mu_2(y-1),$$
where $\mu_1, \mu_2 \ge 0$ are the KKT multipliers. The KKT conditions for a minimum are
\[
\begin{aligned}
& \text{(stationarity)} && \frac{\partial \mathcal{L}}{\partial y} = 2A(\tau)y + B(\tau) - \mu_1 + \mu_2 = 0, \\
& \text{(primal feasibility)} && 0 \leq y \leq 1, \\
& \text{(dual feasibility)} && \mu_1 \geq 0,\;\; \mu_2 \geq 0, \\
& \text{(complementarity)} && \mu_1 y = 0,\;\; \mu_2(y-1)=0
\end{aligned}
\]
We analyze the solution by considering which constraints are active.
\begin{itemize}
\item \tb{Interior Solution ($0 < \hat{y} < 1$).}
Since $y \neq 0$ and $y-1 \neq 0$, complementary slackness requires $\mu_1=0$ and $\mu_2=0$. The stationarity condition simplifies to
$$ 2A(\tau)\hat{y} + B(\tau) = 0 \implies \hat{y} = -\frac{B(\tau)}{2A(\tau)} = y^{\mathrm{unc}}(\tau) $$
This solution is valid if and only if it satisfies the case's assumption, i.e., $0 < y^{\mathrm{unc}}(\tau) < 1$.
\item \tb{Boundary Solution ($\hat{y}=0$).}
The condition $y=0$ is active. Complementary slackness requires $\mu_2(0-1)=0 \implies \mu_2=0$. The stationarity condition becomes
$$ 2A(\tau)(0) + B(\tau) - \mu_1 + 0 = 0 \implies \mu_1 = B(\tau) $$
For dual feasibility, we must have $\mu_1 \ge 0$, which means $B(\tau) \ge 0$. Since $A(\tau) > 0$, the condition $B(\tau) \ge 0$ is equivalent to $-\frac{B(\tau)}{2A(\tau)} \le 0$, or $y^{\mathrm{unc}}(\tau) \le 0$. Thus, $\hat{y}=0$ is the solution if $y^{\mathrm{unc}}(\tau) \le 0$.
\item \tb{Boundary Solution ($\hat{y}=1$).}
The condition $y=1$ is active. Complementary slackness requires $\mu_1(1)=0 \implies \mu_1=0$. The stationarity condition becomes
$$ 2A(\tau)(1) + B(\tau) - 0 + \mu_2 = 0 \implies \mu_2 = -2A(\tau) - B(\tau)$$
For dual feasibility, we must have $\mu_2 \ge 0$, which means $-2A(\tau) - B(\tau) \ge 0$, or $B(\tau) \le -2A(\tau)$. Dividing by $2A(\tau)$ (which is positive) gives $\frac{B(\tau)}{2A(\tau)} \le -1$. Multiplying by $-1$ and reversing the inequality gives $-\frac{B(\tau)}{2A(\tau)} \ge 1$, which is $y^{\mathrm{unc}}(\tau) \ge 1$. Thus, $\hat{y}=1$ is the solution if $y^{\mathrm{unc}}(\tau) \ge 1$.
\end{itemize}
Combining the above three scenarios, we get the lower-level optimal response as follows
\begin{equation}
    \hat{y}(\tau) =
    \begin{cases}
        0 & \text{if } y^{\mathrm{unc}}(\tau) \le 0 \\
        1 & \text{if } y^{\mathrm{unc}}(\tau) \ge 1 \\
        y^{\mathrm{unc}}(\tau) & \text{if } 0 < y^{\mathrm{unc}}(\tau) < 1
    \end{cases}
    \label{eq:ystar_piecewise}
\end{equation}
This can be written more compactly as
\begin{equation}
    \hat{y}(\tau) = \min\!\big\{\,1,\; \max\{\,0,\; y^{\mathrm{unc}}(\tau)\,\}\big\}
    \label{eq:ystar_projection}
\end{equation}
For the interior case where $0 < \hat{y}(\tau) < 1$, the closed-form solution is
\begin{equation}
    \hat{y}(\tau) = -\frac{B(\tau)}{2A(\tau)} = \frac{b(b-a) + d(d-c-\tau)}{(a-b)^2 + (c+\tau-d)^2}
    \label{eq:ystar_closed}
\end{equation}

\subsubsection{Three Canonical Regimes (Worked Cases)}\label{app_13}
To understand the interplay between the leader and follower, we analyze three parameter sets $(a,b,c,d)$ that exhibit distinct follower responses.
\begin{itemize}
\item \tb{Case 1: Never Tolled (Avoidance).}
Consider parameters $(a,b,c,d)=(2,1,3,1)$. Here, the tolled route is inherently less attractive (slower travel time and higher base cost). Let's derive the follower's response.
\begin{align*}
    A(\tau) &= (2-1)^2 + (3+\tau-1)^2 = 1 + (\tau+2)^2 = \tau^2 + 4\tau + 5 \\
    B(\tau) &= 2(1)(2-1) + 2(1)(3+\tau-1) = 2 + 2(\tau+2) = 2\tau + 6
\end{align*}
The unconstrained minimizer is
$$
y^{\mathrm{unc}}(\tau) = -\frac{B(\tau)}{2A(\tau)} = -\frac{2\tau+6}{2(\tau^2+4\tau+5)} = -\frac{\tau+3}{\tau^2+4\tau+5}
$$
For any non-negative toll $(\tau \ge 0)$, the numerator is positive and the denominator is positive. Therefore, $y^{\mathrm{unc}}(\tau) < 0$ for all $\tau \ge 0$. According to \eqref{eq:ystar_piecewise}, the follower's best response is always to be at the lower boundary
$$
\hat{y}(\tau) = 0 \quad \forall \tau \ge 0
$$
Consequently, the leader’s revenue is $R(\tau) = \tau \cdot 0 = 0$. The leader cannot extract any revenue.

\item \tb{Case 2: Always Tolled up to a Threshold.}
Consider parameters $(a,b,c,d)=(1,3,1,3)$. In this case, the tolled route is far superior (faster and cheaper).
\begin{align*}
    A(\tau) &= (1-3)^2 + (1+\tau-3)^2 = 4 + (\tau-2)^2 = \tau^2 - 4\tau + 8 \\
    B(\tau) &= 2(3)(1-3) + 2(3)(1+\tau-3) = -12 + 6(\tau-2) = 6\tau - 24
\end{align*}
The unconstrained minimizer is
$$
y^{\mathrm{unc}}(\tau) = -\frac{B(\tau)}{2A(\tau)} = -\frac{6\tau-24}{2(\tau^2-4\tau+8)} = \frac{12-3\tau}{\tau^2-4\tau+8}
$$
The follower will use the tolled route exclusively ($\hat{y}=1$) as long as the unconstrained solution, $y^{\mathrm{unc}}(\tau) \ge 1$. We find the threshold toll $\tau_{th}$ by solving $y^{\mathrm{unc}}(\tau_{th}) = 1$
\begin{align*}
    \frac{12-3\tau_{th}}{\tau_{th}^2-4\tau_{th}+8} &= 1 \\
    12 - 3\tau_{th} &= \tau_{th}^2 - 4\tau_{th} + 8 \\
    0 &= \tau_{th}^2 - \tau_{th} - 4
\end{align*}
The positive root of this quadratic is $\tau_{th} = \frac{1+\sqrt{1-4(1)(-4)}}{2} = \frac{1+\sqrt{17}}{2} \approx 2.56$.
For $0 \le \tau \le \tau_{th}$, we have $y^{\mathrm{unc}}(\tau) \ge 1$, so the follower's response is $\hat{y}(\tau)=1$. In this regime, the leader's revenue is $R(\tau) = \tau$. To maximize this, the leader should choose the largest possible toll, pushing it to the threshold value: $\tau^\star = \tau_{th} \approx 2.56$. At this point, revenue is $R^\star \approx 2.56$. If $\tau > \tau_{th}$, usage begins to drop according to the interior solution.
\begin{figure}[htb]
    \centering
    \includegraphics[width=0.6\textwidth]{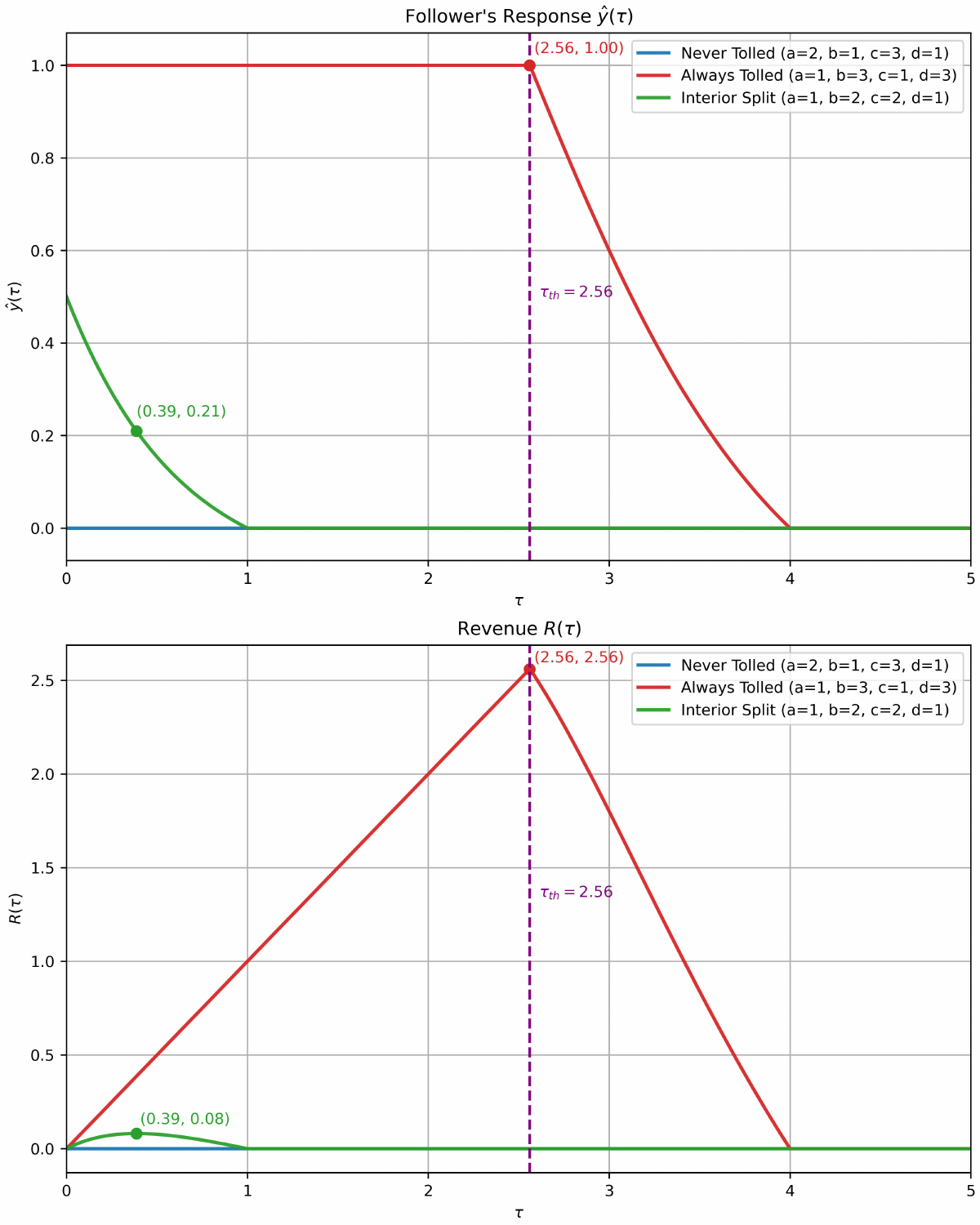}
    \caption{Follower’s response and leader revenue across three regimes in the toll‑setting bilevel model.}
    \label{fig:toll_diagram}
\end{figure}
\begin{description}
    \item[\tb{Analysis for Tolls above the Threshold ($\tau > \tau_{th}$):}] To confirm that $\tau_{th}$ is the global optimum, we must analyze the leader's revenue when the toll is set higher than this threshold. For $\tau > \tau_{th} \approx 2.56$, the unconstrained response $y^{\mathrm{unc}}(\tau)$ will be less than 1. We must also check where it becomes non-positive by setting the numerator to zero: $12-3\tau=0 \implies \tau=4$. This defines two sub-regions for $\tau > \tau_{th}$.
    \begin{description}
        \item[\tb{Interior Solution Region ($\tau_{th} < \tau < 4$).}] In this range, $0 < y^{\mathrm{unc}}(\tau) < 1$. The follower's response is the interior solution
    $$
    \hat{y}(\tau) = y^{\mathrm{unc}}(\tau) = \frac{12-3\tau}{\tau^2-4\tau+8}
    $$
    The leader's revenue function is
    $$
    R(\tau) = \tau \cdot \hat{y}(\tau) = \frac{12\tau - 3\tau^2}{\tau^2-4\tau+8}
    $$
    Computing the derivative $\frac{dR}{d\tau}$ using the quotient rule. Let $N(\tau) = 12\tau - 3\tau^2$ and $D(\tau) = \tau^2-4\tau+8$
    $$
    \frac{dR}{d\tau} = \frac{N'(\tau)D(\tau) - N(\tau)D'(\tau)}{[D(\tau)]^2} = \frac{(12-6\tau)(\tau^2-4\tau+8) - (12\tau-3\tau^2)(2\tau-4)}{(\tau^2-4\tau+8)^2}
    $$
    The sign of the derivative is determined by its numerator, which simplifies to
    \begin{align*}
    \text{Num}(\tau) &= (-6\tau^3 + 36\tau^2 - 96\tau + 96) - (-6\tau^3 + 36\tau^2 - 48\tau) \\
    &= -96\tau + 96 + 48\tau \\
    &= 96 - 48\tau = 48(2-\tau)
    \end{align*}
    The derivative is zero when $48(2-\tau)=0$, which implies $\tau=2$. This stationary point is outside the region of interest ($\tau > 2.56$). For any $\tau$ in the interval $(\tau_{th}, 4)$, the term $(2-\tau)$ is negative. Therefore, $\frac{dR}{d\tau} < 0$ throughout this entire region. This proves that the revenue function is strictly decreasing for all tolls immediately following the threshold.
    \item[\tb{Zero Usage Region ($\tau \ge 4$).}] In this range, $y^{\mathrm{unc}}(\tau) \le 0$. The follower's response becomes $\hat{y}(\tau) = 0$. Consequently, the leader's revenue is $R(\tau) = 0$. 
    \end{description}
\end{description}

\item \tb{Case 3: Interior Split (Partial Usage).}
Consider parameters $(a,b,c,d)=(1,2,2,1)$, where neither route strictly dominates the other. The coefficients of the follower's quadratic cost function are
\begin{align*}
    A(\tau) &= (1-2)^2 + (2+\tau-1)^2 = 1 + (\tau+1)^2 = \tau^2+2\tau+2 \\
    B(\tau) &= 2(2)(1-2) + 2(1)(2+\tau-1) = -4 + 2(\tau+1) = 2\tau-2
\end{align*}
The unconstrained minimizer is
$$
y^{\mathrm{unc}}(\tau) = -\frac{B(\tau)}{2A(\tau)} = -\frac{2\tau-2}{2(\tau^2+2\tau+2)} = \frac{1-\tau}{\tau^2+2\tau+2}
$$
For any toll $\tau \in [0,1]$, the numerator $(1-\tau)$ is in $[0,1]$ and the denominator is $\ge 2$. Thus, we have $0 \le y^{\mathrm{unc}}(\tau) \le 1/2$. This confirms that for this range of tolls, the follower's optimal response is always an interior solution, $\hat{y}(\tau) = y^{\mathrm{unc}}(\tau)$
\begin{equation}
    \hat{y}(\tau) = \frac{1-\tau}{\tau^2 + 2\tau + 2}, \qquad \text{for } \tau \in [0,1]
    \label{eq:split_formulas_kkt}
\end{equation}
The leader's problem is to maximize the revenue function $R(\tau) = \tau \hat{y}(\tau)$ over the feasible interval for $\tau$
$$
\begin{aligned}
\max_{\tau} \quad & R(\tau) = \frac{\tau(1-\tau)}{\tau^2 + 2\tau + 2} \\
\text{subject to} \quad & \tau \ge 0 \\
& \tau \le 1
\end{aligned}
$$
This is a constrained non-linear optimization problem. We solve it using the KKT conditions. First, we write the constraints in the standard form $g_i(\tau) \le 0$
\begin{align*}
    g_1(\tau) = -\tau \le 0 \\
    g_2(\tau) = \tau - 1 \le 0
\end{align*}
The Lagrangian function $\mathcal{L}$ for this maximization problem is
$$
\mathcal{L}(\tau, \lambda_1, \lambda_2) = R(\tau) - \lambda_1(-\tau) - \lambda_2(\tau-1) = R(\tau) + \lambda_1\tau + \lambda_2(1-\tau),
$$
where $\lambda_1, \lambda_2$ are the non-negative Lagrange multipliers. The KKT conditions for an optimal solution $\tau^\star$ are
\[
\begin{aligned}
& \text{(stationarity)} && \frac{\partial \mathcal{L}}{\partial \tau}\Big|_{\tau^\star} 
= \frac{dR}{d\tau}\Big|_{\tau^\star} + \lambda_1 - \lambda_2 = 0, \\
& \text{(primal feasibility)} && 0 \leq \tau^\star \leq 1, \\
& \text{(dual feasibility)} && \lambda_1 \geq 0,\;\; \lambda_2 \geq 0, \\
& \text{(complementarity)} && \lambda_1(-\tau^\star) = 0,\;\; \lambda_2(\tau^\star - 1) = 0
\end{aligned}
\]
The derivative of the revenue function using the quotient rule, where $N(\tau) = \tau-\tau^2$ and $D(\tau) = \tau^2+2\tau+2$  is given as
$$
\frac{dR}{d\tau} = \frac{N'(\tau)D(\tau) - N(\tau)D'(\tau)}{[D(\tau)]^2} = \frac{(1-2\tau)(\tau^2+2\tau+2) - (\tau-\tau^2)(2\tau+2)}{(\tau^2+2\tau+2)^2}
$$
$$
\frac{dR}{d\tau} = \frac{-3\tau^2 - 4\tau + 2}{(\tau^2 + 2\tau + 2)^2}
$$
We now analyze the possible cases based on which constraints are active.
\begin{description}
    \item[\tb{Interior solution ($0 < \tau^\star < 1$):}] From complementary slackness, since $\tau^\star \neq 0$ and $\tau^\star-1 \neq 0$, we must have $\lambda_1 = 0$ and $\lambda_2 = 0$. The stationarity condition simplifies to $\frac{dR}{d\tau} = 0$. This requires the numerator to be zero
$$
-3(\tau^\star)^2 - 4\tau^\star + 2 = 0 \quad \implies \quad 3(\tau^\star)^2 + 4\tau^\star - 2 = 0
$$
The solutions are $\tau = \frac{-4 \pm \sqrt{16 - 4(3)(-2)}}{6} = \frac{-4 \pm \sqrt{40}}{6}$. The only positive root is
$$
\tau^\star = \frac{-4 + \sqrt{40}}{6} = \frac{-2+\sqrt{10}}{3} \approx 0.387
$$
This value lies within the interval $(0,1)$, satisfying the primal feasibility assumption for an interior solution. Thus, $\tau^\star \approx 0.387$ with $\lambda_1=\lambda_2=0$ is a valid KKT point.
\item[\tb{Boundary solution at $\tau^\star = 0$:}] The constraint $g_1(\tau)$ is active. From complementary slackness, $\lambda_2(\tau^\star-1) = \lambda_2(-1) = 0 \implies \lambda_2=0$. The stationarity condition becomes $\frac{dR}{d\tau}\Big|_{\tau=0} + \lambda_1 = 0$. Let's evaluate the derivative at $\tau=0$
$$
\frac{dR}{d\tau}\Big|_{\tau=0} = \frac{-3(0)^2 - 4(0) + 2}{(0^2 + 2(0) + 2)^2} = \frac{2}{4} = \frac{1}{2}
$$
The stationarity condition thus requires $\frac{1}{2} + \lambda_1 = 0 \implies \lambda_1 = -1/2$. This violates the dual feasibility condition $\lambda_1 \ge 0$. Therefore, $\tau^\star=0$ is not the maximum.
\item [\tb{Boundary solution at $\tau^\star = 1$:}] The constraint $g_2(\tau)$ is active. From complementary slackness, $\lambda_1(-\tau^\star) = \lambda_1(-1) = 0 \implies \lambda_1=0$. The stationarity condition becomes $\frac{dR}{d\tau}\Big|_{\tau=1} - \lambda_2 = 0$. Evaluating the derivative at $\tau=1$
$$
\frac{dR}{d\tau}\Big|_{\tau=1} = \frac{-3(1)^2 - 4(1) + 2}{(1^2 + 2(1) + 2)^2} = \frac{-5}{25} = -\frac{1}{5}
$$
The stationarity condition requires $-\frac{1}{5} - \lambda_2 = 0 \implies \lambda_2 = -1/5$. This violates the dual feasibility condition $\lambda_2 \ge 0$. Therefore, $\tau^\star=1$ is not the maximum.
\end{description}
The only solution satisfying all KKT conditions is the interior point. The optimal toll is therefore $\tau^\star \approx 0.387$. At this toll, the follower's usage and the leader's revenue are
\begin{align*}
y^\star(\tau^\star) &\approx \frac{1 - 0.387}{(0.387)^2 + 2(0.387) + 2} \approx \frac{0.613}{2.924} \approx 0.21 \\
R(\tau^\star) &= \tau^\star y^\star(\tau^\star) \approx 0.387 \times 0.21 \approx 0.081
\end{align*}

For any toll $\tau > 1$, the tolled route becomes economically unattractive ($y^{\mathrm{unc}}(\tau)<0$), and the follower's optimal response is to completely avoid it, resulting in zero usage ($\hat{y}(\tau)=0$). Consequently, the leader's revenue $R(\tau)$ is also zero, making any such toll a suboptimal choice compared to the interior solution.
\end{itemize}
The solutions of the toll‑setting problem for different parameter regimes are given in Table~\ref{tab:toll_optima}. The phase diagram given by Figure~\ref{fig:toll_diagram} depicts a dominated tolled route (never use), a dominant tolled route (always use up to a threshold), and comparable routes (interior split). They illustrate how the leader’s non-convex objective arises from the follower’s simple, convex optimal response function.

\begin{table}[ht]
\centering
\caption{Optimal tolls, user's optimal responses, and maximum revenues in different regimes of the toll-setting problem.}
\label{tab:toll_optima}
\renewcommand{\arraystretch}{1.6}
\begin{tabular}{@{}p{8cm} p{7.0cm}@{}}
\toprule
\textbf{Regime Type} & \textbf{Optimal Solutions and Objective Values} \\
\midrule

Interior Split (Partial Usage)
& Parameters: $(a,b,c,d) = (1,2,2,1)$ \newline
Optimal toll: $\tau^\star = \tfrac{-2+\sqrt{10}}{3} \approx 0.39$ \newline
Follower’s optimal response: $y^* = \approx 0.21$ \newline
Leader’s maximum revenue: $R^\star \approx 0.08$ \\

\addlinespace
Never Tolled (Avoidance) 
& Parameters: $(a,b,c,d) = (2,1,3,1)$ \newline
Follower’s optimal response: $y^* = 0$ \newline
Optimal toll: $\tau^\star =$ arbitrary (any $\tau \ge 0$) \newline
Leader’s maximum revenue: $R^\star = 0$ \\

\addlinespace
Always Tolled up to a Threshold
& Parameters: $(a,b,c,d) = (1,3,1,3)$ \newline
Threshold toll: $\tau_{th} \approx 2.56$ \newline
Optimal toll: $\tau^\star \approx 2.56$ \newline
Follower’s optimal response: $y^* = 1$ \newline
Leader’s maximum revenue: $R^\star \approx 2.56$ \\

\bottomrule
\end{tabular}
\end{table}

\noindent \tb{Remarks.} The convexity of the follower’s problem guarantees a unique equilibrium response~$\hat{y}(\tau)$ for each admissible toll rate. However, the leader’s objective remains non-convex and piecewise due to discontinuities introduced by the follower’s best-response mapping. Consequently, optimal toll values may occur either at the boundary cases, where the user fully commits to one route ($\hat{y}\in\{0,1\}$), or within the interior, where mixed choices emerge. This structural property is emblematic of many bilevel formulations encountered in practice, such as pricing, resource allocation, and hyperparameter optimization, where the leader must anticipate rational but constrained follower behavior. These interactions often yield non-smooth optimization landscapes that require specialized analytical and computational treatment.

\subsection{Real and Standard Formulations}\label{app_real_standard_formulations}
The real and standard formulations for the optimistic bilevel problem are described as follows. Their differences are illustrated using a simple example. Similar descriptions can be developed for the pessimistic bilevel problem.

Consider the reaction set defined as
\[
\Psi(x) = \left\{ \hat{y} \;:\; \hat{y} \in \argmin_{y} \left\{ f(x, y) \mid g(x, y) \leq 0,\; h(x, y) = 0 \right\} \right\}
\]

\begin{description}

\item[\tb{Real Optimistic Formulation:}] 
The real optimistic formulation is conceptually more appropriate because, from the perspective of the upper-level decision maker, the value of the objective function is of primary importance rather than the explicit attainment of a particular lower-level solution within the reaction set. It is defined as follows
\begin{equation} \label{real_opt}
    \min_{x} \; \inf_{\hat{y} \in \Psi(x)} F(x, \hat{y}) 
    \quad \text{subject to} \quad 
    G(x, \hat{y}) \leq 0,\; H(x, \hat{y}) = 0
\end{equation}

\item[\tb{Standard Optimistic Formulation:}] 
The standard optimistic formulation is more tractable. It is equivalent to the real optimistic formulation in the case where the infimum in the inner minimization problem of formulation~\eqref{real_opt} is attained. The standard optimistic formulation is defined as follows
\begin{equation}\label{std_opt}
    \min_{x, \hat{y}} \; F(x, \hat{y}) 
    \quad \text{subject to} \quad 
        \hat{y} \in \Psi(x), \; G(x, \hat{y}) \leq 0,\; H(x, \hat{y}) = 0
\end{equation}

\item[\tb{Example:}] 
Consider the following bilevel problem
\begin{equation}
\operatorname*{\text{``min''}}_{x} \; F(x,\hat{y}) = -\hat{y}
\quad \text{subject to} \quad \hat{y} \in \argmin_{y} \{ x - y \mid 0 \le y < x \}, 
\quad x \in [0,1]
\end{equation}

\noindent
The lower-level problem is given by
\[
\min_{0 \le y < x} (x - y),
\]
which is equivalent to maximizing \(y\) over the interval \([0,x)\). The supremum of \(y\) is \(x\), but it is not attained due to the strict inequality. Hence, the reaction set is $\Psi(x) = [0,x)$.

\medskip
\noindent
\textbf{Standard Optimistic Formulation.} The problem can be written as
\begin{equation}
\min_{x,\hat{y}} \; -\hat{y}
\quad \text{subject to} \quad 
\hat{y} \in [0,x), \; x \in [0,1]
\end{equation}
For any fixed \(x\), the objective is minimized by choosing \(\hat{y}\) as large as possible, i.e., \(\hat{y} \to x^{-}\). This yields an infimum value of \(-x\), which is not attainable since \(\hat{y} \neq x\). Consequently, the overall infimum of the problem is \(-1\), obtained as \(x \to 1\) and \(\hat{y} \to 1^{-}\), but no feasible solution attains this value. Therefore, the standard optimistic formulation does not admit an optimal solution.

\medskip
\noindent
\textbf{Real Optimistic Formulation.} The corresponding real optimistic formulation is
\begin{equation}
\min_{x \in [0,1]} \; \inf_{\hat{y} \in \Psi(x)} (-\hat{y})
\end{equation}
Substituting \(\Psi(x) = [0,x)\), we obtain
\[
\inf_{\hat{y} \in [0,x)} (-\hat{y}) = -x
\]
Thus, the problem reduces to
\[
\min_{x \in [0,1]} (-x),
\]
which admits the optimal solution \(x^* = 1\) with optimal value \(-1\).

\medskip
\noindent
\textbf{Discussion.} This example highlights a fundamental difference between the two formulations. The standard optimistic formulation requires the explicit selection of a lower-level optimal solution, which may fail when the lower-level optimum is not attained. In contrast, the real optimistic formulation relies on the infimum over the reaction set and remains well-defined even in the absence of an attained lower-level solution. Consequently, the real formulation admits an optimal solution in this example, whereas the standard formulation does not.

\end{description}

\subsection{Solutions under Optimistic and Pessimistic Formulations}\label{app_solution_opt_pess_formulations}
Recall the bilevel formulation
\begin{align*}
\operatorname*{\text{``min''}}_{x} \; F(x, \hat{y}) = x^2 + \hat{y}^2 + \hat{y} - x + \tfrac{1}{2} \\
& \hspace{-5cm} \text{s.t.} \\
& \hspace{-5cm}-1 \le x \le 1, \\
& \hspace{-5cm} -x + \hat{y} \le 1, \\
& \hspace{-4.5cm} x - \hat{y} \le 1, \\
& \hspace{-4.5cm} \hat{y} \in 
   \argmin_{y}
   \Big\{
      f(x, y) = (y - x)^4 - (y - x)^2
      \quad \text{s.t.} \quad  x^2 + y^2 \le 1, \; x + y \le 1
   \Big\}
\end{align*}
\begin{figure}[htb]
    \centering
    \includegraphics[width=0.7\textwidth]{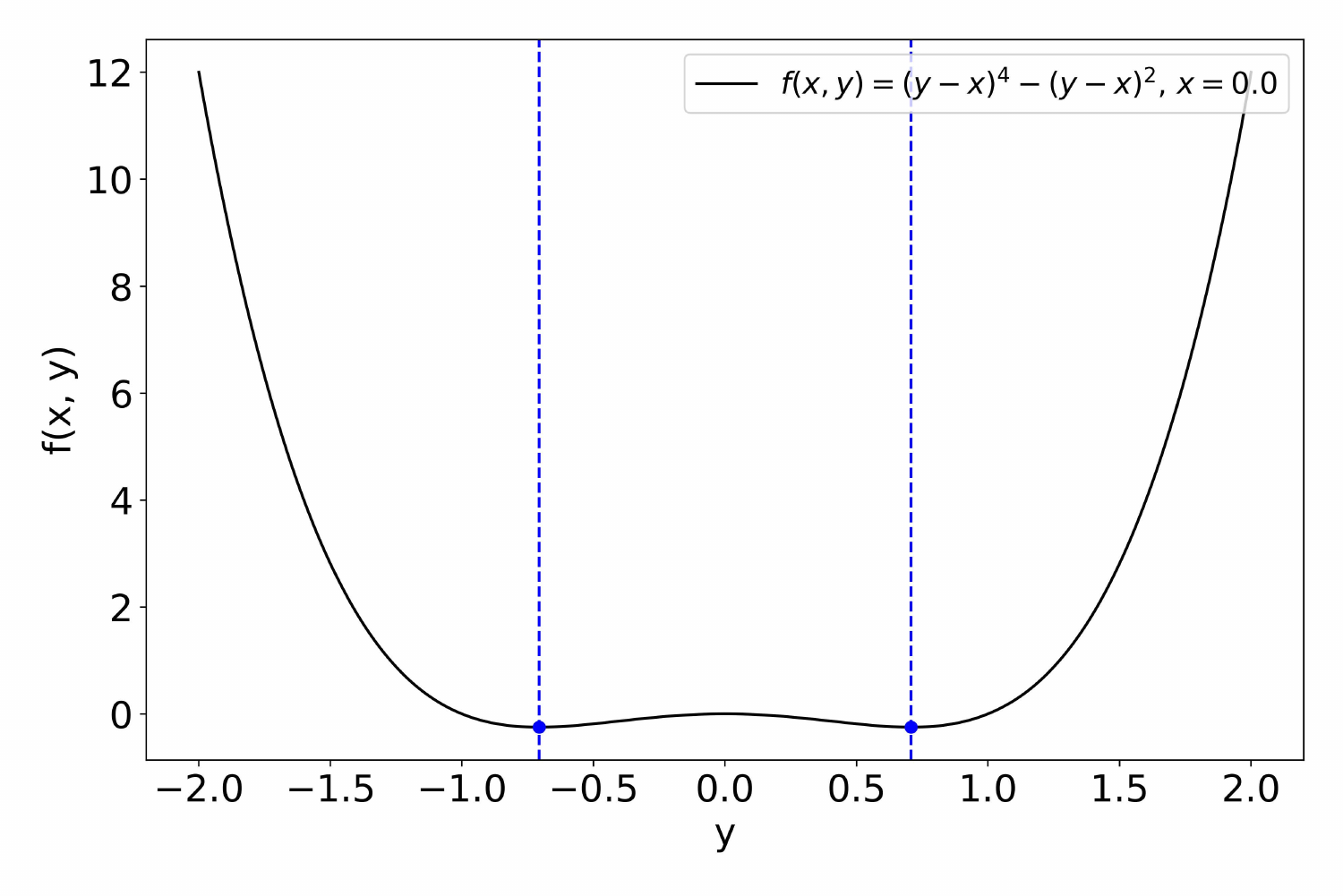}
    \caption{Lower-level objective and its minima for a fixed upper-level decision.}
    \label{fig:reaction_set}
\end{figure}

Figure~\ref{fig:reaction_set} visualizes the lower-level objective function corresponding to a fixed upper-level decision \((x = 0)\). The plot illustrates the minima \(\hat{y} = \pm \frac{1}{\sqrt{2}}\) of the lower-level function, which are used to derive the reaction set in the context of bilevel optimization. To analyze the follower’s behavior, we construct the Lagrangian of the lower-level problem
\[
\mathcal{L}(x,y,\lambda_1,\lambda_2) = (y-x)^4 - (y-x)^2 + \lambda_1(x^2+y^2-1) + \lambda_2(x+y-1),
\]
where $\lambda_1, \lambda_2 \ge 0$ are the multipliers associated with the follower constraints. The KKT conditions are
\[
\begin{aligned}
& \text{(stationarity)} && 4(y-x)^3 - 2(y-x) + 2\lambda_1 y + \lambda_2 = 0, \\
& \text{(primal feasibility)} && x^2+y^2 \le 1,\;\; x+y \le 1, \\
& \text{(dual feasibility)} && \lambda_1 \ge 0,\;\; \lambda_2 \ge 0, \\
& \text{(complementarity)} && \lambda_1(x^2+y^2-1)=0,\;\; \lambda_2(x+y-1)=0
\end{aligned}
\]
When $\lambda_1=\lambda_2=0$, stationarity reduces to
\[
4(y-x)^3 - 2(y-x) = 0,
\]
which yields three critical points given by
\[
\hat{y} = x, \quad \hat{y} = x \pm \frac{1}{\sqrt{2}}
\]
Evaluating the second derivative,
\[
\frac{\partial^2 f}{\partial y^2} = 12(y - x)^2 - 2,
\]
we find that $\hat{y} = x$ corresponds to a local maximum \(\left( \frac{\partial^2 f}{\partial y^2} = -2 < 0 \right)\), whereas $\hat{y} = x \pm \frac{1}{\sqrt{2}}$ are global minima \(\left( \frac{\partial^2 f}{\partial y^2} = 4 > 0 \right)\). Hence, the lower-level (KKT) solution set is
\[
\Psi(x) = \left\{ x - \frac{1}{\sqrt{2}},\; x + \frac{1}{\sqrt{2}} \right\}
\]

In the optimistic formulation, the follower selects the $\hat{y}$ that minimizes the upper-level objective $F(x,\hat{y})$ among its global minimizers. Substituting $\hat{y}_o = x - \tfrac{1}{\sqrt{2}}$ gives
\[
F(x,\hat{y}_o) = 2\left(x - \tfrac{1}{2\sqrt{2}}\right)^2 + \left(\tfrac{3}{4} - \tfrac{1}{\sqrt{2}}\right),
\]
which is minimized at
\[
x^* = \tfrac{1}{2\sqrt{2}}, \qquad y^* = -\tfrac{1}{2\sqrt{2}},
\]
with corresponding value
\[
F(x^*,y^*) = \tfrac{3}{4} - \tfrac{1}{\sqrt{2}}
\]

At this stage the BOP is equivalently cast as a Mathematical Program with Equilibrium Constraints (MPEC), in which the equilibrium is provided by the follower’s KKT conditions. This special case is known as a Mathematical Program with Complementarity Constraints (MPCC). The KKT system comprises stationarity, primal feasibility, dual feasibility, and complementarity. At the optimistic optimizer $(x^*,y^*)$, the follower’s constraints are strictly inactive, so $\lambda_1=\lambda_2=0$. Hence, only the stationarity condition enters the leader’s KKT, through the multiplier $\beta$. With
\[
s(x,y,\lambda_1,\lambda_2)=4(y-x)^3-2(y-x)+2\lambda_1 y+\lambda_2=0,
\]
the leader’s KKT requires
\[
\nabla_x F + \beta\,\partial_x s=0, \qquad \nabla_y F + \beta\,\partial_y s=0,
\]
which at $(x^*,y^*)$ yield
\[
\beta=\tfrac{1}{4}\Big(\tfrac{1}{\sqrt{2}}-1\Big)
\]

In the pessimistic formulation, the follower instead selects the $\hat{y}$ that maximizes $F(x,\hat{y})$ among its minimizers. Substituting $\hat{y}_p = x + \tfrac{1}{\sqrt{2}}$ gives
\[
F(x,\hat{y}_p) = 2\left(x + \tfrac{1}{2\sqrt{2}}\right)^2 + \left(\tfrac{3}{4} + \tfrac{1}{\sqrt{2}}\right),
\]
with minimum attained at
\[
x^* = -\tfrac{1}{2\sqrt{2}}, \qquad y^* = \tfrac{1}{2\sqrt{2}},
\]
and objective value
\[
F(x^*,y^*) = \tfrac{3}{4} + \tfrac{1}{\sqrt{2}}
\]
Here too, the problem reduces to an MPCC via the follower’s KKT embedding. As in the optimistic case, the follower’s constraints are inactive at the solution, leaving only the stationarity condition active in the leader’s KKT. This yields the multiplier
\[
\beta=-\tfrac{1+1/\sqrt{2}}{4}
\]
In both optimistic and pessimistic solutions, the follower’s feasibility constraints are slack, so their multipliers vanish and complementarity holds automatically. Consequently, only the follower’s stationarity condition contributes to the leader’s KKT. If a follower constraint were active at the solution, the associated multiplier would be nonzero and the corresponding condition would appear explicitly in the leader’s KKT system.
\end{document}